\documentclass[sigconf]{acmart}

\copyrightyear{2022}
\acmYear{2022} 
\setcopyright{rightsretained}

\acmConference[CCS '22]{Proceedings of the 2022 ACM SIGSAC Conference on Computer and Communications Security}{November 7--11, 2022}{Los Angeles, CA, USA}
\acmBooktitle{Proceedings of the 2022 ACM SIGSAC Conference on Computer and Communications Security (CCS '22), November 7--11, 2022, Los Angeles, CA, USA}
\acmDOI{10.1145/3548606.3560694}
\acmISBN{978-1-4503-9450-5/22/11}

\usepackage{etoolbox}
\makeatletter
\patchcmd{\maketitle}{\@copyrightpermission}{
   \begin{minipage}{0.3\columnwidth}
     \href{https://creativecommons.org/licenses/by/4.0/}{\includegraphics[width=0.90\textwidth]{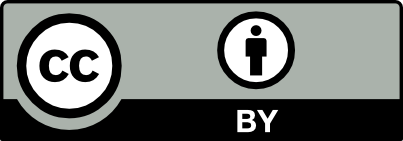}}
   \end{minipage}\hfill
   \begin{minipage}{0.7\columnwidth}
     \href{https://creativecommons.org/licenses/by/4.0/}{This work is licensed under a Creative Commons Attribution International 4.0 License.}
   \end{minipage}

   \vspace{5pt}
}{}{}

\makeatother

\newcommand{\dtrain}{D}

\newcommand{\Ex}[1]{E[{#1}]}

\newcommand{\prob}[1]{\texttt{Pr}({#1})}
\newcommand{\ATE}[1]{\textsc{ATE}({#1})}
\newcommand{\todo}[1]{\textcolor{black}{#1}}

\newcommand{\learnalg}{\mathcal{A}}
\newcommand{\trainparam}{\pi}

\newcommand{\tool}{{\sc Etio}\xspace}

\newcommand{\traintestgap}{train-to-test accuracy gap\xspace}

\newcommand{\model}{f}

\newcommand{\norm}[1]{\left\lVert#1\right\rVert_2}
\newcommand{\accdiff}{\texttt{AccDiff}\xspace}

\newcommand{\lossdiff}{\texttt{LossDiff}\xspace}
\newcommand{\testvar}{\texttt{TestVar}\xspace}
\newcommand{\trainvar}{\texttt{TrainVar}\xspace}
\newcommand{\testacc}{\texttt{TestAcc}\xspace}
\newcommand{\trainacc}{\texttt{TrainAcc}\xspace}
\newcommand{\testbias}{\texttt{TestBias}\xspace}
\newcommand{\trainbias}{\texttt{TrainBias}\xspace}
\newcommand{\testloss}{\texttt{TestLoss}\xspace}
\newcommand{\trainloss}{\texttt{TrainLoss}\xspace}
\newcommand{\oakacc}{\texttt{ShadowAcc}\xspace}
\newcommand{\mlleakacc}{\texttt{MLLeakAcc}\xspace}
\newcommand{\mlleaklacc}{\texttt{MLLeakAcc-l}\xspace}
\newcommand{\mlleaktopacc}{\texttt{MLLeakTop3Acc}\xspace}
\newcommand{\mlleaktoplacc}{\texttt{MLLeakTop3Acc-l}\xspace}
\newcommand{\threshacc}{\texttt{ThreshAcc}\xspace}
\newcommand{\numparams}{\texttt{NumParams}\xspace}
\newcommand{\trainsize}{\texttt{TrainSize}\xspace}
\newcommand{\centroid}{\texttt{CentroidDist}\xspace}

\newcommand{\miacc}{\texttt{MIAcc}\xspace}

\newcommand{\memguard}{\texttt{MemGuardAcc}\xspace}

\newcommand{\numattacks}{$6$\xspace}
\newcommand{\numdefenses}{$2$\xspace}
\newcommand{\numconstraints}{$19$\xspace}
\newcommand{\predpower}{$3-22\%$}
\newcommand{\numbetterpred}{$17/24$}

\newcommand{\predacc}{$0.90$}
\newcommand{\remainingpred}{$7/24$}

\newcommand{\cebigpvalue}{$1$}
\newcommand{\msebigpvalue}{$4$}
\newcommand{\totalcequeries}{$16$}
\newcommand{\totalmsequeries}{$16$}

\newcommand{\totalpriorq}{$18$\xspace}
\newcommand{\falsepriorq}{$7$\xspace}
\newcommand{\truepriorq}{$9$\xspace}

\usepackage{balance}

\usepackage{amsmath,amsfonts, amsthm}
\usepackage{mathtools}
\usepackage{algorithmicx,algpseudocode}
\usepackage{graphicx,rotating}
\usepackage{textcomp}
\usepackage{url}
\usepackage{xspace}
\usepackage[ruled,vlined, linesnumbered]{algorithm2e}
\SetAlFnt{\small}
\SetAlCapFnt{\small}
\SetAlCapNameFnt{\small}
\usepackage{subcaption}
\usepackage{bbm}
\usepackage{pifont}
\usepackage{enumitem}

\usepackage{tikz}
\usepackage{multirow}
\usepackage{thmtools, thm-restate}
\usepackage{hyperref}

\usepackage{color, colortbl}
\definecolor{LightCyan}{rgb}{0.88,1,1}

\usepackage[T1]{fontenc}

\hypersetup{
    colorlinks=true,
    linkcolor=blue,
    filecolor=magenta,
    urlcolor=cyan,
}

\usepackage{siunitx}
\usepackage{booktabs}
\usepackage{csvsimple}
\csvstyle{mystyle}{
    tabular=ccccc,
    table head=\toprule,
    table foot=\bottomrule,
    no head,
    late after line=\\,
    late after first line=\\\midrule,
    }
\newcommand\independent{\protect\mathpalette{\protect\independenT}{\perp}}\def\independenT#1#2{\mathrel{\rlap{$#1#2$}\mkern2mu{#1#2}}}

\begin{document}

\title{Membership Inference Attacks and Generalization:\\ A Causal Perspective}

\author{Teodora Baluta}
\affiliation{%
  \institution{National University of Singapore}
  \country{Singapore}}
\orcid{0000-0003-3655-9810}
\email{teobaluta@comp.nus.edu.sg}

\author{Shiqi Shen}
\affiliation{%
  \institution{National University of Singapore}
  \country{Singapore}}
\orcid{0000-0002-0469-3663}
\email{shiqi04@u.nus.edu}

\author{S. Hitarth}
\authornote{Work done while interning at National University of Singapore.}
\affiliation{%
  \institution{Hong Kong University of Science and Technology}
  \country{Hong Kong}}
\orcid{}
\email{hsinghab@connect.ust.hk}

\author{Shruti Tople}
\affiliation{
  \institution{Microsoft Research}
  \city{Cambridge}
  \country{United Kingdom}}
\email{shruti.tople@microsoft.com}

\author{Prateek Saxena}
\affiliation{%
  \institution{National University of Singapore}
  \country{Singapore}}
\email{prateeks@comp.nus.edu.sg}

\begin{abstract}
Membership inference (MI) attacks highlight a privacy weakness in present stochastic training methods for neural networks. It is {\em not} well understood, however, why they arise. Are they a natural consequence of imperfect generalization only? Which underlying causes should we address during training to mitigate these attacks? Towards answering such questions, we propose the first approach to explain MI attacks and their connection to generalization based on principled {\em causal reasoning}. We offer causal graphs that quantitatively explain the observed MI attack performance achieved for $6$ attack variants. We refute several prior non-quantitative hypotheses that over-simplify or over-estimate the influence of underlying causes, thereby failing to capture the complex interplay between several factors. Our causal models also show a new connection between generalization and MI attacks via their shared causal factors. Our causal models have high predictive power (\todo{\predacc}), i.e., their analytical predictions match with observations in unseen experiments often, which makes analysis via them a pragmatic alternative.

\end{abstract}

  


\maketitle

\section{Introduction}

As the use of machine learning proliferates, privacy has become a key concern in machine learning~\cite{gdpr} with several classes of attacks being discovered.
Membership inference (MI) attacks, which have led to a flurry of works recently~\cite{long2018understanding,choquette2021label,hui2021practical,liu2021ml,zanella2020analyzing,shokri2021privacy,liu2021ml,li2021membershipleakage,song2019membership,song2021systematic,hayes2019logan,nasr2019comprehensive,nasr2018machine,zhang2021membership,liu2021encodermi}, capture the advantage of an adversary in distinguishing samples used for training from those that were not.
There is clear empirical evidence that MI attacks are effective and many new attack variants are emerging. At the same time, there is currently no systematic framework to understand {\em why} standard training procedures leave deep nets susceptible to MI attacks.

There are two incumbent approaches to understanding why deep networks are susceptible to MI attacks~\cite{yeom2018privacy,sablayrolles2019white,song2021systematic,liu2021ml,truex2019demystifying}. The first tries to offer {\em fully mechanistic} explanations derived from theoretical analysis. For instance, a line of work tries to mathematically model the stochastic mechanism of training models (e.g., using stochastic gradient descent (SGD)) with enough precision~\cite{sablayrolles2019white}. The most commonly accepted mechanistic explanation is that ML models leak training data because they fail to generalize, measured through quantities like overfitting gap, accuracy gap, and so on~\cite{yeom2018privacy,sablayrolles2019white}. 
This approach is appealing and is actively progressing, but at the same time, modeling the training process with closed-form mathematical expressions is an inherently difficult problem. Predictions from these mechanistic explanations often do not agree with observations in experiments, because the approximations or assumptions made in the analysis may not hold in practice~\cite{yeom2018privacy}. Furthermore, generalization offers a one-way explanation: if models generalize almost perfectly in a particular sense, then MI attacks are expected to be ineffective on average. It does not say how well MI attacks will work for models that may have not generalized perfectly, which is often the case in practice. As a result, there have been many different hypothesized root causes that go beyond direct measures of classical generalization, such as model capacity and architecture. In short, no single coherent mechanistic explanation today predicts the average performance of existing MI attacks well.

A second approach to explaining MI attacks is based on {\em statistical testing} of hypotheses: Researchers intuit about the root cause, run experiments, and then report statistical correlations between the hypothesized cause and the performance of the MI attacks~\cite{shokri2017membership,truex2019demystifying,liu2021ml,li2021membership,song2021systematic}. For example, several works have suggested that the empirically observed overfitting gap or the accuracy difference between training and testing sets explain MI attacks. This approach, while being important in its own right, fails to provide satisfactory explanations often as well. Guessing which root causes are really responsible for attacks is difficult; after all, the stochastic process of standard training procedures is complex and is affected by multiple possible sources of randomness, hyper-parameter selection, and sampling bias. Furthermore, correlation is not always causation. Confusing the two can result in overly simplistic explanations of the true phenomenon at hand and lead to paradoxes. Lastly, the purely empirical approach leaves no room to accommodate mechanistic axioms (things we know that ought to be true from our theoretical understanding)---if the observations do not correlate with mechanistically derived facts, then they remain unexplained.  

\paragraph{Our approach.} We propose a new approach that explains MI attacks through a {\em causal model}. A causal model is a graph where nodes are random variables that abstractly represent properties of the underlying stochastic process and edges denote cause-effect relationships between them. We can model the process of sampling data sets, picking hyper-parameters like the size of the neural network, output vectors, generalization parameters like bias and variance, and predictions from MI attack procedures as random variables. These random variables can be measured empirically during experiments. We can then both encode and infer causal relationships between nodes quantitatively through equations. Edges in our causal model are of two types: 1) mechanistically derived edges denote known mathematical facts derived from domain knowledge (prior work, definitions, etc.); and 2) relations inferred from experimental observations using {\em causal discovery} techniques.

Our causal approach is substantially different from prior works and enables much deeper and principled analysis. 
The causal model, once learnt, acts like a predictive model---one can ask what will be the expected performance of a particular MI attack if the ``root causes'' (random variables in the model) were to have certain values not observed during prior experiments. Such estimation can be done without running expensive experiments. A causal model allows us to "single out" the effect of one variable on the MI attack performance. Such queries go beyond just observing statistical correlations because they need to reason about other variables that might affect both the cause and the outcome (the attack performance). To carefully solve these queries, we leverage the principled framework of causal reasoning known as {\em do-calculus}.
It allows us to perform systematic refutation tests, which avoids confusing causation with correlation.
 Such tests quantitatively tell us how well the model fits the observed data and answer ``what if'' style of questions about surmised root causes.
Further, we can compare causal models obtained for two different attacks to understand how their manifestation differs, or compare models with and without an intervention (e.g., by applying a defense) for a given attack to understand which root causes it neutralizes. Causal models offer a more principled and interactive way of examining MI attacks.

\paragraph{Resulting Findings.} To showcase the utility of our approach, we study  \todo{\numattacks} well-known MI attacks and \todo{\numdefenses} defenses for deep neural networks trained using standard SGD training procedures. We analyze a list of intuitive ``root causes'' which have been suggested in prior works and formally specify them as \todo{$9$} causal hypotheses. We analyze each of these \todo{9} hypotheses for ML models with \todo{2} types of loss functions, so we evaluate on \todo{$18$} formalized hypotheses. Several salient findings have resulted from our causal analysis. First, we refute \todo{\falsepriorq/\totalpriorq} previously hypothesized causes, highlighting the perils of understanding MI attacks purely from intuition or from statistical correlational analysis. 
Second, we find that different causes contribute differently to the average attack accuracy, dispelling the idea that a single explanation suffices for all \todo{\numattacks} MI attacks we study.
Our causal approach also models the attacks well (\todo{\predacc}), i.e., predict the observed attack accuracy. This betters prior single-cause explanations by \todo{\predpower} in \todo{\numbetterpred}. 
Third, we show that two stochastic parameters inherent in the training process, namely \textbf{Bias} and \textbf{Variance}, govern both generalization achieved by ML models~\cite{jiang2018predicting,yang2020rethinking,neal2018modern} and MI attack accuracy. This offers a more nuanced lens to connect generalization and MI attack accuracy from that offered by prior works~\cite{yeom2018privacy,long2018understanding,carlini2019secret}.
Fourth, we show that defenses against MI attacks based on regularization reduce the influence of some of root causes, but fail to completely remove their effect.

\paragraph{Summary of Contributions.}
We propose the first use of causal analysis for studying membership inference attacks on deep neural networks. We derive causal models for $6$ MI attacks by combining both known domain-specific assumptions and observations made from experiments. Our key contribution is a new quantitative connection between MI attacks and generalization, which enables refuting claims about causation with finer accuracy.

\subsection*{Availability}
Our prototype implementation is publicly available on GitHub\footnote{At \url{https://github.com/teobaluta/etio}.}.

\section{Motivation}
\label{sec:motivation}

Many intuitive explanations for privacy leakage have been put forward in prior works.
The most widely accepted claim is that ``{\em overfitted classifiers are more susceptible to MI attacks}'', which has been backed by experimental correlational analyses~\cite{shokri2017membership,truex2019demystifying,li2021membership,liu2021ml,long2018understanding,leino2020stolen,yeom2018privacy,sablayrolles2019white}.
To evaluate the level of overfitting though, two different metrics have been proposed: the difference in the loss of training and non-training samples~\cite{yeom2018privacy,sablayrolles2019white}, as well as the train-to-test accuracy gap~\cite{shokri2017membership,truex2019demystifying,li2021membership,liu2021ml,long2018understanding,leino2020stolen}. However, empirical evidence to the contrary has also been observed--both MI attacks and extraction attacks have been reported on well-generalized models~\cite{carlini2019secret,carlini2020extracting,long2018understanding}. Other potential contributing factors, such as model complexity / structure~\cite{shokri2017membership,song2021systematic,nasr2019comprehensive}, the size of the training set~\cite{shokri2017membership,nasr2019comprehensive}, the diversity of the training samples~\cite{shokri2017membership}, how close a target model to attack is to the shadow model~\cite{salem2019ml}, and so on have been proposed, creating an unclear picture of why MI attack arise. We highlight \todo{$9$} common hypotheses claimed in prior works below:

\begin{enumerate}[label=(H\arabic*)]

    \item The overfitting gap is the cause of MI attacks that use multiple shadow models in the inference~\cite{shokri2017membership}.
    
    \item ``The main idea behind our shadow training technique is that similar models trained on relatively similar data records using the same service behave in a similar way''~\cite{shokri2017membership}.
    
    \item Beyond overfitting, model complexity influences the membership inference attack accuracy~\cite{shokri2017membership}.
    
    \item The size of the training set is a contributing factor to the success of MI attacks~\cite{shokri2017membership}.
    
    \item The shadow model attack works even if the attack uses only one shadow model~\cite{salem2019ml}.
    
    \item ``If a model is more overfitted, then it is more vulnerable to membership inference attack.'' - for MI attacks that use a single shadow model~\cite{salem2019ml}.

    \item There is no difference in the attack accuracy if we use the top-3 predictions in descending order vs. the whole prediction vectors ~\cite{salem2019ml}.
    
    \item Shadow model-based attacks transfer when there is a clear decision boundary between members and non-members~\cite{salem2019ml}.
    
    \item The average generalization error explains the advantage of the threshold-based adversary (even when this attack assumes that the error is normally distributed)~\cite{yeom2018privacy}.
\end{enumerate}

It is natural to ask: To what extent are these explanations correct? Do these hypothesized factors universally explain all MI attacks equally?  What does achieving a certain level of generalization (eliminating overfitting) imply towards reducing MI attacks? 
Answering such queries requires a principled framework for reasoning even to phrase the right statistical quantities to measure---it is something that is easily prone to fallacious reasoning.

\subsection{Pitfalls of Testing with Correlations}

Let us consider two of the prior work hypotheses: (H1) Higher difference between train and test accuracy leads to higher MI attacks; and (H3) An increase in model complexity increases privacy leakage, i.e., larger models are more susceptible.
One of the most prominent approaches to validating such hypotheses today is experimental validation through statistical correlation analysis~\cite{truex2019demystifying,shokri2017membership,long2018understanding,liu2021ml}. The analysis proceeds by observing how the train-test accuracy gap and attack accuracy for a chosen MI attack changes under different choices of model complexity (number of model parameters).
For concrete illustration, we run a small-scale experiment for the multiple shadow model attack~\cite{shokri2017membership}. We train \todo{$2$} deep nets with varying number of parameters on \todo{CIFAR10} dataset. We average the observed training and testing accuracy of the deep neural networks under multiple samples of the training datasets. For each of these models, we also run the shadow model attack separately~\cite{shokri2017membership}, using a disjoint part of the training dataset for the shadow model training.

\begin{figure}[t]
\centering
\includegraphics[width=0.75\linewidth]{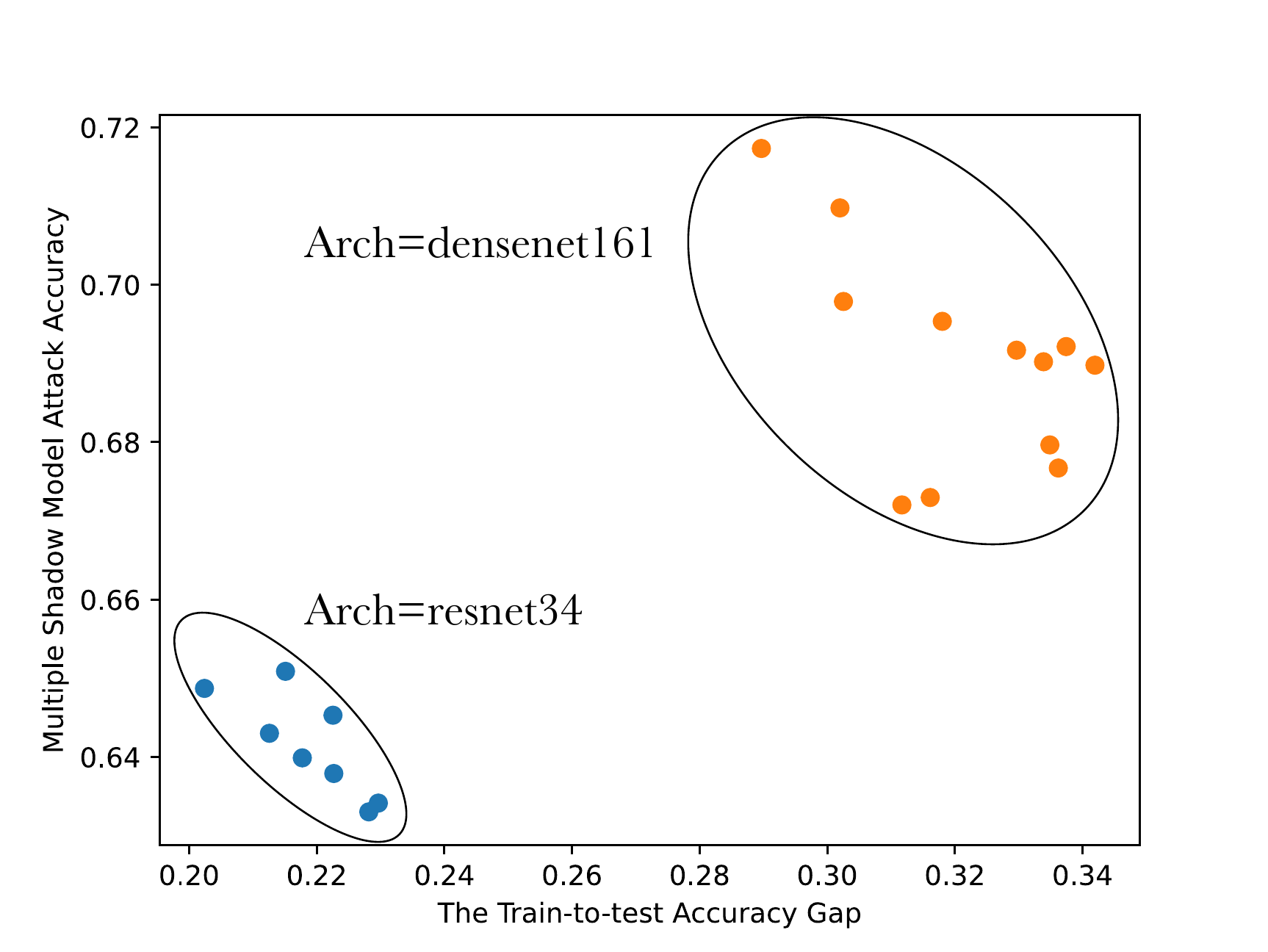}
\caption{High \traintestgap correlates with high attack accuracy in general, but if clustered on architecture type, an inverse relationship is visible.}
\label{fig:stats-example}
\end{figure}

\paragraph{Hidden Causes.}
In Fig.~\ref{fig:stats-example}, on the X-axis we plot the average \traintestgap against the attack accuracy on the Y-axis. The overall trend of the relationship between \traintestgap and the attack accuracy is positive, i.e., the larger the \traintestgap, the higher the attack accuracy.
But this trend is not seen in Fig.~\ref{fig:stats-example}, when 
we group the trained neural nets by other criteria such as model complexity and architecture. We observe they are clumped into two clusters. Within those clusters, the observed trend is the {\em reverse}: the larger the accuracy gap, the lower the attack accuracy!
This turns out to be a fallacious conclusion, because it fails to account for the effect of other factors such as model complexity or architecture indirectly on \traintestgap or directly on the attack accuracy.

This paradox arises because there are {\em confounding factors} or {\em confounders}~\cite{pearl2000models,simpson1951interpretation} wherein different sub-populations of the data have contradictory statistical properties. Similar paradoxes can arise due to {\em selection bias}~\cite{pearl2000models,cai2008identifying,bareinboim2012controlling} or  {\em collider bias}~\cite{berkson1946limitations,pearl2000models}. Without resolving such issues, it is difficult to decide whether one should try reducing the train-to-test accuracy gap or model complexity.

\paragraph{Singling Out.}
When we want to decide which factors influence the end outcome more than the others, such as when designing practical defenses, one would like to ``single out'' the main causes and quantify how much they affect the expected MI attack accuracy.
In our running example, the model complexity affects both variables, the train-to-test accuracy gap as well as the MI attack accuracy. It would be difficult to {\em quantify} how much it affects attack accuracy directly and how much indirectly via train-to-test accuracy gap without a more principled analysis of the observed data.

\begin{figure}[t]
\centering
\includegraphics[width=0.9\linewidth]{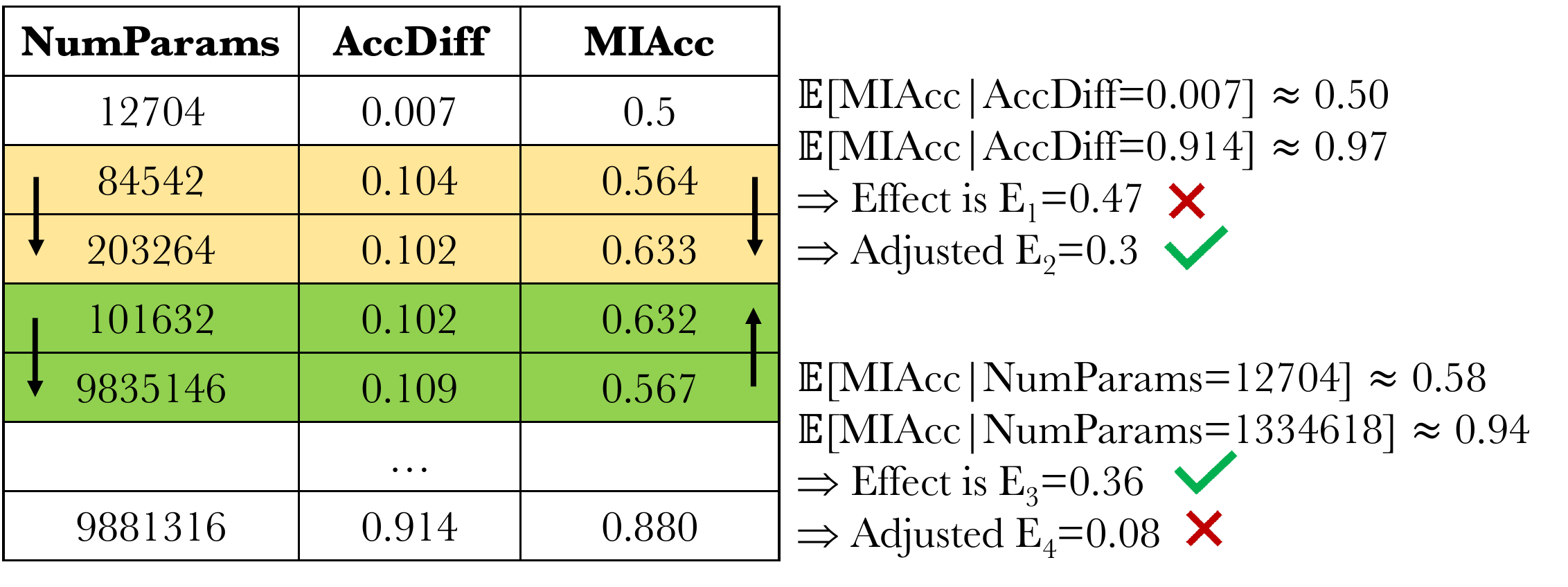}
\caption{Reporting average conditional probabilities is not always correct. For example, if we are to estimate the effect of \traintestgap on the MI attack accuracy from the data shown, the conditional over-estimates the effect by \todo{$0.17$}. For measuring the effect of model size though, the second estimate shown is correct since over-controlling for the \accdiff incorrectly decreases the effect to \todo{$0.08$} from \todo{$0.36$}.}
\label{fig:confounding}
\end{figure}

To understand the challenge, let us say we want to estimate how changing the \traintestgap from \todo{$a=0.007$} to \todo{$b=0.914$} (which we observe in practice) affects the MI attack accuracy. Let us assume we train more NNs and that we now know that the model complexity is a confounding factor for both. A naive way to analyze this is to statistically estimate the following quantity: $E_1=\Ex{\miacc|\accdiff\approx 0.9}- \Ex{\miacc|\accdiff\approx 0}$ where the \miacc is the attack accuracy and the \accdiff is the \traintestgap.
Fig.~\ref{fig:confounding} shows our new experimental observations and conditional probability estimates from data which reveals that the estimated expected effect is $\todo{E_1=0.47}$.
It is misleading to conclude that a change in \traintestgap will have a large impact on the attack performance since we know that there is a confounder. 

If we want to single out the effect of changes in \traintestgap, the correct way is the following: Find samples with the same values for the model complexity but different values of \traintestgap, from which we then compute the difference these produce on the attack accuracy. This is called analytically ``controlling'' for the confounding factor\footnote{Controlling for a variable means binning data according to measured values of the variable}.
This corresponds to analytically computing how the system would behave under randomized values of model complexity. Such randomization ``nullifies'' or ``smoothens out'' the effect of model complexity. If we do this carefully, it turns out that the actual estimated effect $E_2$ when \accdiff ranges from $a$ to $b$ is expressed by the following quantity:
\begin{align*}
E_2=\sum_{z}\Ex{\miacc|\accdiff=a,\numparams=z}\prob{\numparams=z}\\-
\sum_{z}\Ex{\miacc|\accdiff=b,\numparams=z}\prob{\numparams=z}
\end{align*}
This leads to an estimated effect of \todo{$E_2=0.3$}, as per our data (Section~\ref{sec:eval})---significantly lower than the naive analysis above.

\begin{figure}[t]
\centering
\begin{subfigure}[t]{0.5\linewidth}
    \centering
    \includegraphics[width=0.8\linewidth]{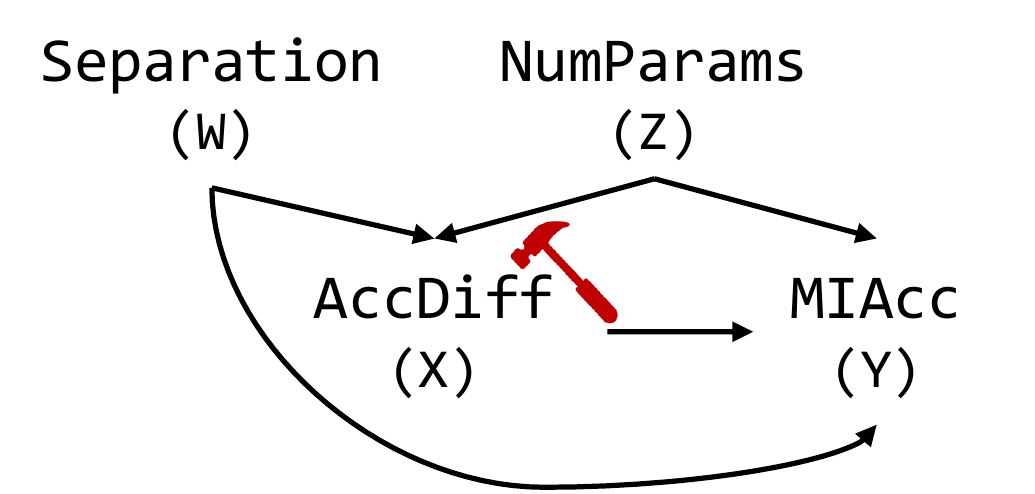}
    \subcaption[]{}
    \label{fig:H8-direct-model}
\end{subfigure}%
\begin{subfigure}[t]{0.5\linewidth}
    \centering
    \includegraphics[width=0.8\linewidth]{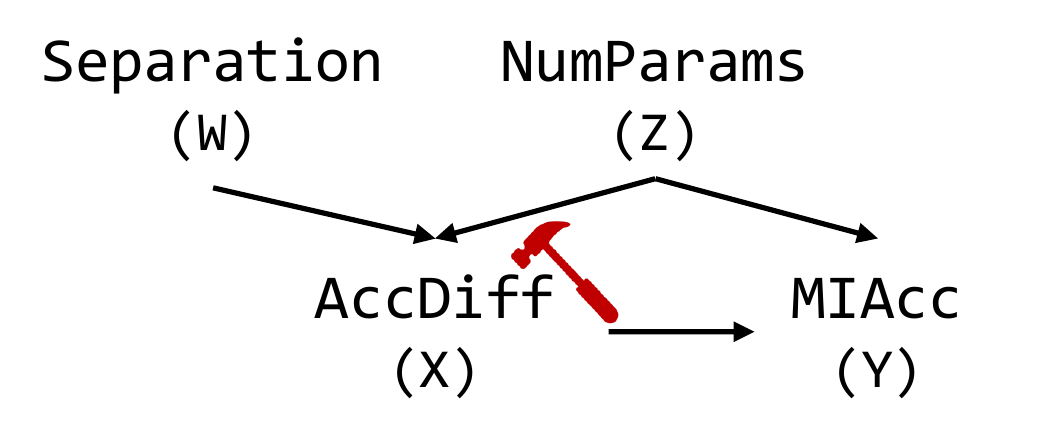}
    \subcaption[]{}
    \label{fig:H8-only-indirect-model}
\end{subfigure}
\caption{
The query to estimate varies by assumptions chosen. If we assume that separation score influences the MI attack accuracy (Fig.~\ref{fig:H8-direct-model}), we should control for two confounding factors, the separation score and the number of parameters. The resulting effect is \todo{$0.12$}. %
If we assume otherwise (Fig.~\ref{fig:H8-only-indirect-model}), we should not ``control'' for the separation score, otherwise it results in a much larger effect of \todo{$0.68$}.}
\label{fig:spec-assumptions}
\end{figure}

\paragraph{Avoiding Over-controlling.} It may be tempting to control for all factors that may influence the outcome. But arbitrarily controlling for variables leads to fallacious reasoning as well. For example, if we want to estimate the effect of the model complexity on the attack accuracy, should we now control for the \traintestgap? If we were to control for the \traintestgap, then the estimated effect $E_4$ when \numparams varies from $a'$ to $b'$ is given by (also shown in Fig.~\ref{fig:confounding}):
\begin{align*}
E_4=\sum_{z}\Ex{\miacc|\numparams=a',\accdiff=z}\prob{\accdiff=z}\\-
\sum_{z}\Ex{\miacc|\numparams=b',\accdiff=z}\prob{\accdiff=z}
\end{align*}
The above expression is analogous to the case where we controlled for model complexity, except we are controlling for the \traintestgap now. The estimated value from experiments for this statistic is \todo{$E_4=0.08$}. This is, however, an incorrect analysis.  If we analytically control for the \traintestgap, then we are actually biasing the total effect that the model complexity has on the attack, as we are ``blocking'' (failing to distinguish) its indirect effect through the \traintestgap.
The correct statistical quantity, in this case, turns out to be  $E_3=\Ex{\miacc|\numparams=a'}-\Ex{\miacc|\numparams=b'}$, i.e., the total effect model complexity has on the attack accuracy. 
The estimated effect is \todo{$E_3=0.36$}---a lot higher than that obtained from the incorrect analysis, and corresponds to the second unadjusted estimate in Fig.~\ref{fig:confounding}.
Another similar example of bad control or over-controlling is the bias amplification problem or pre-treatment control~\cite{pearl2012class} (illustrated later in Fig.~\ref{fig:spec-assumptions}).
The main takeaway is that a principled framework would tell us which quantities to estimate, avoiding over-controlling in experiments and false conclusions.

\paragraph{Large Number of Possible Factors.} When moving beyond a couple of factors to consider, the reasoning can become more complicated.
To build on our previous example, let us now introduce another factor, related to (H8): the separation between members and non-members. Our hypothesis is that the separation between members and non-members is influenced by both the model complexity and the \traintestgap, and in turn it influences the attack accuracy.
How does changing the separation then affect the attack accuracy? 
To answer this question, notice that the separation is influenced by the model complexity and the \traintestgap, both of which influence the attack accuracy.
Similar to what we described so far, one will then have to make sure to randomize these factors in order to obtain the effect of the separation on the MI attack accuracy.
For every additional factor, though, we need to do enough experiments to ``randomize'' our estimates so that they correctly compute the effect on the attack accuracy. It is easy to see that the number of experiments one needs to run quickly starts to grow large as the number of factors considered increases.

\paragraph{Importance of Specifying Assumptions.} 
So far, we have considered cases where certain causal relationships exist and we are trying to correctly estimate the effect of certain factors on the outcome. But, how can we start to test our assumptions, i.e., whether a causal relationship exists at all? Such refutation is hard to do, in general. A practical recourse is that one can specify their assumed beliefs and hope to refute quantitatively under the assumptions. The choice of assumptions matters critically to the outcome.
To illustrate this, consider the hypothesis (H8) again, which introduces a separation score that measures the distance between members and non-members as a factor for the single shadow model MI attack~\cite{salem2019ml} (so far, we have considered the multiple shadow one~\cite{shokri2017membership}). Deciding whether the separation score has any direct influence on the MI attack is critically important---if we choose to assume so, we get one set of conclusions, if we do not, we get another. 
When the separation has a direct influence on MI attack, then the principled analysis to estimate the effect of \traintestgap is similar to the case of estimating the model complexity.
We estimate the effect on the attack accuracy is \todo{$0.12$} on our set of experiments (the details of our experimental setup are in Section~\ref{sec:eval-setup}). 
In the alternative scenario, the correct quantity to estimate is below, leading to the estimated effect of the \traintestgap to be \todo{$E_5=0.68$} when it varies from the $a$ to $b$.  \begin{align*}
 E_5=\sum_z\Ex{\miacc|\accdiff=a,\numparams=z}\prob{\numparams=z}\\-\sum_z\Ex{\miacc|\accdiff=b,\numparams=z}\prob{\numparams=z}
\end{align*}

We illustrate the differences in the two sets of assumptions in Fig.~\ref{fig:spec-assumptions} which exacerbates the bias amplification problem.
We point out that prior works do {\em not} specify such assumptions or beliefs explicitly, making it impossible to refute or validate such hypotheses.

\paragraph{Goodness of Explanations.}
Given the subjectivity of assumptions and computational limits on the number of experiments one can run, it is difficult to analytically argue
that a given explanation is ``correct'' or certain hypothesis is conclusively ``incorrect''.
How then can we measure how good or correct is an explanation? A practical way to do so is look at the {\em predictive power} of a given explanation, i.e., measure how accurately it can predict the outcome (MI attack accuracy) under experimental settings {\em not} seen during creating the explanation. The highlighted prior hypothesis (H1)-(H9) often have predicted power well below \todo{$85\%$} on average, offering less satisfying results. In contrast, our approach has predictive power of \todo{\predpower} higher than the prior work hypotheses, for most attacks we study.

\section{The Causal Modelling Approach}
\label{sec:approach-overview}

The prior common hypotheses, some of which are derived from mechanistic explanations or theoretical analyses, provide a good starting point to reason about potential factors of the MI attacks. But, as shown throughout Section~\ref{sec:motivation}, there are several pitfalls in identifying the factors and estimating their effect.
Our aim is to infer a model defined over a set of potential factors and a given MI attack, not just a simple correlation of each factor separately with the MI attack.
The model explicitly defines relationships between factors and the MI attack and between themselves. It should also provide a query interface for the following query types:
\begin{itemize}
    \item {\em Prediction Queries}: Given some observed values $a_1, \ldots, a_n$ of the potential factors $X_1,\ldots,X_n$, what is the predicted MI attack accuracy $Y$: $\Ex{Y|X_1=a_1,\ldots,X_n=a_n}$?
    \item {\em Interventional Queries}: What is average effect of a potential factor on the attack accuracy if that factor had taken a different value from the observed one? 
\end{itemize}

The prediction query consists of a set of assignments of observed  (from running experiments) values for a set of factors, and the target variable $Y$. The output of this query is the expected MI attack accuracy conditioned on the observed values. Such queries help us measure how well the causal model agrees with observations in experiments.
The interventional query is a ``what if'' query. It consists of two variables: the potential cause variable, called the treatment variable $X$, and the desired outcome variable $Y$. For instance, to estimate the effect of the \traintestgap on the attack, we ask if the \traintestgap had taken the value $0.1$ compared to having no \traintestgap, what is the expected MI attack accuracy? 
We want our causal model to be 1) {\bf Accurate}, i.e., to have a goodness of fit, and 2) {\bf Principled}, i.e., the estimated effect is rigorously computed.

\begin{figure}[t]
\centering
\includegraphics[width=0.65\linewidth]{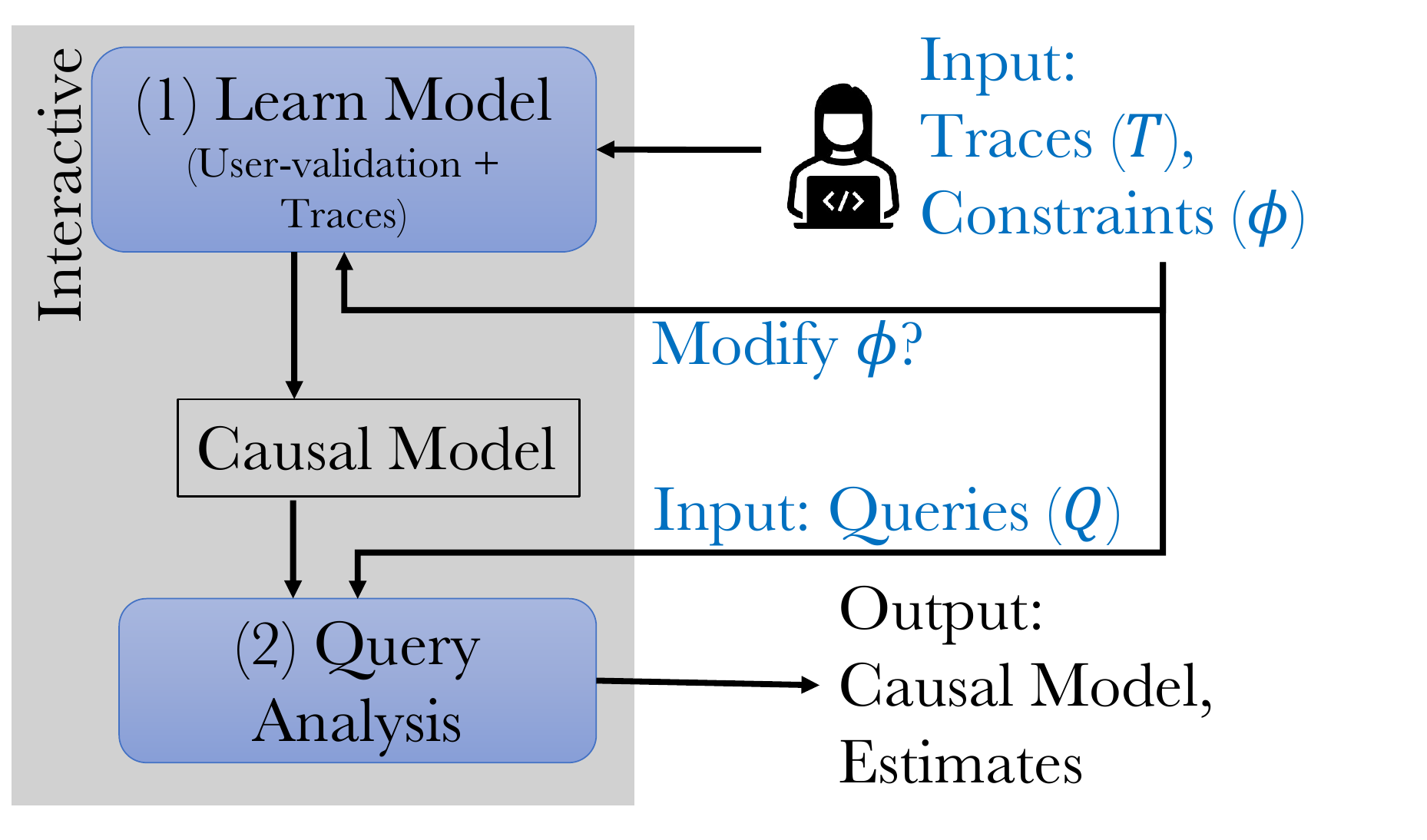}
\caption{The user provides the domain knowledge $\phi$ and the traces $T$. The traces contain observations of the values that the factors of interest take for the training algorithm $\mathcal{A}$ and attack $A$. After an interactive step, the user fixes on a causal graph on which the input queries are analyzed.
}
\label{fig:overview}
\end{figure}

We introduce a novel methodological shift: Our approach proposes to use {\em causal reasoning} to disambiguate potential factors of MI attacks while satisfying the goals highlighted above.
We combine mechanistic explanations from domain knowledge with automated inferences from empirical data to infer a causal model. Specifically, causal models are directed acyclic graphs (DAGs) defined over a set of variables and a set of directed edges\footnote{We thus interchangeably use causal model and causal graph.} where each directed edge from variable $X$ to $Y$ represents a ``$X$ causes $Y$'' relationship. On top of the graph structure, causal models are quantitative: each node in the graph has associated an equation that describes the cause-and-effect relationship between the node and its parents.
Our approach is necessarily synergistic: without domain knowledge constraints, purely observational data cannot distinguish cause-and-effect; and without observational data, we cannot test our intuitions or extract more insights from experiments.
The whole process is interactive and it is illustrated in Fig.~\ref{fig:overview}.
Initially, the user identifies potential factors or variables of interest of the underlying training and attack procedure such as training hyper-parameters, \traintestgap and the outcome MI attack accuracy.
We model these as ``random variables'' that can be observed and measured.
The user generates the set of observations for these variables which we call traces $T$, by effectively running experiments and recording the values of the variables.
The domain knowledge constraints ($\phi$) formally describe the mechanistic explanations, facts or assumptions that stem from the data-generating process, e.g., the training and attack procedures.
Given the traces and the domain constraints, we output a causal graph which the user can choose to further refine (\textit{Modify $\phi$} step in Fig.~\ref{fig:overview}).
Finally, the causal models encode cause-and-effect relationships by construction and can support the $2$ types of queries.
The user can specify these queries ($Q$) formally and obtain estimates on the inferred causal graph (step (2) in Fig.~\ref{fig:overview}).

\noindent\textbf{Inputs \& Outputs.}
We have prototyped our approach in an interactive tool called \tool\footnote{Available at \url{https://github.com/teobaluta/etio}.} and envision model practitioners and researchers as its main users.
\tool minimally requires a set of traces corresponding to the runs of a specific training algorithm ($\learnalg$) over the training dataset ($D$).
These traces record values of a set of properties about the training algorithm, model, and the performance metrics of the attack procedure ($A$)--all of which we call variables ($V$).
The user additionally specifies a set of domain knowledge constraints which encode knowledge that two variables are not in a causal relationship, e.g., if they are caused by the same unmeasurable/confounder variable (which we denote as \textsc{Forbid} constraints) or that there is a causal relationship between two variables (denoted as \textsc{Enforce} constraints). Then the domain constraints $\phi$ are a concatenation of the \textsc{Forbid} and \textsc{Enforce} sets of constraints.
For the studied MI attacks, we describe the variables of interest $V$ in Section~\ref{sec:generalization} and our domain constraints $\phi$ in Section~\ref{sec:dk-constraints} in detail.
In addition to the inputs necessary to infer the causal model, \tool allows the user to pose well-reasoned queries about potential factors of MI attacks, as per the query interface.

\begin{figure}[t]
\centering
\begin{subfigure}[t]{0.5\linewidth}
    \centering
    \includegraphics[width=0.6\linewidth]{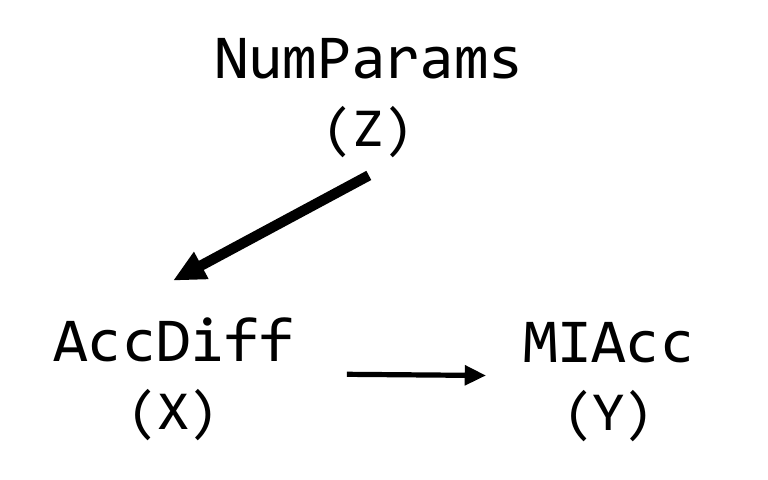}
    \subcaption[]{}
    \label{fig:makes-sense}
\end{subfigure}%
\begin{subfigure}[t]{0.5\linewidth}
    \centering
    \includegraphics[width=0.6\linewidth]{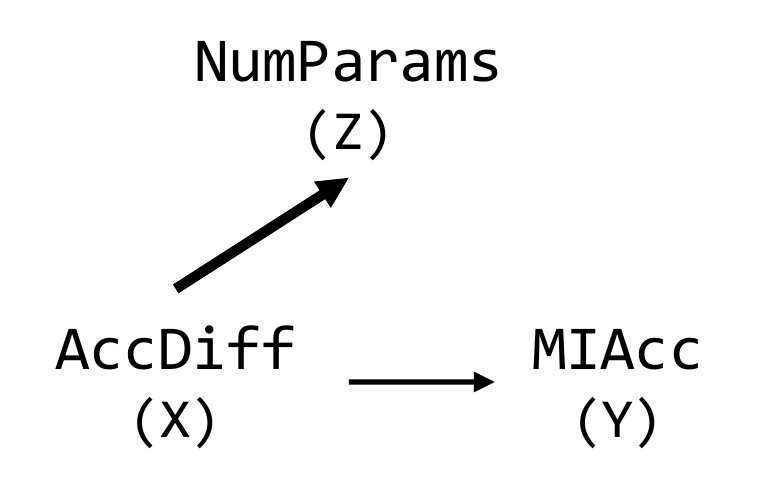}
    \subcaption[]{}
    \label{fig:non-sense}
\end{subfigure}
\caption{Two causal models that are not indentifiable (distinguishable) from observations, since both result in the same conditional (in)dependence relations, but require different quantities to estimate in a causal analysis.}

\label{fig:motivate-dk}
\end{figure}

\subsection{Learning the Causal Model}

Despite the clear advantage of explicitly expressing assumptions in the form of an interpretable causal graph, constructing one is challenging. 
The fundamental issue is that while associations or correlational analysis are useful for predicting outcome, they do not always reflect the causal relationship.
Associations can at most reveal relationships of dependence or (conditional) independence.

To illustrate this point, we show two models that describe the same conditional independence relationships in Fig.~\ref{fig:motivate-dk}, but are causally different. 
In Fig.~\ref{fig:makes-sense}, the model encodes that the model complexity affects the \traintestgap which in turn influences the MI attack accuracy. In contrast, the model in Fig.~\ref{fig:non-sense} describes that the accuracy difference affects both the model complexity and the MI attack accuracy. 
The models in Fig.~\ref{fig:makes-sense} and~\ref{fig:non-sense}, though, are indistinguishable from one another purely from observations, they both encode that $\numparams \independent \miacc | \accdiff$. But, in Fig.~\ref{fig:non-sense}, the model complexity has no causal effect on the MI attack, whereas in Fig.~\ref{fig:makes-sense} the model complexity causes the MI attack to change through the \traintestgap. 
In fact, from how the experiment is set up, the second relationship does not have any real-world interpretation, i.e., the model complexity is decided beforehand as a hyper-parameter to the training process.
Thus, our approach must rely on domain knowledge, a specification of which is missing in prior works in the literature.

Formally, causal models $(G,\theta)$ consist of (a) a DAG $G=(V,E)$ called a causal graph, over a set $V$ of vertices and (b) a joint probability distribution $\mathcal{P}_\theta(V)$, parameterized by $\theta$ over the variables in $V$\footnote{To simplify notation, we denote the vertex and its corresponding variable the same.}.
The set of variables $V$ can take either discrete or continuous values.
Our framework is orthogonal to the underlying representation of parameters. We choose a linear model to represent the relationship between the nodes of the graph.
For predictive queries, the parameters have a probabilistic interpretation $\mathcal{P}_\theta(V)=\Pi_i\prob{X_i|pa_{X_i}}, X_i\in V$. Each node has associated with it a probability function based on its parent nodes $pa_{X_i}$. In this work, we utilize the linear Gaussian model: $X_i$ is a linear Gaussian of its parents $X_j$: $X_i=\beta_0 + \sum_j \beta_j X_j + \epsilon$ where $X_j \in pa_{X_i}$ and $\epsilon \sim \mathcal{N}(0,\sigma^2)$. %

Note that our choice of linear equations and Gaussian probability functions are not fundamental---these can be changed if necessary. These choices have been sufficient to create causal models with good predictive power for the attacks we analyze (see Section~\ref{sec:pred-power}).

To learn the causal graph, there are two sub-steps: (a) learning the structure of the graph $G=(V,E)$ from the traces $T$ and constraints $\phi$ and (b) learning the parameters of the causal graph.
Conceptually, the first sub-goal is to maximize the posterior probability $\prob{G|T}=\prob{G}\prob{T|G}$, where $\prob{G}$ is a prior on the graph (i.e., $G$ contains the edges represented by the \textsc{Enforce} list and all graphs with edges that are part of the \textsc{Forbid} have $0$ probability) and $\prob{T|G}$ is the predictive probability of the graph $G$. 
Ideally, the posterior probability concentrates around a single structure $G_{MAP}$, the optimal directed acyclical graph.
Learning the optimal DAG though is intractable for most problems as the number of DAGs is super exponential with the number of nodes $O(n!2^{\binom{n}{2}})$~\cite{robinson1977counting}.
In fact, recovering the optimal DAG with a bounded in-degree $\geq 2$ has been shown to be NP-hard~\cite{chickering1995learning}.

We choose to instantiate our approach with a standard hill-climbing algorithm~\cite{wu2008dynamic,daly2007methods}, an iterative Greedy approach that starts from the graph with nodes representing the variables $V$ and the edges that are part of the $\textsc{Enforce}$.
The algorithm does not guarantee that the produced graph is the optimal one but it is scalable. Since our goal is to disambiguate between many different possible factors (see Section~\ref{sec:motivation}), this technique allows the user to add new variables of interest and has a good predictive accuracy (goodness of fit) in practice. 
The algorithm iteratively tries to add, remove, or reverse the direction of a directed edge from the graph at the previous step. It uses a scoring function to choose between these operations. The scoring function maps a graph to a numeric value. We use a type of score based on log-likelihood $LL(G|T)$ but that prefers simpler graphs ($LL(G|T)-p$, where $p$ is a penalizing term that grows with more complex causal models with more parameters).
This is known as Bayesian Information Criterion~\cite{schwarz1978estimating}.
For each such operation, the hill-climbing algorithm computes the change in the score if that operation had been performed. It then picks the operation that results in the best score and stops when no further improvements are possible.
Moreover, several distinct graphs $G$ can have similarly high posterior probabilities which is common when the data size is small compared to the domain size~\cite{friedman2003being}.
This is in part due to the the causal ambiguity of learning from data.

Instead of learning a single graph, \tool uses a bootstrapping technique~\cite{friedman2013data}.
The bootstrapping process resamples the traces $T$ with replacement. It then returns a set $\mathcal{S}$ of multiple bootstrap datasets $S$. 
For each bootstrap dataset, \tool uses the graph learning algorithm to learn the structure of the graph $G'$.
For every arc present in the set of graphs $AG$, \tool estimates the strength or confidence that each possible edge $e_i$ is present in the true DAG as
$\hat{p}_{e_i}=\frac{1}{|\mathcal{S}|}\sum_{b\in \mathcal{S}} \mathbbm{1}_{\{e_i \in E_b\}}$, where $\mathbbm{1}_{\{e_i \in E_b\}}$ returns $1$ if $e_i \in E_b$, else returns $0$. 
The purpose is to prune out the edges that are below a certain confidence threshold $t$. There are existing techniques to estimate the confidence threshold such as the $L_1$ estimator~\cite{scutari2011identifying} which \tool uses to fix the confidence threshold.
Using the most significant arcs, \tool constructs a graph that contains all of the significant arcs (the averaged graph $G$).
Our approach does not guarantee that the obtained graph represents the true causal graph--inferring one in our setup is infeasible.
Thus, we take the practical approach and aim to infer a graph with a good predictive power.

\subsection{Answering Queries}
\label{sec:causal-inference}

Predictive queries ask what is the output of the model given that certain input factors have certain values.
Given a set of previously unseen set of assignments for the variables $\{X_i=a_i\}, X_i\in V$, the outcome corresponding to the MI attack node $Y\in V$ is computed by the expression $\Ex{Y|pa_Y}$. This expression is recursively expanded until it is conditioned on the $X_i$ values and can be evaluated with the given concrete values.
To learn the coefficients associated with each node, we use a standard maximum likelihood estimation approach to fit each node's observed data conditioned on its parents.

In principle, we can answer interventional queries, which measure how much changes in an factor's value affects the outcome, by conducting experiments where we manipulate the training process such that input variables take desired values. Such manipulations are called interventions. Formally, given a set of variables $V=\{X_1, \ldots, X_n\}$, an intervention on a set $W \subset V$ of the variables is an experiment where the experimenter controls each variable $w \in W$ to take a value of another independent (from other variables) variable $u$, i.e., $w=u$.
This operation, and how it affects the joint distribution, has been formalized as the {\em do} operator by Pearl~\cite{pearl2009causality}.
For example, in Fig.~\ref{fig:makes-sense}, we can intervene on the model complexity independently of the other variables.
However, in some cases modifying variables directly is not feasible in practice (e.g., the \traintestgap) as it requires knowledge of the data distribution that the model is trying to learn in the first place. So, we cannot really conduct such interventional experiments.

The key insight to answer intervention queries is that we can reason about such queries with only the causal graph and the data--
\tool applies the principles of $do$-calculus to analytically compute the causal relationship expressed by the $do$-query. 
The $do$-calculus rules have been proven to be sound and complete~\cite{shpitser2006identification,huang2006pearl}. They are complete in that if repeated application of the rules of $do$-calculus cannot obtain a conditional probability, then the algorithm outputs that the causal relationship cannot be identified without additional assumptions. 
If we do obtain an ordinary conditional probability, then we say that the causal estimate can be identified, i.e., the graph has enough assumptions or no ambiguity.
Then, the obtained expression (called the {\em estimand}) represents the correct translation of the causal query to a conditional probability (soundness).
Such guarantees are powerful tools: Given the formal query and the causal model, this approach avoids paradoxes that might arise from over-controlling or not controlling (Section~\ref{sec:motivation}).
We will explain a small fragment of this calculus through an example.

\begin{figure}[t]
\centering
\begin{subfigure}[t]{0.45\linewidth}
    \centering
    \includegraphics[width=\linewidth]{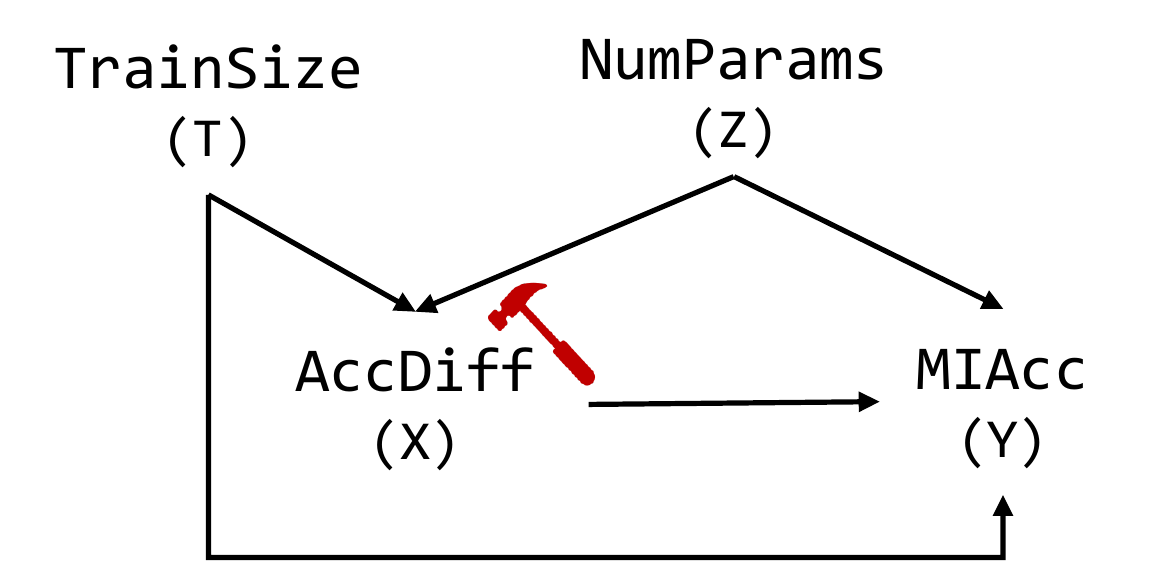}
    \subcaption[]{}
    \label{fig:backdoor-simple}
\end{subfigure}%
\begin{subfigure}[t]{0.55\linewidth}
    \centering
    \includegraphics[width=\linewidth]{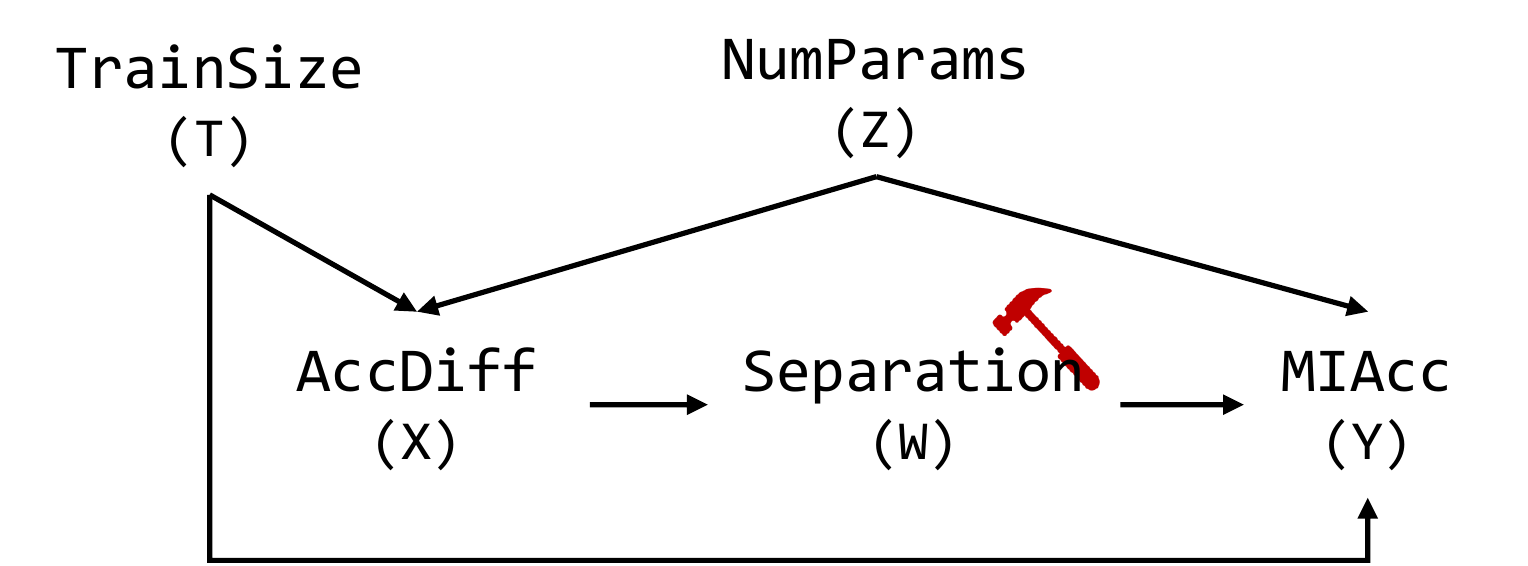}
    \subcaption[]{}
    \label{fig:backdoor-collider}
\end{subfigure}
\caption{Importance of selecting the right control variables to avoid selection bias. $T$ and $Z$ are both in adjustment set for $X$ in \ref{fig:backdoor-simple}. In Fig.~\ref{fig:backdoor-collider}, to estimate the effect of separation score ($W$), we can control for $T$, $Z$ or $X$ but controlling for the \traintestgap ($X$) introduces selection bias.
}
\label{fig:backdoor}
\end{figure}

\noindent\textbf{Example: Backdoor Paths.}
Let us consider the examples in Fig.~\ref{fig:backdoor}. In Fig.~\ref{fig:backdoor-simple}, the query (Q1) is to estimate the effect of \traintestgap on the MI attack accuracy, given only observations of the variables of interest. In Fig.~\ref{fig:backdoor-collider}, we introduce a new variable, the separation score between members and non-members (H8), which is caused by the \traintestgap. The query (Q2) asks to estimate the effect of the separation distance on the attack accuracy. 
The causal model in Fig.~\ref{fig:backdoor-simple} was previously discussed in Section~\ref{sec:motivation}---the expected attack accuracy is computed over both of the confounding factors, the model complexity, and the training set size. From the graph structure, observe that the confounding happens because of the two undirected paths from the node corresponding to the \traintestgap to the MI attack node ($\pi_1: \accdiff \leftarrow \numparams \rightarrow \miacc$ and $\pi_2: \accdiff \leftarrow \trainsize \rightarrow \miacc$).
Such paths are called {\em backdoor paths}.
A backdoor path is a non-causal path from $X$ to $Y$. This is a path that would exist in the graph even if we were to remove the outgoing edges from the node of interest.
When there are backdoor paths, there are sources of association (which we can observe statistically) in addition to the causal ones.
In Fig.~\ref{fig:backdoor-collider}, it seems the query (Q2) requires a similar control as in Fig.~\ref{fig:backdoor-simple}. However, which of the nodes on the paths $\pi_1$ and $\pi_2$ paths, should we control for? The node corresponding to the \traintestgap is one such candidate. Notice though that unlike other nodes on the backdoor paths, it has two incoming edges, meaning that controlling for it biases the observed relationship between its parents. Such nodes are called ``colliders'' and can introduce bias (see~Section~\ref{sec:motivation}). One has to carefully determine exactly when to control for colliders. 

There is a set of principled rules to ``block'' backdoor paths. We summarize these rules informally here but interested readers can refer to~\cite{pearl2009causality,pearl2012class} for more thorough background.
A path is blocked if 1) we control for a non-collider on that path or 2) we do not control for a collider on that path. For any given path, only one of these conditions is required to block the path. So, if there exists a path between $X$ and $Y$ that contains an uncontrolled collider, that path is blocked without controlling on any other variables.
Guided by these rules (called the {\em backdoor criteria}), in (Q2) we should control for X when estimating the effect of $W$ on the MI attack.

\paragraph{Estimating the Causal Effect}
Recall that we are interested in the average treatment effect, which is the average difference in the outcomes given that the treatment takes two values: the \textit{treatment} value and the \textit{control} value.
A straight-forward way to compute the average treatment effect (ATE) is by using the difference in the mean of the outcome {\em conditioned} on the treatment variable ($\Ex{Y|X=a}-\Ex{Y|X=b}$). 
However, this method of computation suffers from statistical pitfalls, such as sampling bias and confounding bias which we highlighted in Section~\ref{sec:motivation}.
Instead, what we want to quantify is the average treatment effect as a $do$-query as defined below.

\begin{definition}[Average Treatment Effect]\label{def:ate}
The average treatment effect of a variable of interest $X$ (called the treatment) on the target variable $Y$ (called the outcome) is:
\begin{align*}
 \ATE{X, Y, a, b} = \Ex{Y|do(X = a)} - \Ex{Y|do(X = b)},
\end{align*}
\end{definition}
where $a,b$ are constants for which $X$ is defined. We omit the constants when the query is over the  domain of $X$.

\tool will translate the $do$-query $\Ex{Y|do(X=a)}$ into an ordinary (conditional) expectation expression from given the causal model (e.g., using the backdoor criteria).
It then learns an estimator that allows computing the ordinary expectation using the available data. We choose a linear regression model to estimate the quantities of interest.
Its estimates are interpretable: a positive $\ATE{X,Y}$ value means that an increase of the feature $X$ causes an increase in the MI attack accuracy $Y$, and vice-versa for negative $\ATE{X,Y}$.

In summary, we have described a methodology encapsulated in \tool to analyze causally potential factors. While our methodology uses techniques standard in causality, we have carefully laid out the technical choices that allow us to achieve our goals: 1) used linear equations to capture causal effects; 2) combined Greedy structured algorithm with bootstrapping to scale the creation of models; and 3) defined the average treatment effect as our measured outcome.

\section{Connecting MI and Generalization}
\label{sec:approach}

Our main technical novelty is how use \tool to study the connection between MI attacks and classical generalization in ML. We now show how to create causal models for $6$ different attacks and {\em formalize} hypotheses (H1)-(H9) made in prior works.

\subsection{Variables of Interest}
\label{sec:generalization}

The generalization notions and other potential causes identified in H1-H9 (Section~\ref{sec:motivation}) are properties of the training algorithm.
The training algorithm $\mathcal{A}$ takes as input a training dataset $\dtrain$ consisting of $N$ samples $\dtrain=\{(x_1, y_1), \ldots, (x_N, y_N)\}$, $D\sim P^N$, each independently identically drawn from $P$ where $P$ is a distribution over $\mathcal{X}\times \mathcal{Y}$, $\mathcal{X}$ is the input space and $\mathcal{Y}$ is the output space. The training algorithm also takes as input a set of training hyperparameters $\trainparam$ and the loss function $l$.
The training algorithm produces a model $f:\mathcal{X}\rightarrow\mathcal{Y}$.
All of the prior works have studied MI attacks on neural networks trained with stochastic gradient descent (SGD), so we focus on SGD primarily in this work.

The generalization error is a measure of how well a learned model $f$ can correctly predict previously unseen data samples. 
For a given a model $f$ and a sample $z=\{x,y\} \sim P$, the generalization error is defined as $\mathbb{E}_{z\sim P}[l(f,z)]$. The learned model $f$ depends on the drawn training dataset $\dtrain$. As a result, the generalization error of $\learnalg$ is $\testloss=\mathbb{E}_{z\sim P,D}[l(f_D,z)]$. 
We denote the generalization error as \testloss since if we were to sample $z\sim P$ it would be highly unlikely for that sample to belong to the training set $\dtrain$.
\par\smallskip

\noindent\textbf{Bias-variance decomposition.}
A fundamental principle to understand generalization in machine learning models is the bias-variance decomposition~\cite{kohavi1996bias,geman1992neural,yang2020rethinking}, which establishes that the \textbf{generalization error} directly factors into \textbf{Bias} and \textbf{Variance} as shown in Table~\ref{tab:features}.
The bias represents how well the hypothesis class $\mathcal{H}$ to which the model $f$ belongs to fits the true
data $\mathcal{Y}$, while the variance represents how much the model varies across different
samples of data.
For example, with sufficient training time, a model that is overly-parametrized can have a low bias (since it
fits the data very well) and high variance (because it can fit all the
``accidental regularities'' or idiosyncrasies of the sampled data). 
Our causal models use bias and variance as variables and therefore these serve as a new lens to explain how they affect MI attacks in addition to generalization in the same representation.
To compute the bias and variance, we follow the methodology outlined in prior work~\cite{yang2020rethinking}. We first compute an (unbiased) estimator for the variance term in the bias-variance decomposition. Next, from the generalization error (or loss), we subtract the variance to derive the bias. The full mathematical formulation of bias and variance is in Appendix~\ref{sec:appdx-bias-variance}.
\par\smallskip

\noindent\textbf{MI Attack Accuracy.} 
(H1)-(H9) are claims relating the effect of potential causes on the MI attack susceptibility. To measure susceptibility, we consider the random variable corresponding to the MI attack accuracy, for each prior work attack. In total, we study three MI attacks: the multiple shadow model (\oakacc)~\cite{shokri2017membership}, the single shadow model attack (\mlleaktopacc)~\cite{salem2019ml}, and threshold-based attack (\threshacc)~\cite{yeom2018privacy}.
We additionally perform similar attacks to~\cite{salem2019ml} where we use one shadow model (\mlleakacc).
We take the whole prediction vector for CIFAR10, and the top-$10$ predictions for CIFAR100 as input features to train the attack model. Other considered variants of this attack include the correct label in the input features of the attack model. We denote these as \mlleaktoplacc and \mlleaklacc.
For each learned network $f_D$, we evaluate the MI attack on members $\in D$ and a dataset of previously unseen samples, non-members $\not\in D$. The final result is the expected accuracy on members and non-members averaged over multiple samples $f_D$.

\noindent\textbf{Other Model Properties.} Some of the hypotheses in prior work involve training hyperparameters and model properties such as training set size (H4) and model complexity (H3). For H3, we use the number of parameters in the model to measure model complexity. Specifically, we count the number of parameters (e.g., weights and biases) that are updated during the training phase.
\par\smallskip

The potential factors that appear in H1-H9 are summarized in Table~\ref{tab:features}. Our aim is to infer a causal model over these variables. Next, we require traces, so we run the training algorithm to collect observations of these variables. We leave the detailed process to generate traces of these variables for Section~\ref{sec:eval-setup}. Besides traces, we formulate domain knowledge constraints as input to \tool.

\begin{table}[]
\centering
\caption{Summary of variables we consider when building our causal graphs to answer queries Q1-Q9. We build the causal graphs for each MI attack $A$~\cite{shokri2017membership,salem2019ml,yeom2018privacy}.
For a given sample $z$, the MI attack outputs whether it is a member ($m$) or not ($\neg m$).
We illustrate the variance term only for MSE where $\bar{f}(x)=\mathbb{E}_{D}[f(x,D)]$. 
}
\label{tab:features}
\resizebox{0.8\linewidth}{!}{%
\begin{tabular}{c|c}
\textbf{Variables} & \textbf{Formula}   \\ \hline
\trainacc        & $\mathbb{E}_{z \sim D,D}[f_D(x) = y]$         \\ \hline
\testacc         & $\mathbb{E}_{z\sim P,D}[f_D(x) = y]$          \\ \hline
\accdiff         & \trainacc - \testacc                              \\ \hline
\trainloss        & $\mathbb{E}_{z\sim D,D}[l(f_D,z)]$          \\ \hline
\trainvar &
  $\mathbb{E}_{x \sim D,D}[\norm{f_D(x)-\bar{f}(x)}^2]$  \\ \hline
\trainbias       & \trainloss-\trainvar                     \\ \hline
\testloss         & $\mathbb{E}_{z\sim P,D}[l(f_D,z)]$                                                                                       \\ \hline
\testvar &
  $\mathbb{E}_{x \sim P,D}[\norm{f_D(x)-\bar{f}(x)}^2]$  \\ \hline
\testbias       & \testloss - \testvar                    \\ \hline
\lossdiff        & \testloss - \trainloss                  \\ \hline
\numparams  & $|f_D|$                                  \\ \hline
\trainsize        & $\in \{1k, 5k\}$      \\ \hline
\oakacc &
   $\mathbb{E}_{z,D}[A(z) = m|z \sim D \land A(z) =  \neg m|z \sim P]$  \\ \hline
\begin{tabular}[c]{@{}c@{}} \mlleakacc,\\ \mlleaktopacc \end{tabular} &
  $\mathbb{E}_{z,D}[A(z) = m|z \sim D \land A(z) = \neg m|z \sim P]$  \\ \hline
\multicolumn{1}{c|}{\threshacc} &
  $\mathbb{E}_{z,D}[A(z) = m|z \sim D \land A(z) = \neg m|z \sim P]$  \\ \hline
\multicolumn{1}{c|}{\centroid} & $\mathbb{E}_{D}[\norm{C(D) - C(P)}]$
\end{tabular}%
}
\end{table}

\subsection{Domain Knowledge as Constraints}
\label{sec:dk-constraints}

As our domain-specific constraints, we leverage simple insights that force the hill-climbing algorithm to infer models that have causal meaning.
For instance, one constraint encodes that the root nodes of the model should correspond to variables that are part of the training algorithm's hyper-parameters such as \trainsize and \numparams. This constraint belongs to the \textsc{Forbid} list.
In addition to these, we have identified the following constraints:

\begin{itemize}
    \item {\em There are no outgoing edges from the attack node.} Without this constraint, the structure learning algorithm could learn that the attack causes one of the features--the direction of the edge cannot be inferred by observations only.
    \item {\em There is no edge from a node that is neither a root node nor \trainvar nor \trainloss to \trainbias.} We add this constraint because the \trainbias is computed from the \trainloss and \trainvar. Any influence on \trainbias should be mediated by its two parents.
    \item {\em There is no edge from a node that is not a root node  to \trainvar.} The variance on training samples is computed directly on the prediction vectors.
    \item {\em There is no edge from a node that is neither a root node, \testvar nor \testloss to \testbias.} We add this constraint because the \testbias is computed from the \testloss and \testvar. Any influence on \testbias should be mediated by its two parents.
    \item {\em There is no edge from a node that is not a root node to \testvar.} The variance on testing samples is computed directly on the prediction vectors.
    \item {\em Constraints in \textsc{Enforce}.} There is an edge from \trainacc to \accdiff and \testacc to \accdiff. There is an edge from \trainloss to \lossdiff and \testloss to \lossdiff.
\end{itemize}

In total, we have \todo{\numconstraints} \textsc{Forbid} and \textsc{Enforce} constraints (see Appendix~\ref{sec:appdx-dk}). Such constraints are easy to derive, as they either stem from the definitions in Table~\ref{tab:features} or from the data-generating process.

\subsection{From Hypotheses to Queries}
\label{sec:causal-queries}

We can obtain a causal model from the traces and the domain knowledge constraints using \tool. The last step is to formulate queries on the causal models. Hypotheses H1-H9 can all be formally described as interventional queries Q1-Q9 respectively, as follows. 

\noindent\textbf{(H1,H6,H9) $\rightarrow$ (Q1,Q6,Q9): Generalization Metrics.} Existing work uses two different metrics to quantify generalization precisely: the \traintestgap (\accdiff in Table~\ref{tab:features}) and the average generalization error (\lossdiff in Table~\ref{tab:features}).
We use the metric that was cited by the respective original works to determine the query for each of the studied attacks.

\noindent\textbf{(H2) $\rightarrow$ (Q2): Formalizing ``Closeness''.}
We first quantify ``closeness'' more formally between a shadow model and a target model when these have the same architecture, as is the case for the MI attack by Shokri et al.~\cite{shokri2017membership}. Our observation is that the variance term in the generalization error is a metric of ``closeness''. Intuitively, since the shadow model is a realization of a different subset $\dtrain_i$ of $\dtrain$, the variance is a measure of the expectation of whether training on different sampled training sets (of the same size) from the data distribution outputs neural networks with very ``different'' prediction vectors. Since this particular MI attack trains shadow models with the same architecture as the target model, the larger the variance, the more likely we are to obtain a shadow model that is on average more ``distant'' from the target model.
Hence, H2 can be reformulated to check if the variance of the learning algorithm is a cause of the attack. We consider the variance of $\learnalg$ on the members (training samples, denoted as \trainvar) and the variance of $\learnalg$ on non-members (testing samples, denoted as \testvar), separately. The causal query asks if \trainvar, and, respectively, \testvar, affect the MI attack accuracy \oakacc, i.e., if a change in \trainvar (or \testvar) causes the \oakacc to change. Formally, the hypothesis translates to two $do$-queries (Q2): $\ATE{\oakacc,\trainvar}$ and $\ATE{\oakacc,\testvar}$.

\noindent\textbf{(H5) $\rightarrow$ (Q5): Single Shadow Model.}
We want to check if the closeness (as measured by \trainvar and \testvar) is a cause for the single shadow model attack. If it does not contribute to the attack, the hypothesis is correct.

\noindent\textbf{(H7) $\rightarrow$ (Q7): Different Causes.}
Prior work showed that in terms of MI attack performance there is no significant difference between the performance of attack models trained on the top three predictions of the prediction output vs. the whole prediction vector~\cite{salem2019ml} (page 5, Fig.4). We denote the attack using the whole prediction vector as \mlleakacc.
Here, we are interested if the attack model's performance changes with the target model for the variants of the attack. Thus, we formalize it as checking whether \testvar and \trainvar are causes for \mlleakacc(-l) and \mlleaktopacc(-l).

\noindent\textbf{(H8) $\rightarrow$ (Q8): Formalizing Decision Boundaries.} A few works give credit to the clear decision boundary between members and non-members for the success of shadow model-based attacks~\cite{salem2019ml}.
To quantify the ``distinguishability'' between members and non-members, we first compute the centroid of members ($C(D)$) and non-members ($C(P)$) as the following:
$C(D) = \mathbb{E}_{z \sim D}[f_D(z)]$ and $C(P) = \mathbb{E}_{z \sim P}[f_D(z)]$.
Then, we use the Euclidean distance between the above two centroids to measure the distinguishability for the given training set.
For each training setup ($\mathcal{A},\trainparam$ and architecture), we compute the averaged centroid distance over multiple different training sets. Note that the user can specify any such existing statistic as a distinguishability measure between the training and testing set as input to \tool.

\paragraph{Implementation.}
Our implementation consists of two parts: generating the traces, i.e., training models and running attacks, and implementing \tool. For the traces, we use the standard machine learning library PyTorch 1.7.1+cu110~\cite{pytorch} to train the models and run the attacks. For \tool, we use two libraries for analyzing the MI attacks. First, we use the R library called bnlearn 4.7~\cite{bnlearn} to infer the causal models. This library offers several off-the-shelf algorithms for structured learning and Bayesian inference. Second, we use dowhy 0.7.0~\cite{dowhy,dowhypaper} to implement the average treatment effect queries.

\section{Evaluation}
\label{sec:eval}

We evaluate \tool on two grounds:
\begin{itemize}
    \item (EQ1) {\em Goodness of fit}: Are the causal models predicting the MI attack more accurately on unseen samples than  correlational analyses?
    \item (EQ2) {\em Utility}: Is \tool useful in refuting or confirming prior hypotheses? Does it provide useful insights to how MI attacks connect to generalization and how defenses work?
\end{itemize}

We study \todo{\numattacks} MI attacks: multiple shadow model~\cite{shokri2017membership}, 4 variations of the single shadow model~\cite{salem2019ml} and threshold-based~\cite{yeom2018privacy}.
We present the results with respect to the $9$ queries for two loss functions, so in total we have $18$ prior work hypotheses. 
In addition, we study \todo{\numdefenses} practical defenses proposed in prior work. The first is L2-regularization (also known as weight decay) that was proposed as a mitigation strategy for the multiple shadow model attack~\cite{shokri2017membership}. It has been used as a baseline for other defenses~\cite{nasr2018machine}.
The second defense we consider is MemGuard~\cite{jia2019memguard}, a defense that changes the prediction vectors without changing the accuracy of the model.
We choose this defense as it is effective against the attacks we also considered in this work~\cite{shokri2017membership,salem2019ml}.
The causal models for all of the evaluated attacks and defenses are available in Apppendix~\ref{sec:appdx-graphs}.

We present details later and summarize our key findings below:
\begin{itemize}
    \item There is no one-size-fits-all explanation for the $6$ MI attacks---the factors contribute differently in different attacks. 
    
    \item Our analysis refutes \todo{\falsepriorq/\totalpriorq} prior hypothesis we formalized, and confirms \todo{\truepriorq/\totalpriorq} as valid. %
    
    \item Our causal models have predictive accuracy of \todo{\predacc} for unobserved experiments, which are not used for causal model creation. This is comparable or better by \todo{\predpower} than simple correlational analysis between the single cause and the MI attack, in all cases we study.
    
    \item Bias and Variance observed during training can quantitatively predict both generalization measures (e.g., \testloss, \testbias, and \testacc) and MI attack performance, providing new insights. These factors play a disproportionately larger role in explaining MI attack performance, compared to other factors such as model complexity, dataset size, or even generalization measures themselves. 
    
    \item Defenses reduce certain causes of the MI attack, but not all and not completely. They reduce the effect of variance, but fail to eliminate factors such as the \traintestgap or the distance between members and non-members. 
\end{itemize}

\subsection{Experimental Setup}
\label{sec:eval-setup}

\noindent\textbf{Datasets.} We select $3$ common image datasets: MNIST, CIFAR10, and CIFAR100.
MNIST has $60$k training and $10$k testing samples of $28\times 28$ grayscale images of handwritten digits. CIFAR10 and CIFAR100 have $50$k training and $10$k testing samples of $32\times32$ color images uniformly distributed in $10$ and $100$ classes, respectively. 

\noindent\textbf{Models.} For each dataset, we train multiple models with different architectures and hyperparameters. For MNIST, we use multilayer perceptron (MLP) with one hidden layer to build the target model. We change the number of units used in the hidden layer ($\{16,32,64,128,256\}$) to change the \textit{width} or the model complexity. 
For CIFAR10 and CIFAR100, we use various convolutional neural network (CNN) architectures: AlexNet, DenseNet161, and ResNet34.  For changing the width, we vary the number of filters of these models. 
The widths considered for AlexNet were $\{16,32,64,128,256\}$, and for DenseNet161 and ResNet34 they were $\{2,4,8,16,32\}$.

\noindent\textbf{Training Algorithms.} We trained all models using stochastic gradient descent (SGD) with momentum 0.9, two weight-decay rates $\{5 \times 10^{-4}, 5\times 10^{-3}\}$ and two kinds of loss functions: mean squared error (MSE) and cross entropy (CE). The higher weight decay models are used to evaluate L2-regularization.
We summarize the training configurations we considered in Appendix, Table~\ref{tab:training-methods}. For the models trained with the scheduler, we used the step learning rate scheduler with the learning rate decay factor of $10$ and for $200$ epochs.

\noindent\textbf{Variables of Interest.} To estimate the variables of interest for each training setup, we follow the procedure proposed by Yang et al. \cite{yang2020rethinking}. Specifically, we randomly generate disjoint splits of training samples $\dtrain = \dtrain_1 \cup \ldots \cup \dtrain_n$, where size of each $|\dtrain_i|= s$. We train $n$ models for each architecture width $\model_{\dtrain_i}$ over different training sets $\dtrain_i$. 
For CIFAR10 and CIFAR100, we use $n\in\{10, 50\}$ and, respectively, $s\in\{5000,1000\}$, while we use $n\in\{12, 60\}$ for MNIST because MNIST has a larger training set.

\noindent\textbf{Attacks.}
The shadow model training size is equal to the training size of the target model, i.e., either $1000$ or $5000$. This set forms the member set for the attack model. Additionally, an equal-sized set is used to form the non-member training set for the attack model. The evaluation set for the attack model consists of the $1000$ or $5000$ training samples of the target model and an equal-sized set not previously seen by either target and shadow models. For each architecture and width, we perform the attack $30$ times for different samples of the datasets from the original training set $D$. The same splits are used for all datasets and single shadow model attacks.

\begin{table}[]
\caption{The graphs that \tool generates have consistently equal or better predictive power than simple correlations.
\textbf{Best Corr.} stands for the Pearson correlation coefficient, and \textbf{\tool Pred.} stands for predictive correlation.}
\label{tab:pred-acc}
\centering
\resizebox{0.49\textwidth}{!}{%
\begin{tabular}{cccccc}
                     &                                 & \multicolumn{4}{c}{\textbf{Normal / High Weight Decay}} \\ \cline{3-6} 
\textbf{Loss} &
  \textbf{Attack} &
  \textbf{\begin{tabular}[c]{@{}c@{}}Best Corr.\\ Variable\end{tabular}} &
  \textbf{\begin{tabular}[c]{@{}c@{}}Best Corr.\\ Value\end{tabular}} &
  \textbf{\begin{tabular}[c]{@{}c@{}}\tool\\ Pred.\end{tabular}} &
  \textbf{\begin{tabular}[c]{@{}c@{}}\tool\\ MSE\end{tabular}} \\ \hline
\multirow{2}{*}{ce}  & \multirow{2}{*}{\mlleakacc}     & \multirow{2}{*}{\accdiff}      & 0.8543     & 0.9358     & 2.31E-03    \\
                     &                                 &                                & 0.8621	 & 0.9083   &  2.54E-03    \\
\multirow{2}{*}{mse} & \multirow{2}{*}{\mlleakacc}     & \multirow{2}{*}{\accdiff}      & 0.8493	 & 0.9170	  &  1.80E-03    \\
                     &                                 &                                & 0.8675	 & 0.9033	  & 9.55E-05    \\
\multirow{2}{*}{ce}  & \multirow{2}{*}{\mlleaklacc}    & \multirow{2}{*}{\accdiff}      & 0.8209	 & 0.9685	  & 1.17E-03    \\
                     &                                 &                                & 0.8414	 & 0.9486	  & 2.05E-03    \\
\multirow{2}{*}{mse} & \multirow{2}{*}{\mlleaklacc}    & \multirow{2}{*}{\accdiff}      & 0.9397	 & 0.9740  &  7.30E-04    \\
                     &                                 &                                & 0.9371	 & 0.9694	& 6.72E-05    \\
\multirow{2}{*}{ce}  & \multirow{2}{*}{\mlleaktopacc}  & \centroid                      & 0.9257 & 	    0.9663 &	1.16E-03   \\
                     &                                 & \accdiff                       & 0.8595 &	    0.9645 & 1.16E-03    \\
\multirow{2}{*}{mse} & \multirow{2}{*}{\mlleaktopacc}  & \accdiff                      & 0.8743 &	    0.9466 &	1.26E-03    \\
                     &                                 & \accdiff                       & 0.9041 &	    0.9216 &	1.14E-04    \\
\multirow{2}{*}{ce}  & \multirow{2}{*}{\mlleaktoplacc} & \centroid                      & 0.9166 &	    0.9641 &	2.21E-03    \\
                     &                                 & \accdiff                       & 0.8409 &	    0.9579 &	1.58E-03    \\
\multirow{2}{*}{mse} & \multirow{2}{*}{\mlleaktoplacc} & \accdiff                      & 0.8447 &	    0.9244 &	1.73E-03    \\
                     &                                 & \accdiff                       & 0.9000 &	    0.9361 &	9.51E-05    \\
\multirow{2}{*}{ce}  & \multirow{2}{*}{\oakacc}        & \multirow{2}{*}{\accdiff}      & 0.9752 &    0.9817 &	6.45E-04    \\
                     &                                 &                                & 0.9694 &	    0.9762 &	9.18E-04    \\
\multirow{2}{*}{mse} & \multirow{2}{*}{\oakacc}        & \multirow{2}{*}{\accdiff}      & 0.9526 &    0.9733 &	5.36E-04    \\
                     &                                 &                                & 0.9626 &    0.9689 &	1.61E-04    \\
\multirow{2}{*}{ce}  & \multirow{2}{*}{\threshacc}     & \multirow{2}{*}{\lossdiff}     & 0.7517 &	    0.9739 &	9.00E-04    \\
                     &                                 &                                & 0.7730 &	    0.9425 &	1.77E-03    \\
\multirow{2}{*}{mse} & \multirow{2}{*}{\threshacc}     & \multirow{2}{*}{\lossdiff}     & 0.9823 &	    0.9906 &	2.43E-04    \\
                     &                                 &                                & 0.9775 &    0.9880 &	2.41E-05   
\end{tabular}%
}
\end{table}

\subsection{Predictive Power of Causal Models}
\label{sec:pred-power}

To evaluate the predictive accuracy of the graphs on unseen observations, we use two metrics regularly used in evaluating Bayesian nets: 1) {\em mean predictive correlation} and 2) {\em mean squared error} (MSE). 
We compute these two metrics using standard cross-validation over multiple runs (\todo{$20$}). For each run, we use a \todo{$80/20$} split of the observations for the train-test sets.
For each run of the cross-validation, the predictive correlation measures the (linear) correlation between the observed and the predicted values for the MI attack node.

\noindent\textbf{Baseline.} 
We use a simple baseline that can predict the accuracy of the attack: we compute the Pearson correlation between the observed values of the MI attack and the observed values of the other variables of interest we identified (Section~\ref{sec:generalization}).
A high Pearson correlation (close to $1$) means that there is a linear relationship perfectly describing the MI attack and the variable. If so, to predict the MI attack accuracy, measuring this one variable and learning the coefficients of the relationship from data is enough.
\par\smallskip

In total, we evaluate \todo{$24$} setups for \todo{\numattacks} attacks for models trained with two loss functions, with and without L2-regularization.
For all of the attacks on both undefended and defended models, the predictive correlation is above \todo{\predacc} (Table~\ref{tab:pred-acc}).
Compared to the correlation baseline, the graphs \tool produces are consistently equal or better for all \todo{$24$} setups. For \todo{\numbetterpred} of the setups, the predictive correlation improves \todo{\predpower}.
For the remaining \todo{\remainingpred}, the causal models are on par with baselines or slightly better, within \todo{$3\%$}.
The mean MSE for predictions is low (\todo{$0.001$}) for all of our evaluated attacks and models.
We find that in the case of \oakacc, \tool does not significantly improve the accuracy, as the correlation values are already higher than \todo{$0.95$}. This confirms what prior works suggest~\cite{yeom2018privacy,leino2020stolen,song2021systematic}: using an attack based on the prediction correctness yields, on average, similar performance to the \oakacc. We observe that the \accdiff is almost in a perfectly linear relationship with the accuracy of the multiple shadow model attack. Similarly, the metric we formalize, the centroid distance between clusters of members and non-members (\centroid) is almost perfectly linear with the single shadow model attack.

\begin{table}[]
\centering
\caption{We translate the prior work hypothesis H1-H9 to $\ATE{\text{Feature},\text{Attack}}$ queries. For (Q2,Q5,Q7) which are made of several ATE queries, we $\checkmark$ only if all ATE queries support the prior hypothesis. $*$ means the p-value $> 0.05$ and we mark such queries with $\circ$.}
\label{tab:final-prior-work}
\resizebox{0.48\textwidth}{!}{%
\begin{tabular}{cccccc}
               &            & \multicolumn{2}{c}{\textbf{CE}}                       & \multicolumn{2}{c}{\textbf{MSE}}                               \\ \cline{3-6} 
\textbf{Attack} &
  \textbf{Feature} &
  \textbf{ATE} &
  \textbf{\begin{tabular}[c]{@{}c@{}}Query\\ Result\end{tabular}} &
  \textbf{ATE} &
  \textbf{\begin{tabular}[c]{@{}c@{}}Query\\ Result\end{tabular}} \\ \hline
\oakacc        & \accdiff   & 0.30            & Q1: $\checkmark$                  & 0                          & Q1: $\times$                      \\ \hline
\oakacc        & \trainvar  & 0.02            & \multirow{2}{*}{Q2: $\checkmark$} & 0.03                     & \multirow{2}{*}{Q2: $\circ$} \\
\oakacc        & \testvar   & 0.94            &                                   & 0.20 ($*$)                     &                                   \\ \hline
\oakacc        & \numparams & 0.15            & Q3: $\checkmark$                  & -0.005                          & Q3: $\checkmark$                      \\ \hline
\oakacc        & \trainsize & -0.11           & Q4: $\checkmark$                  & -0.09                   & Q4: $\checkmark$                  \\ \hline
\mlleakacc     & \trainvar  & -0.34  & \multirow{6}{*}{Q7: $\times$}     & -0.05 ($*$)                  & \multirow{6}{*}{Q7: $\times$}     \\
\mlleakacc     & \testvar   & 0.81            &                                   & 0           &                                   \\
\mlleaklacc    & \trainvar  & -0.24  &                                   & 0.06                          &                                   \\
\mlleaklacc    & \testvar   & 0.84           &                                   & 0                     &                                   \\
\mlleaktoplacc & \trainvar  & -0.40           &                                   & -0.06 ($*$)                        &                                   \\
\mlleaktoplacc & \testvar   & 0.75            &                                   & 0                     &                                   \\ \hline
\mlleaktopacc  & \accdiff   & 0.18            & Q6: $\checkmark$                  & 0                          & Q6: $\times$                      \\ \hline
\mlleaktopacc  & \trainvar  & -0.34           & \multirow{2}{*}{Q5: $\times$}     & -0.15 ($*$)                         & \multirow{2}{*}{Q5: $\times$}     \\
\mlleaktopacc  & \testvar   & 0.78           &                                   &  0.24                     &                                   \\ \hline
\mlleaktopacc  & \centroid  & 0.27            & Q8: $\checkmark$                  & 0                          & Q8: $\times$                      \\ \hline
\threshacc     & \lossdiff  & 1.47 ($*$)            & Q9: $\circ$                       & \multicolumn{1}{l}{-0.67} & Q9: $\checkmark$                 
\end{tabular}%
}
\end{table}

\subsection{Testing of Prior Hypotheses}

We confirm using our analysis that \todo{\truepriorq/\totalpriorq} of the prior work hypotheses are true ($\checkmark$ in Table~\ref{tab:final-prior-work}). We also find \todo{\falsepriorq/\totalpriorq} prior hypotheses do {\em not} identify a cause for the studied MI attack ($\times$ in Table~\ref{tab:final-prior-work}).
As some of the hypotheses involve more than one potential cause or they are comparing causes between attacks, we have a total of \todo{$32$} ATE values, \todo{$16$} for each loss function.

We refute prior hypotheses in broadly two instances. First, when there is no universal explanation: prior hypotheses often overlook the differences between NNs trained with different loss functions and the specifics of the attack. 
Second, because prior hypotheses do not consider the connections between the parameters of the training process and variables that naturally appear as part of SGD such as loss difference and variance, etc. 
In summary: 
\begin{itemize}
    \item A single causal factor does not explain all attacks. In fact, causes vary per attack and differ by the loss function used.
    \item (Q1, Q6) The \traintestgap does cause the MI attack accuracy, though for MSE-trained models, the loss difference is a more suitable metric.
    \item (Q2) The ``closeness'' of the shadow model influences the MI attack accuracy, more so for CE-trained models than for MSE-trained models. 
    \item (Q7) There are differences between the variants of the single shadow model attack, and the single shadow model with top-3 is more robust to changes in the shadow model.
    \item (Q3, Q4) Training size is a factor that affects the MI attack accuracy for all of the evaluated attacks. Model complexity is a cause for all evaluated attacks.
    \item (Q5) We find that the variance of the outputs of the models is also a cause for the single shadow model attack, to various degrees depending on the type of attack. Prior work overlooks the differences in the prediction vectors between the target and shadow model.
    \item (Q8) Our formalized distance between the clusters of members and non-members is one of the largest causes for the single shadow model with top-3.
    \item (Q9) The threshold-based attack is influenced with varying degrees by other factors that are related to the loss.
\end{itemize}

\noindent\textbf{CE vs. MSE.} We find that the \traintestgap has the largest influence for CE-trained models, whereas for MSE-trained models it is the loss differences between members and non-members. Similarly, factors such as the variance and the centroid distance that affect MI attack accuracy on CE-trained models are not factors on MSE-trained models.

\noindent\textbf{Detailed Analysis.}
The differences in the prediction vectors of the non-members (as measured by \testvar) has a large causal effect on the average MI attack accuracy. This validates the prior work hypothesis (Q2) that the differences in the shadow and target model affect the MI attack. 
For instance, for CE, the estimated ATE of \trainvar on \oakacc is \todo{$0.02$}, whereas that of \testvar is \todo{$0.94$} (Table~\ref{tab:final-prior-work}).  
We find that for the multiple shadow model attack on MSE-trained models, the number of parameters (Q3) does not show a significant influence on the \oakacc, but for all other attacks and loss functions, the more parameters, the higher the attack accuracy. We thus validate (Q3) as for all evaluated attacks there is a non-zero ATE value.
We find that the variance of the outputs of the models is also a cause for the single shadow model attack, to various degrees depending on the type of attack.
Prior work overlooks the differences in the prediction vectors between the target and shadow model. Thus, our analysis shows that (Q5) is refuted.
The variance of the training algorithm influences MI attacks that take the whole prediction vector more--models tend to agree more on top-3 predictions rather than the whole prediction vector. Thus, the evaluated shadow model MI attacks that take the top-3 predictions are not as sensitive to differences between the shadow and target models' architecture and dataset.
This shows that there are differences between the variants of the single shadow model attack, refuting (Q7).
More details for each attack are available in Appendix~\ref{sec:appdx-experiments}.

We find that a small fraction of queries have low statistical significance (p-value $> 0.05$)---\todo{\msebigpvalue/\totalmsequeries} MSE-trained models and \todo{\cebigpvalue/\totalcequeries} for CE-trained models.
We do not draw any conclusions for these.

\subsection{MI attacks and Generalization}
\label{sec:mi-and-gen}

We find that \textbf{Bias} and \textbf{Variance} values have a high level of influence on both generalization measures as well as the MI attack accuracy. As expected from the bias-variance decomposition theorem, Bias and Variance values are strongly predictive of \testacc, \accdiff, and \centroid values---all of these are generalization measures. Bias and Variance also have a disproportionately high influence on MI attack accuracy. Appendix~\ref{sec:appdx-experiments} gives details; here we summarize their effect on MI attack accuracy which varies by attack.

\noindent\textbf{Variance \& MI.}
The higher the variance on non-members, the higher the MI attack accuracy.
The reverse is true for members: the higher the variance on members, the lower the MI attack accuracy. 
There is less variance on the training samples (i.e., the model learns similar prediction vectors across multiple training datasets) and there is higher variance on the test samples. Our analysis show that the larger the gap between these two, the better the MI attack accuracy.
This suggests that defenses which decrease the gap between the train and test variance (like MemGuard~\cite{jia2019memguard}) will be effective.

\noindent\textbf{Bias \& MI.}
The Bias on non-members is almost always a factor in all types of MI attacks we study. It affects MI attack accuracy through the test accuracy, which in turn affects the \traintestgap or the distance between members and non-members. Recall that Bias is ``how far'' the test set predictions are from the ground truth on average. High Bias on non-members explains why MI attacks, even when not explicitly using the label information, will have a better accuracy. On members, however, the ATE value of the Bias in most cases is close to $0$, i.e., it has almost no effect on MI attack accuracy---this corresponds to networks closely fitting the training set.
Compared to Variance, the Bias has a larger effect, and even more so when the input features to the attack model are the top-3 predicted labels and not the whole prediction vector. 

{\em Why do larger models leak information even if their test loss decreases?}
Complex interplay of loss, variance, bias and the model size have been observed previously under different regimes (Fig.1 in~\cite{yang2020rethinking}).
When the Bias dominates, it and the testing loss decrease with increase in model size of the network, but the Variance does not linearly go down and exhibits a peak (bell-shaped curve). Our analysis shows that Variance {\em by itself} contributes to the MI attack---despite training larger models with lower loss and Bias, the Variance can improve MI attack accuracy (see Variance ATE, Appendix~\ref{sec:appdx-experiments}).

{\em How does CE loss differ from MSE?}
Unlike MSE, for CE we observe that the Variance dominates the loss term. The loss and the Variance exhibit a high peak as the model size increases, while the Bias keeps decreasing (unimodal curve).  This explains why for CE models the Variance has a much larger impact on MI attack accuracy. After Variance peaks, as the model size increases, the loss drops to where both Bias and Variance are low. This explains why increasing model size can reduce MI attack accuracy beyond a point.

\subsection{Utility in Explaining Defenses}

We analyze the causal models for all \todo{\numattacks} attacks for L2-regularization. For MemGuard, we evaluate the single shadow model (top-3) attack on the defended models as done in the original work. Thus we have $2+12=14$ causal graphs in total.
For each case, we analyze how much the ATE for a cause differs from ATE in the corresponding causal graphs for {\em undefended} ML model.  

We find that L2-regularization alleviates some causes but not all, and not in equal measure for all attacks. For instance, it majorly reduces the ATE of the test variance on all of the evaluated MI attacks--even as drastically as from \todo{$0.84$} to \todo{$0.16$} for \mlleaklacc. The effect of the \traintestgap on the \oakacc remains the same, but it increases for the single shadow model attacks. The ATE of the \centroid does not change after regularization is applied, showing that there are still exploitable signals left.

The MemGuard defense reduces the Variance significantly but many factors remain unaddressed even after the defense. For instance, the distance between members remains a factor.
The ATE of the \accdiff is reduced from \todo{$0.19$} to \todo{$0.08$}. MemGuard is more effective overall in removing causes than regularization. 

More details of our analysis of defenses is in Appendix~\ref{sec:appdx-defenses}.

\section{Related Work}

\noindent\textbf{Generalization.}
Generalization in machine learning is a fundamental topic. Several studies investigate the bias-variance decomposition in neural networks~\cite{neal2018modern,yang2020rethinking,geman1992neural}.
Yang et al.~\cite{yang2020rethinking} explore the dependence of bias and variance to network width and depth, e.g., deeper models tend to have lower bias but higher variance. Our work connects MI attacks to such training and architectural choices.
Other works propose new measures of generalization~\cite{dziugaite2020search,jiang2018predicting}.

\noindent\textbf{Membership Inference Attacks.}
There has been a recent line of work proposing MI attacks and providing useful attack taxonomy. Shokri et al.~\cite{shokri2017membership} present the first membership inference attack. They show that overfitting is correlated with their attack performance. They suggest that besides overfitting, the structure and type of the model also contribute to the privacy leakage through membership inference attacks. Several new attacks have emerged~\cite{liu2021ml,murakonda2020ml,he2021node,mireshghallah2022quantifying,li2021membershipleakage,liu2021encodermi,zhang2021membership,choquette2021label,leino2020stolen} and attack taxonomies~\cite{truex2019demystifying,li2021membership} have started to categorize them. These attacks serve as tools to evaluate the privacy risk of machine learning models through attack procedures.
Our work distinguishes itself from all of these by providing a causal framework to explain why these attacks arise.
Notably, our work provides a new lens into how generalization and MI attacks connect---through a systematic measurement and reasoning of bias, variance, and other stochastic variables that arise in training.

Several works have provided mechanistic explanations connecting MI attacks to generalization prior to our work. 
Yeom et al.~\cite{yeom2018privacy} provide a theoretical connection between a notion of generalization called the average generalization error and a bounded-loss adversary which does not apply to training using a CE loss. They also propose a threshold-based attack which has knowledge of the loss distribution which we have also evaluated in our framework \tool. 
The attack assumes that the loss is normally distributed, and thus can be connected to the adversary advantage in a closed-form expression.
Our work shows that the assumptions made in their work may not always hold. We show that MI attack performance is linked to the average generalization error for models with MSE loss but does not always for CE loss. Subsequent work by Song et al.~\cite{song2019privacy} propose a similar threshold-based attack, but on the confidences of the prediction. Nasr et al.~\cite{nasr2019comprehensive} also analyze the connection between membership inference attacks and overfitting, while proposing white-box membership inference attacks. They also empirically observe the correlation of the attack performance to the model capacity.
Song et al.~\cite{song2019membership} evaluate membership inference attacks against adversarially robust models and point out that these models have a larger \traintestgap when considering adversarial examples.
Our work shows that other factors beyond the \traintestgap contribute to the privacy leakage.

\noindent\textbf{Causality.}
Causality is an active area of research with recent advances improving learning of causal models ~\cite{kocaoglu2019characterization,shanmugam2015learning}, as well as better inference procedures~\cite{jaber2018causal,lee2021causal}.
While extensively applied in sciences~\cite{triantafillou2017predicting,glymour2019review,heckman2008econometric,marini1988causality,brady2008causation}, causality has only been recently connected to privacy~\cite{tschantz2020sok,tople2020alleviating}.
In our work, we introduce the causal lens to understand MI and generalization.
Since our proposed methodology is synergistic, combining learning with domain knowledge, we can benefit from such advances to improve our causal models and analysis. 
In addition to learning and inference, methods to test the causal assumptions have also been proposed such as sensitivity analysis~\cite{rosenbaum2005sensitivity,robins2000sensitivity} and simulated dataset-approach~\cite{neal2020realcause}.
Again, our approach can leverage such tests for the constructed causal models.

\section{Conclusion}

We have proposed the first use of causal graphs to capture how stochastic factors---such as bias, variance, model size, data set size, loss values, and so on---causally interact to give rise to MI attacks, providing a new connection between these attacks and generalization. We hope this framework helps formally re-analyze statistical conclusions
and pinpoint root causes more accurately.

\section*{Acknowledgements}
We thank the anonymous reviewers for their valuable feedback. We extend a special thanks to Amit Sharma for helping us with the queries about the DoWhy library and using it in our implementation.
We are grateful to Arnab Bhattacharyya and Kuldeep S. Meel for their useful feedback on an earlier draft of this paper.
This research was supported by the Crystal Centre at National University of Singapore and its sponsors, the National Research Foundation Singapore under its NRF Fellowship Programme [NRF-NRFFAI1-2019-0004] and the Ministry of Education Singapore Tier 2 grant MOE-T2EP20121-0011. Teodora Baluta is also supported by the Google PhD Fellowship.

\bibliographystyle{ACM-Reference-Format}
\balance
\bibliography{paper}

\appendix
\begin{appendix}

\section{Bias-Variance}
\label{sec:appdx-bias-variance}

\subsection{Bias-Variance Decomposition}
\label{sec:bv-details}

Generalization is defined as the expected error on all possible samples and all possible datasets, i.e., $\mathbb{E}_{x,y}\mathbb{E}_\dtrain[l(f_D(x), y)]$.

\paragraph{MSE loss.}
We first consider that the neural network was trained using squared error loss. The expected error or generalization error is~\cite{yang2020rethinking,geman1992neural}:

\begin{align*}
  \mathbb{E}_{x,y}&\mathbb{E}_{D}[(y-f_D(x))^2] \\
  &= \mathbb{E}_{x,y}\mathbb{E}_{D}[y^2 - 2yf_D(x) + f_D(x)^2] \\
            &= \mathbb{E}_{x,y}[y^2 - 2y\mathbb{E}_{D}[f(x,D)] +  \mathbb{E}_{D}[f_D(x)^2] \\
            &= \mathbb{E}_{x,y}[y^2 - 2y\bar{f}(x) + \bar{f}(x)^2] + Var[f_D(x)] \\
            &= \mathbb{E}_{x,y}[(y-\bar{f}(x))^2] + \mathbb{E}_{x,y}[Var[f_D(x)]]\\
            &= \mathbb{E}_{x,y}[(y-\bar{f}(x))^2] + \mathbb{E}_{x,y}\mathbb{E}_{D}[(f_D(x)-\bar{f}(x))^2], \\
\end{align*}

where $\bar{f}(x)=\mathbb{E}_{D}[f_D(x)]$ are the averaged predictions over different training sets.
The first term $\mathbb{E}_{x,y}[(y-\bar{f}(x))^2]$ is the bias and the second term $\mathbb{E}_{x,y}{\mathbb{E}_{D}[(f_D(x)-\bar{f}(x))^2}]$ represents the variance.

\paragraph{Cross-entropy Loss.}
We follow prior work's generalized decomposition for the cross-entropy loss~\cite{yang2020rethinking}. Let $\pi_0(x)\in \mathbb{R}^c$ be the one-hot encoding of the ground truth label.
The cross-entropy is $H(\pi, \pi_0)=\sum_{l=1}^c\pi_0[l] \log{\pi[l]}$, where$\pi[l]$ is the $l$-th element of $\pi$.

\begin{align*}
\Ex{H(\pi_0,\pi}{\dtrain} = D_{KL}(\pi_0 || \hat{\pi}) + \Ex{D_{KL}(\hat{\pi}||\pi)}{\dtrain},
\end{align*}
where $\hat{\pi}$ is the average of log-probability after normalization: $\hat\pi[l] \approx exp(\Ex{\log{\pi[l]}})$ for $l = 1,\ldots, c$.

\paragraph{Estimating the Bias and Variance.}
For MSE loss, we use the following unbiased estimator for variance:
\begin{align*}
    \widehat{Var}(x,D) = \frac{1}{n-1} \sum_{j=1}^{n} \Big\lVert f_{D_i}(x) -  \sum_{j=1}^{n} f_{D_j}(x) \Big\rVert 
\end{align*}

The final value of bias and variance, \trainbias and \trainvar, are obtained by taking the average over the members $x \in D$. Similarly, the \testbias and \testvar are averaged over the non-members $x \sim P$.
We repeat this computation with $N=3$ different random disjoint splits and take the average of the estimate to decrease the variance of the estimator. In total, for each model architecture $\mathcal{M}_w$ (e.g., $\mathcal{M}=$Resnet34 with width $w=2$), we train $n\cdot N$ $f_D$ over a given dataset $D$.

\section{Domain Knowledge Constraints}
\label{sec:appdx-dk}

Each of the constraints described informally are explicitly expanded as a pair of nodes (``from'', ``to'') and added in the \textsc{Enforce} and \textsc{Forbid} lists. The resulting number of constraints are thus \todo{$\numconstraints$}.

\begin{itemize}
    \item {There are no edges from the \centroid to \testloss and, respectively, \testacc.} We add this edge because the direction of influence should be the other way around, if such edges are inferred.
    \item {There is no edge from the \testvar to \testloss.} Similarly, these two quantities are computed from the prediction vector for \testvar and, prediction vectors and labels for \testloss, so there is no edge between them.
\end{itemize}

\section{Detailed Experiments}
\label{sec:appdx-experiments}

We provide more detailed explanations of our results in this section.
Our causal models are available in Appendix~\ref{sec:appdx-graphs}.

In Table~\ref{tab:training-methods}, we show the different configurations we trained.

\def\checkmark{\tikz\fill[scale=0.4](0,.35) -- (.25,0) -- (1,.7) -- (.25,.15) -- cycle;}

\begin{table}[]
\caption{Different configurations of models we trained and analyzed.}
\label{tab:training-methods}
\resizebox{0.47\textwidth}{!}{%
\begin{tabular}{|c c c c c|}
\hline
\multirow{2}{*}{\textbf{Dataset}} & \multicolumn{2}{c}{\textbf{CE}} &  \multicolumn{2}{c}{\textbf{MSE}}   \\
                 {} &  With Scheduler & Without Scheduler &  With Scheduler & Without Scheduler  \\ \hline
CIFAR10   & \checkmark   & \checkmark   & \checkmark   & \checkmark    \\ \hline
CIFAR100  & \checkmark   & \checkmark   & -  &   - \\ \hline
MNIST  & \checkmark   & \checkmark   &  \checkmark  &   \checkmark \\ \hline
\end{tabular}%
}
\end{table}

\subsection{Analysis of MI Attacks}

\noindent\textbf{Variance and Bias.}
We link the variance from the bias-variance decomposition with the ``closeness'' of the shadow models' prediction. We find that the variance generally plays a role in the MI attack, as our causal graphs identify a path from it to the MI attack accuracy.
In particular, the variance on unseen samples has generally a larger impact for the multiple shadow model attack and the single shadow model attack with the label feature as input.
\par\smallskip

\noindent\textbf{CE vs. MSE-trained models.}
The MI attacks have different mechanisms not just per attack but also depending on the loss function.
We find that the variances for MSE models do not have a significant impact on the MI attack performance.
We observe that for MSE-trained models, the variances of both the training data (members) and testing data (non-members) are typically smaller than their CE counterpart. This means that there is less variance among models and, thus, for MSE-trained models, shadow models' prediction vectors would have a similar distribution to that of the target model.
This case explains the prior works' intuition that one does not require multiple shadow models---even one shadow model captures the behaviour of the target model closely.
\par\smallskip

\noindent\textbf{Training set size and model complexity.}
For all of the evaluated attacks and loss functions, we find that larger training set size causes a lower MI attack accuracy. We also validate that a larger model complexity causes a better MI attack performance.
The changes in these features are related to generalization, not only the MI attack performance.
Such findings validate the prior work hypotheses (Q3 and Q4 in Table~\ref{tab:final-prior-work}). Our analysis, though, singles out the causal effect of the training size on the MI attack accuracy, when it is independent of the model complexity. 
If we simultaneously changed both of them, we would be able to find a sweet spot of the best MI attack accuracy and how well the model generalizes. While this has been studied in prior work, with our method we can confirm how these two factors independently influence the privacy leakage.
\par\smallskip

\paragraph{Multiple Shadow Model Attack.}
For CE-trained models, a larger overfitting gap (\accdiff) causes the MI attack accuracy to increase (\oakacc), even when controlling for bias. This validates Q1 from prior work (Table~\ref{tab:ce-undefended-prior-work}).
The differences in the behaviour of the model, e.g., its unique distribution of the prediction vector, influences the attack performance.
We find that the variance of the prediction vectors for the training set (members) causes the accuracy of the multiple shadow model attack to increase very slightly. 
The differences in the prediction vector of the non-members (as measured by \testvar) has a greater effect on the MI attack accuracy. This validates the prior work hypothesis (Q2) that the differences in the shadow and target model affect the MI attack.
The estimated ATE of \trainvar on \oakacc is \todo{$0.02$}, whereas that of \testvar is \todo{$0.94$} (Table~\ref{tab:ce-undefended-prior-work}).  
For MSE-trained models, the gap in \accdiff does not cause an increase in the multiple shadow model attack accuracy--the inferred model does not have a causal path to the MI attack.
The variance in the non-member predictions has a causal effect on the MI attack accuracy, though it is less than for CE-trained models (Table~\ref{tab:mse-undefended-prior-work}). There is no causal effect of the variance of the members on the MI attack accuracy.
These two findings invalidate prior work hypotheses for MSE-trained models.

\begin{table}[htb]
\caption{We compute the average effect on the $6$ evaluated MI attacks of the causes mentioned in prior works for CE-trained models.}
\label{tab:ce-undefended-prior-work}
\centering 
\resizebox{0.8\columnwidth}{!}{%
\begin{filecontents*}{data/causal_estimates-wd_0.0005-scaler-w_dk-new-out.ce.csv}
,Attack,Feature,ATE,p-value
40,\mlleakacc,\accdiff,    0.1798,4.82E-58
43,\mlleakacc,\centroid,    0.0000,0.00E+00
39,\mlleakacc,\lossdiff,    0.0000,0.00E+00
41,\mlleakacc,\numparams,    0.1609,1.42E-04
36,\mlleakacc,\testbias,    3.9476,3.55E-06
38,\mlleakacc,\testvar,    0.8342,1.26E-06
35,\mlleakacc,\trainbias,    0.0000,0.00E+00
42,\mlleakacc,\trainsize,   -0.0890,1.24E-06
37,\mlleakacc,\trainvar,   -0.3385,4.25E-02
49,\mlleaklacc,\accdiff,    0.1817,2.03E-95
52,\mlleaklacc,\centroid,   -0.2877,1.53E-07
48,\mlleaklacc,\lossdiff,    0.0000,0.00E+00
50,\mlleaklacc,\numparams,    0.1607,1.64E-05
45,\mlleaklacc,\testbias,    3.3946,5.15E-05
47,\mlleaklacc,\testvar,    0.8382,3.30E-04
44,\mlleaklacc,\trainbias,    0.0000,0.00E+00
51,\mlleaklacc,\trainsize,   -0.1052,2.10E-09
46,\mlleaklacc,\trainvar,   -0.2395,1.95E-02
5,\memguard,\accdiff,    0.0805,5.04E-11
8,\memguard,\centroid,    0.1410,6.87E-16
4,\memguard,\lossdiff,    0.0000,0.00E+00
6,\memguard,\numparams,    0.0393,1.91E-02
1,\memguard,\testbias,    1.7732,4.67E-03
3,\memguard,\testvar,    0.1690,5.56E-02
0,\memguard,\trainbias,    0.0000,0.00E+00
7,\memguard,\trainsize,   -0.0587,1.77E-09
2,\memguard,\trainvar,   -0.0345,1.99E-01
14,\oakacc,\accdiff,    0.2949,1.54E-136
13,\oakacc,\lossdiff,    0.0000,0.00E+00
15,\oakacc,\numparams,    0.1536,1.03E-05
10,\oakacc,\testbias,    2.8590,3.14E-05
12,\oakacc,\testvar,    0.9483,1.40E-04
9,\oakacc,\trainbias,    0.0000,0.00E+00
16,\oakacc,\trainsize,   -0.1137,2.03E-13
11,\oakacc,\trainvar,    0.0243,3.20E-02
58,\threshacc,\accdiff,    0.2715,5.63E-83
57,\threshacc,\lossdiff,    1.4711,2.48E-01
59,\threshacc,\numparams,    0.1374,1.33E-01
54,\threshacc,\testbias,    1.6541,1.85E-04
56,\threshacc,\testvar,    1.0284,1.44E-03
53,\threshacc,\trainbias,    0.0000,0.00E+00
60,\threshacc,\trainsize,   -0.0893,2.64E-09
55,\threshacc,\trainvar,    0.2026,2.19E-03
22,\mlleaktopacc,\accdiff,    0.1773,3.14E-59
25,\mlleaktopacc,\centroid,    0.2741,8.65E-21
21,\mlleaktopacc,\lossdiff,    0.0000,0.00E+00
23,\mlleaktopacc,\numparams,    0.1415,8.46E-02
18,\mlleaktopacc,\testbias,    3.8692,1.36E-06
20,\mlleaktopacc,\testvar,    0.7784,6.85E-07
17,\mlleaktopacc,\trainbias,    0.0000,0.00E+00
24,\mlleaktopacc,\trainsize,   -0.0937,8.03E-08
19,\mlleaktopacc,\trainvar,   -0.3431,3.55E-02
31,\mlleaktoplacc,\accdiff,    0.1578,4.96E-54
34,\mlleaktoplacc,\centroid,    0.2753,6.41E-19
30,\mlleaktoplacc,\lossdiff,    0.0000,0.00E+00
32,\mlleaktoplacc,\numparams,    0.1602,6.13E-05
27,\mlleaktoplacc,\testbias,    3.9796,9.55E-07
29,\mlleaktoplacc,\testvar,    0.7538,4.85E-07
26,\mlleaktoplacc,\trainbias,    0.0000,0.00E+00
33,\mlleaktoplacc,\trainsize,   -0.0971,5.57E-08
28,\mlleaktoplacc,\trainvar,   -0.3991,2.93E-02
\end{filecontents*}
\csvreader[mystyle]{data/causal_estimates-wd_0.0005-scaler-w_dk-new-out.ce.csv}{}{\csvcolii & \csvcoliii & \csvcoliv & \csvcolv}
}
\end{table}

\begin{table}[htb]
\caption{We compute the average effect on the $6$ evaluated attacks of each features over the MSE-trained models.}
\label{tab:mse-undefended-prior-work}
\centering 
\resizebox{0.8\columnwidth}{!}{%
\begin{filecontents*}{data/causal_estimates-wd_0.0005-scaler-w_dk-new-out.mse.csv}
,Attack,Feature,ATE,p-value
40,\mlleakacc,\accdiff,    0.0000,0.00E+00
43,\mlleakacc,\centroid,    0.0000,0.00E+00
39,\mlleakacc,\lossdiff,    0.2719,7.31E-63
41,\mlleakacc,\numparams,    0.1787,1.18E-01
36,\mlleakacc,\testbias,    0.0000,0.00E+00
38,\mlleakacc,\testvar,    0.0000,0.00E+00
35,\mlleakacc,\trainbias,    0.0000,0.00E+00
42,\mlleakacc,\trainsize,   -0.0857,3.96E-03
37,\mlleakacc,\trainvar,   -0.0472,9.27E-01
49,\mlleaklacc,\accdiff,    0.0000,0.00E+00
52,\mlleaklacc,\centroid,    0.0000,0.00E+00
48,\mlleaklacc,\lossdiff,    0.3416,1.25E-103
50,\mlleaklacc,\numparams,    0.1647,6.61E-02
45,\mlleaklacc,\testbias,    0.0000,0.00E+00
47,\mlleaklacc,\testvar,    0.0000,0.00E+00
44,\mlleaklacc,\trainbias,    0.0000,0.00E+00
51,\mlleaklacc,\trainsize,   -0.1015,2.11E-03
46,\mlleaklacc,\trainvar,    0.0618,5.31E-01
5,\memguard,\accdiff,    0.0000,0.00E+00
8,\memguard,\centroid,    0.0903,9.28E-01
4,\memguard,\lossdiff,    0.1278,6.82E-03
6,\memguard,\numparams,    0.0469,1.50E-01
1,\memguard,\testbias,    0.0000,0.00E+00
3,\memguard,\testvar,    0.0000,0.00E+00
0,\memguard,\trainbias,    0.0000,0.00E+00
7,\memguard,\trainsize,   -0.0371,9.67E-03
2,\memguard,\trainvar,    0.0023,9.04E-01
14,\oakacc,\accdiff,    0.0000,0.00E+00
13,\oakacc,\lossdiff,    0.6194,2.86E-01
15,\oakacc,\numparams,   -0.0055,9.43E-03
10,\oakacc,\testbias,    0.1960,2.39E-01
12,\oakacc,\testvar,    0.2057,1.25E-01
9,\oakacc,\trainbias,    0.6174,1.28E-05
16,\oakacc,\trainsize,   -0.0881,4.25E-03
11,\oakacc,\trainvar,    0.0277,3.94E-01
58,\threshacc,\accdiff,    0.0000,0.00E+00
57,\threshacc,\lossdiff,   -0.6702,1.20E-12
59,\threshacc,\numparams,    0.0111,3.30E-02
54,\threshacc,\testbias,    0.0631,3.61E-01
56,\threshacc,\testvar,    0.2835,3.88E-01
53,\threshacc,\trainbias,   -0.2563,3.44E-05
60,\threshacc,\trainsize,   -0.0954,2.91E-03
55,\threshacc,\trainvar,    0.0450,2.47E-01
22,\mlleaktopacc,\accdiff,    0.0000,0.00E+00
25,\mlleaktopacc,\centroid,    0.0000,0.00E+00
21,\mlleaktopacc,\lossdiff,    0.2147,8.43E-10
23,\mlleaktopacc,\numparams,    0.1107,6.76E-02
18,\mlleaktopacc,\testbias,   -0.0580,9.64E-02
20,\mlleaktopacc,\testvar,    0.2350,5.08E-02
17,\mlleaktopacc,\trainbias,    0.0000,0.00E+00
24,\mlleaktopacc,\trainsize,   -0.0888,1.72E-03
19,\mlleaktopacc,\trainvar,   -0.1551,7.91E-01
31,\mlleaktoplacc,\accdiff,    0.0000,0.00E+00
34,\mlleaktoplacc,\centroid,    0.0000,0.00E+00
30,\mlleaktoplacc,\lossdiff,    0.2783,1.46E-65
32,\mlleaktoplacc,\numparams,    0.1645,7.26E-02
27,\mlleaktoplacc,\testbias,    0.0000,0.00E+00
29,\mlleaktoplacc,\testvar,    0.0000,0.00E+00
26,\mlleaktoplacc,\trainbias,    0.0000,0.00E+00
33,\mlleaktoplacc,\trainsize,   -0.0935,6.53E-03
28,\mlleaktoplacc,\trainvar,   -0.0627,8.80E-01
\end{filecontents*}
\csvreader[mystyle]{data/causal_estimates-wd_0.0005-scaler-w_dk-new-out.mse.csv}{}{\csvcolii & \csvcoliii & \csvcoliv & \csvcolv}
}
\end{table}

\paragraph{Single Shadow Model Attacks.}
The largest influence on the single shadow model accuracy (\mlleaktopacc(-l)) is the centroid distance between members and non-members (\centroid), thus confirming prior work hypothesis (Q8). Our approach singles out the effect of the \centroid from other variables such as \numparams and \trainsize which indirectly affect the \centroid itself.
In Table~\ref{tab:ce-undefended-prior-work}, the estimated ATE of the centroid distance on the \mlleaktopacc is \todo{$0.27$}.
We find that the variance of the outputs of the models is a cause for the single shadow model attack, to various degrees depending on the type of attack.
Prior work overlooks the differences in the prediction vectors between the target and shadow model. Thus, our analysis refutes prior work (Q5).
We also refute the hypothesis that there are no differences in taking only the top-3 prediction vs. the whole prediction vector (Q7). There are a number of key differences in the causes of these variations of the single shadow model attack.
The variance in the non-members' prediction vectors (\testvar) is a cause for the single shadow model attack that uses the whole prediction vector and the label as input features (\mlleaklacc).
Interestingly, the accuracy of the attacks that take the top-3 predictions (\mlleaktopacc and \mlleaktoplacc) is less sensitive to the variance of the prediction vectors compared to the single shadow model attack that uses the whole prediction vector, as well as the multiple shadow model attack (Table~\ref{tab:ce-undefended-prior-work}).
Our observation is that the variance influences the attacks that consider the whole prediction vector compared to ones that take only the top predictions as models agree on top predictions more than on the last predictions.
A larger causal effect of the variance means that the attack is sensitive to the specific changes in the prediction vector influenced by the dataset or randomness.
Thus, attacks that are robust to these changes on average can more readily transfer membership information beyond the dataset and architecture of the target model.

\paragraph{Threshold-based Attack.}
\lossdiff is a significant cause for the threshold attack accuracy, as expected.
On average, the variance of the prediction vectors significantly influence the average performance of the loss-based attack.
We find the ATE of the \lossdiff on the \threshacc to be close to \todo{$1.37$} (Table~\ref{tab:ce-undefended-prior-work}).
On closer inspection, beyond the prior work hypothesis, we find that there are other causes. For instance, the variance of the prediction vectors causes the MI attack accuracy. Both \trainvar and \testvar have an estimated ATE of around \todo{$0.20$} and \todo{$1.02$}, respectively.
For models trained with MSE, the train-to-test loss difference (\lossdiff) consistently has a causal effect on the MI attack performance rather than the \traintestgap.

\subsection{Analysis of Defenses}
\label{sec:appdx-defenses}

We consider two defenses: L2-regularization~\cite{shokri2017membership} and MemGuard~\cite{jia2019memguard}.

\noindent\textbf{L2-regularization Setup.} 
The first type of defense requires a simple change to one of the parameters of the training algorithm, i.e., the weight decay. The rest of the training procedure is the same, resulting in the same number of models with and without regularization.
We run all of the attacks on the regularized models to evaluate how the defense changes the effect of certain factors on the MI attack accuracy. 

\noindent\textbf{MemGuard Setup.}
The MemGuard defense requires in total $4$ models:
\begin{itemize}
    \item Target Model: the model to be defended.
    \item Defense Model: the attack model trained by defenders. The model is trained with the training set which considers the training set of the target model as members and  the testing set of the target model as non-members.
    \item Shadow Model: the model trained by attackers which has the same architecture as Target Model but is trained with a different dataset
    \item Attack Model: the attack model trained by attackers. The model is trained with the training set which considers the training set of the shadow model as members and  the testing set of the shadow model as non-members. Note that the non-members used to train the attack model need to be different from the non-members used to train the defense model.
\end{itemize}

For our evaluation, we defend $1$ target model per repeat (on average) using MemGuard, using $2$ other models in the same repeat, which gives us $2$ defended models per target model. In total, we generate $6$ new defended models that have almost the same accuracy as the corresponding target model. We evaluate these $6$ models on the testing samples, and compute bias-variance of the outputs of these $6$ models. The resulting graphs are also computed with these updated bias-variance quantities.

\begin{table}[htb]
\caption{We compute the average effect on the $6$ evaluated MI attacks of the causes mentioned in prior works for CE-trained models with L2-regularization.}
\label{tab:ce-defended-analysis}
\centering 
\resizebox{0.8\columnwidth}{!}{%
\begin{filecontents*}{data/causal_estimates-wd_0.005-scaler-w_dk-new-out.ce.csv}
,Attack,Feature,ATE,p-value
31,\mlleakacc,\accdiff,    0.2314,5.78E-84
34,\mlleakacc,\centroid,    0.0000,0.00E+00
30,\mlleakacc,\lossdiff,    0.0000,0.00E+00
32,\mlleakacc,\numparams,    0.1673,2.08E-07
27,\mlleakacc,\testbias,  -75.6790,2.97E-01
29,\mlleakacc,\testvar,    0.1425,3.66E-06
26,\mlleakacc,\trainbias,    0.0000,0.00E+00
33,\mlleakacc,\trainsize,   -0.1267,2.62E-12
28,\mlleakacc,\trainvar,   -0.3409,3.72E-05
40,\mlleaklacc,\accdiff,    0.2833,2.42E-78
43,\mlleaklacc,\centroid,    0.0000,0.00E+00
39,\mlleaklacc,\lossdiff,    1.2445,5.12E-03
41,\mlleaklacc,\numparams,    0.1859,7.27E-09
36,\mlleaklacc,\testbias,    5.5391,9.45E-01
38,\mlleaklacc,\testvar,    0.1580,1.09E-06
35,\mlleaklacc,\trainbias,    0.0000,0.00E+00
42,\mlleaklacc,\trainsize,   -0.1428,9.59E-14
37,\mlleaklacc,\trainvar,   -0.3831,2.80E-06
5,\oakacc,\accdiff,    0.2873,4.69E-178
4,\oakacc,\lossdiff,    0.0000,0.00E+00
6,\oakacc,\numparams,    0.2253,1.21E-10
1,\oakacc,\testbias,   65.4845,3.24E-01
3,\oakacc,\testvar,    0.4880,2.39E-02
0,\oakacc,\trainbias,    0.0000,0.00E+00
7,\oakacc,\trainsize,   -0.1626,3.07E-21
2,\oakacc,\trainvar,   -0.0402,1.28E-07
49,\threshacc,\accdiff,    0.2294,5.86E-125
48,\threshacc,\lossdiff,    0.0000,0.00E+00
50,\threshacc,\numparams,    0.2045,6.11E-12
45,\threshacc,\testbias,  107.4913,1.16E-01
47,\threshacc,\testvar,    0.6449,2.03E-01
44,\threshacc,\trainbias,    0.0000,0.00E+00
51,\threshacc,\trainsize,   -0.1223,3.35E-18
46,\threshacc,\trainvar,   -0.0734,5.86E-05
13,\mlleaktopacc,\accdiff,    0.2288,6.95E-85
16,\mlleaktopacc,\centroid,    0.2446,5.14E-33
12,\mlleaktopacc,\lossdiff,    0.0000,0.00E+00
14,\mlleaktopacc,\numparams,    0.1662,1.14E-07
9,\mlleaktopacc,\testbias,  -45.7867,5.31E-01
11,\mlleaktopacc,\testvar,    0.1115,1.11E-06
8,\mlleaktopacc,\trainbias,    0.0000,0.00E+00
15,\mlleaktopacc,\trainsize,   -0.1322,4.51E-13
10,\mlleaktopacc,\trainvar,   -0.3589,1.20E-05
22,\mlleaktoplacc,\accdiff,    0.2240,3.54E-77
25,\mlleaktoplacc,\centroid,    0.2661,7.45E-32
21,\mlleaktoplacc,\lossdiff,    0.0000,0.00E+00
23,\mlleaktoplacc,\numparams,    0.1650,1.12E-07
18,\mlleaktoplacc,\testbias,  -58.3315,4.38E-01
20,\mlleaktoplacc,\testvar,    0.0932,3.47E-07
17,\mlleaktoplacc,\trainbias,    0.0000,0.00E+00
24,\mlleaktoplacc,\trainsize,   -0.1298,3.26E-12
19,\mlleaktoplacc,\trainvar,   -0.3976,2.50E-05
\end{filecontents*}
\csvreader[mystyle]{data/causal_estimates-wd_0.005-scaler-w_dk-new-out.ce.csv}{}{\csvcolii & \csvcoliii & \csvcoliv & \csvcolv}
}
\end{table}
\begin{table}[htb]
\caption{We compute the average effect on the $6$ evaluated attacks of each features over the MSE-trained models with L2-regularization.}
\label{tab:mse-defended-analysis}
\centering 
\resizebox{0.8\columnwidth}{!}{%
\begin{filecontents*}{data/causal_estimates-wd_0.005-scaler-w_dk-new-out.mse.csv}
,Attack,Feature,ATE,p-value
31,\mlleakacc,\accdiff,   -0.0951,1.81E-02
34,\mlleakacc,\centroid,    0.0000,0.00E+00
30,\mlleakacc,\lossdiff,    0.5262,4.01E-03
32,\mlleakacc,\numparams,    0.0313,7.18E-01
27,\mlleakacc,\testbias,    0.1261,6.01E-05
29,\mlleakacc,\testvar,    0.1238,3.73E-07
26,\mlleakacc,\trainbias,    0.0000,0.00E+00
33,\mlleakacc,\trainsize,   -0.0199,4.07E-03
28,\mlleakacc,\trainvar,    0.1572,1.77E-03
40,\mlleaklacc,\accdiff,    0.0732,3.91E-16
43,\mlleaklacc,\centroid,    0.0000,0.00E+00
39,\mlleaklacc,\lossdiff,    0.6361,4.58E-05
41,\mlleaklacc,\numparams,    0.0529,4.22E-01
36,\mlleaklacc,\testbias,    0.1619,8.45E-07
38,\mlleaklacc,\testvar,    0.1690,3.97E-04
35,\mlleaklacc,\trainbias,    0.0000,0.00E+00
42,\mlleaklacc,\trainsize,   -0.0292,1.62E-02
37,\mlleaklacc,\trainvar,    0.2790,5.77E-05
5,\oakacc,\accdiff,    0.0000,0.00E+00
4,\oakacc,\lossdiff,    0.2941,1.80E-91
6,\oakacc,\numparams,    0.0499,8.78E-01
1,\oakacc,\testbias,    0.0000,0.00E+00
3,\oakacc,\testvar,    0.0000,0.00E+00
0,\oakacc,\trainbias,    0.0000,0.00E+00
7,\oakacc,\trainsize,   -0.0382,7.07E-01
2,\oakacc,\trainvar,    0.4038,1.75E-29
49,\threshacc,\accdiff,    0.2069,5.91E-129
48,\threshacc,\lossdiff,    0.0000,0.00E+00
50,\threshacc,\numparams,    0.0692,6.62E-01
45,\threshacc,\testbias,   -0.2423,5.81E-10
47,\threshacc,\testvar,    0.1398,2.55E-01
44,\threshacc,\trainbias,    0.0000,0.00E+00
51,\threshacc,\trainsize,   -0.0396,1.64E-02
46,\threshacc,\trainvar,    0.2585,3.84E-03
13,\mlleaktopacc,\accdiff,    0.0000,0.00E+00
16,\mlleaktopacc,\centroid,    0.0000,0.00E+00
12,\mlleaktopacc,\lossdiff,    0.2561,4.76E-40
14,\mlleaktopacc,\numparams,    0.0363,5.62E-01
9,\mlleaktopacc,\testbias,    0.0000,0.00E+00
11,\mlleaktopacc,\testvar,    0.0223,4.19E-06
8,\mlleaktopacc,\trainbias,    0.0000,0.00E+00
15,\mlleaktopacc,\trainsize,   -0.0319,1.27E-03
10,\mlleaktopacc,\trainvar,    0.2424,5.69E-02
22,\mlleaktoplacc,\accdiff,    0.0000,0.00E+00
25,\mlleaktoplacc,\centroid,    0.0000,0.00E+00
21,\mlleaktoplacc,\lossdiff,    0.1804,6.18E-79
23,\mlleaktoplacc,\numparams,    0.0014,9.38E-01
18,\mlleaktoplacc,\testbias,    0.0000,0.00E+00
20,\mlleaktoplacc,\testvar,    0.0000,0.00E+00
17,\mlleaktoplacc,\trainbias,    0.0000,0.00E+00
24,\mlleaktoplacc,\trainsize,   -0.0261,1.50E-03
19,\mlleaktoplacc,\trainvar,    0.1858,4.47E-19
\end{filecontents*}
\csvreader[mystyle]{data/causal_estimates-wd_0.005-scaler-w_dk-new-out.mse.csv}{}{\csvcolii & \csvcoliii & \csvcoliv & \csvcolv}
}
\end{table}

\paragraph{Defense through L2-regularization.}
Our first finding is that applying this regularizer as a defense reduces the influence that the variance has on the MI attack for the multiple shadow model attack. 
For all attacks, however, variance is still a cause for the MI attack. 
For \mlleaklacc, the \trainvar on regularized models has a negative effect on the attack performance, i.e., the higher the variance, the lower the attack, which has decreased from \todo{$-0.23$} to \todo{$-0.38$} (more negative effect). In contrast, the \testvar is positive and reduces from \todo{$0.83$} to \todo{$0.15$}. We also find this for \mlleaktopacc where the variance on the non-members has an estimated ATE of \todo{$0.77$} and decreases to \todo{$0.11$} (Table~\ref{tab:ce-defended-analysis}).
We observe that regularization does not help in reducing the effect of \centroid for CE-trained models. 
Regularization does not remove the causal relationship between the main causes of the attack prior to applying this defense. 
For MSE-trained models, the effect of the cause \lossdiff is significantly reduced for the single shadow model attack using top-3 predictions (Table~\ref{tab:mse-defended-analysis}). In fact, the regularization appears to be quite effective for this attack.
The features pertaining to training size and model complexity remain causes for the attack. These have a similar influence on the MI attack accuracy even after applying the high weight decay training.

The attack accuracy \mlleaklacc is not influenced by the distance between members and non-members (\centroid) after regularization. This is visible in the graph itself, i.e., the edge is missing in the \tool graph in Fig.~\ref{fig:ce-MLLeakTop3-MLLeakLAcc} compared to Fig.~\ref{fig:ce-reg-MLLeakTop3-MLLeakLAcc}.

\paragraph{Defense through MemGuard.}
MemGuard reduces the variance for both CE and MSE models, as well as some of the causes, being more effective than L2-regularization in removing the variance effect of the members. MemGuard is more effect on MSE models, as models of the usual signals have been decreased. The effect of Bias on non-members for CE models remains a potential signal, along with \centroid.

\section{Causal Models}
\label{sec:appdx-graphs}

\begin{figure*}[ht!]
\centering
\begin{subfigure}[t]{0.45\linewidth}
    \centering
    \includegraphics[width=0.99\linewidth]{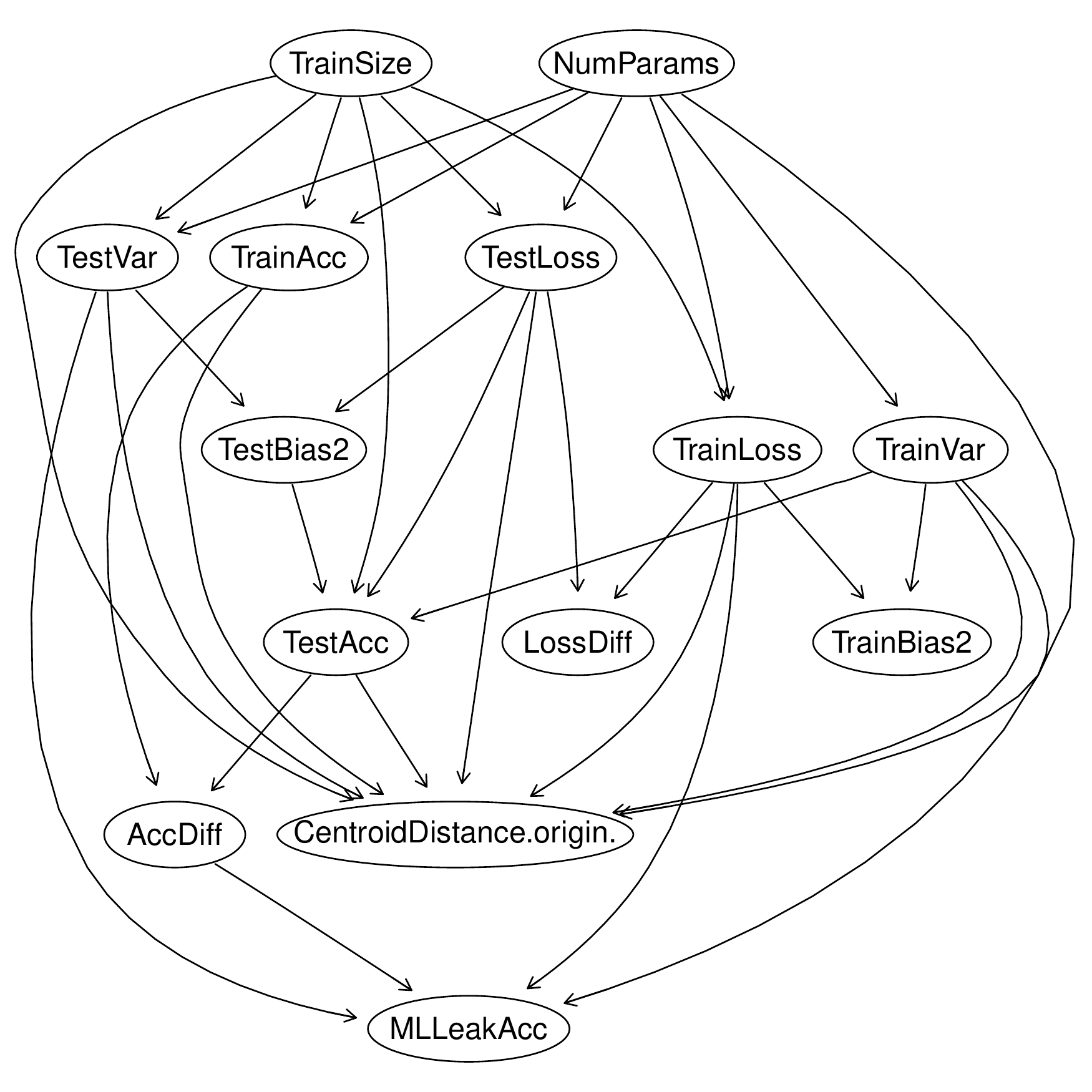}
    \subcaption[]{The causal model \tool infers for the target \mlleakacc 
    (CE-trained models).}
    \label{fig:ce-MLLeak-MLLeakAcc}
\end{subfigure}%
\hfill
\begin{subfigure}[t]{0.45\linewidth}
    \centering
    \includegraphics[width=0.99\linewidth]{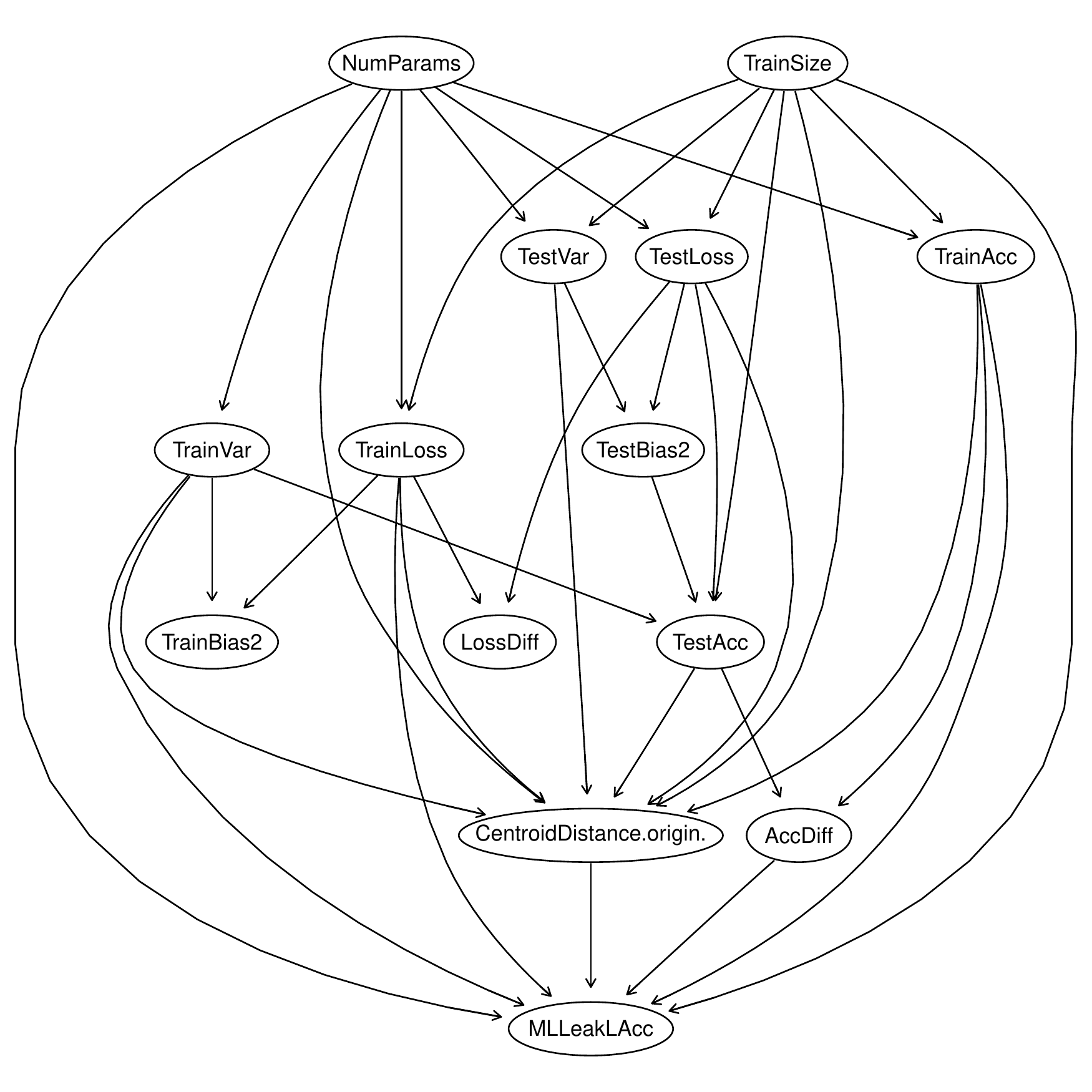}
    \subcaption[]{The causal model \tool infers for the target \mlleaklacc (CE-trained models).}
    \label{fig:ce-MLLeak-MLLeakLAcc}
\end{subfigure}
\begin{subfigure}[t]{0.45\linewidth}
    \centering
    \includegraphics[width=0.9\linewidth]{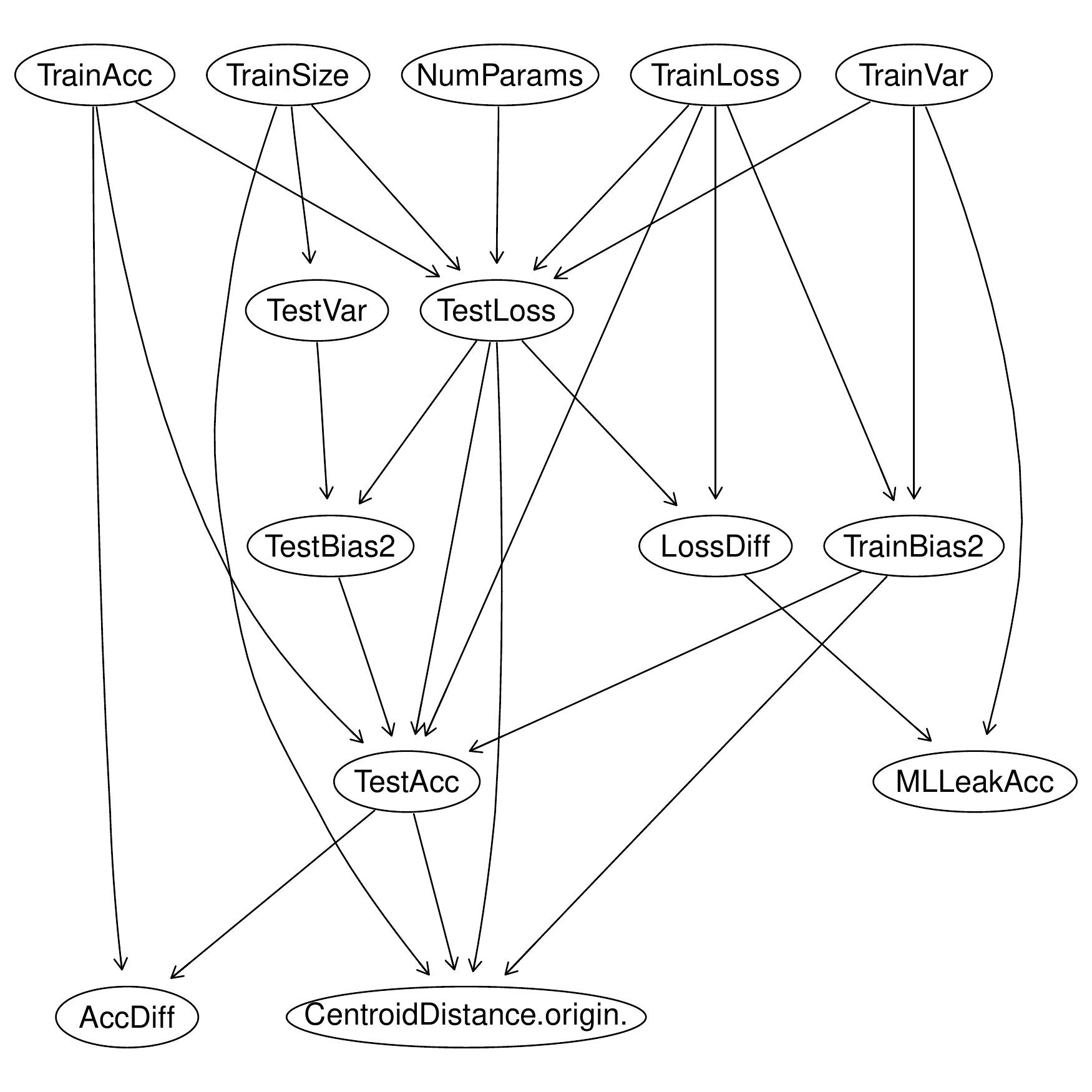}
    \subcaption[]{The causal model \tool infers for the target \mlleakacc (MSE-trained models).}
    \label{fig:mse-MLLeak-MLLeakAcc}
\end{subfigure}%
\hfill
\begin{subfigure}[t]{0.45\linewidth}
    \centering
    \includegraphics[width=0.9\linewidth]{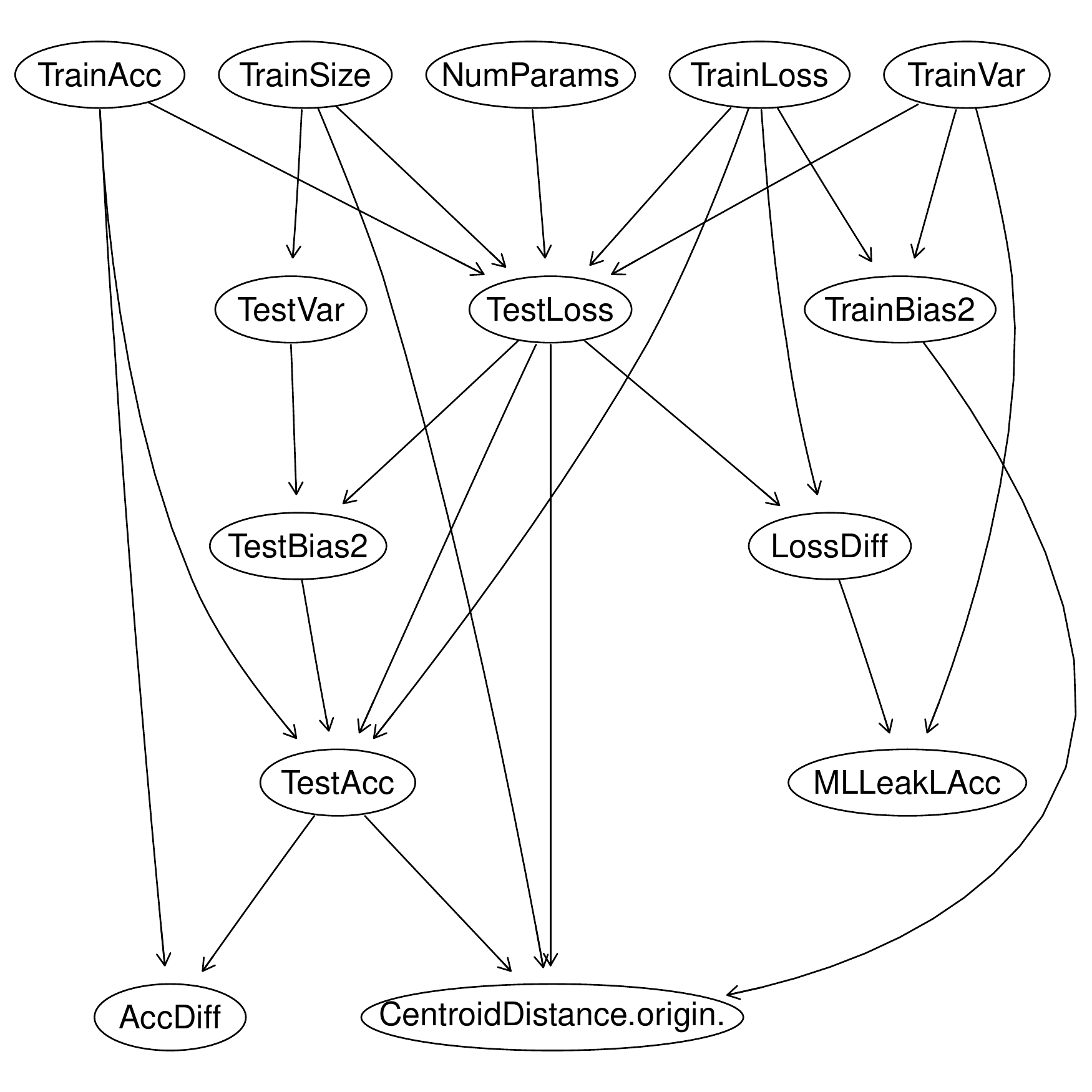}
    \subcaption[]{The causal model \tool infers for the target \mlleaklacc (MSE-trained models).}
    \label{fig:mse-MLLeak-MLLeakLAcc}
\end{subfigure}
\caption{\tool graphs for the single shadow model with top-10 prediction vector (with and without label) as input to the attack model.}
\label{fig:MLLeak}
\end{figure*}

\begin{figure*}[th!]
\centering
\begin{subfigure}[t]{0.45\linewidth}
    \centering
    \includegraphics[width=0.9\linewidth]{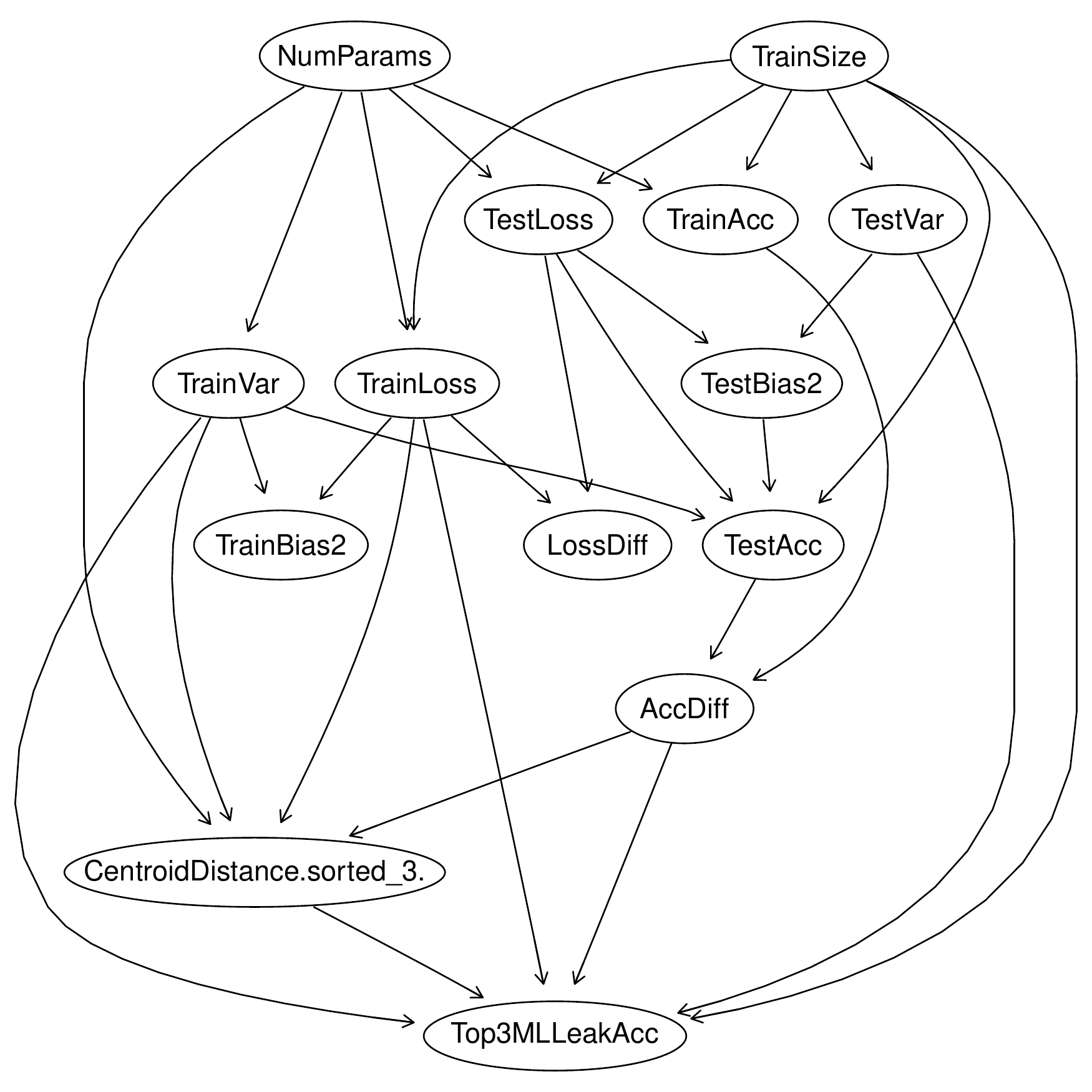}
    \subcaption[]{The causal model \tool infers for the target \mlleaktopacc (CE-trained models).}
    \label{fig:ce-MLLeakTop3-MLLeakAcc}
\end{subfigure}%
\hfill
\begin{subfigure}[t]{0.45\linewidth}
    \centering
    \includegraphics[width=0.9\linewidth]{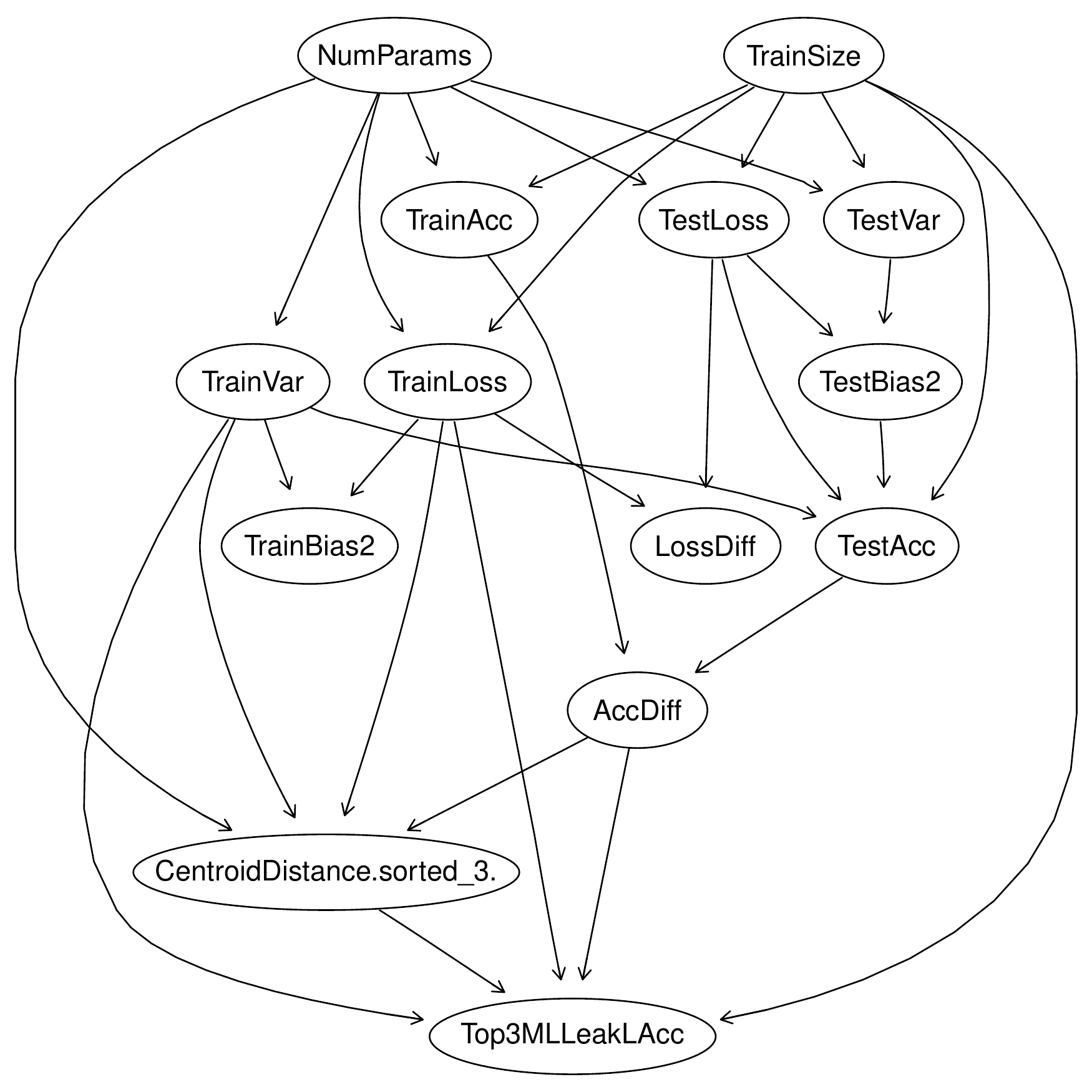}
    \subcaption[]{The causal model \tool infers for the target \mlleaktoplacc (CE-trained models).}
    \label{fig:ce-MLLeakTop3-MLLeakLAcc}
\end{subfigure}
\begin{subfigure}[t]{0.45\linewidth}
    \centering
    \includegraphics[width=0.9\linewidth]{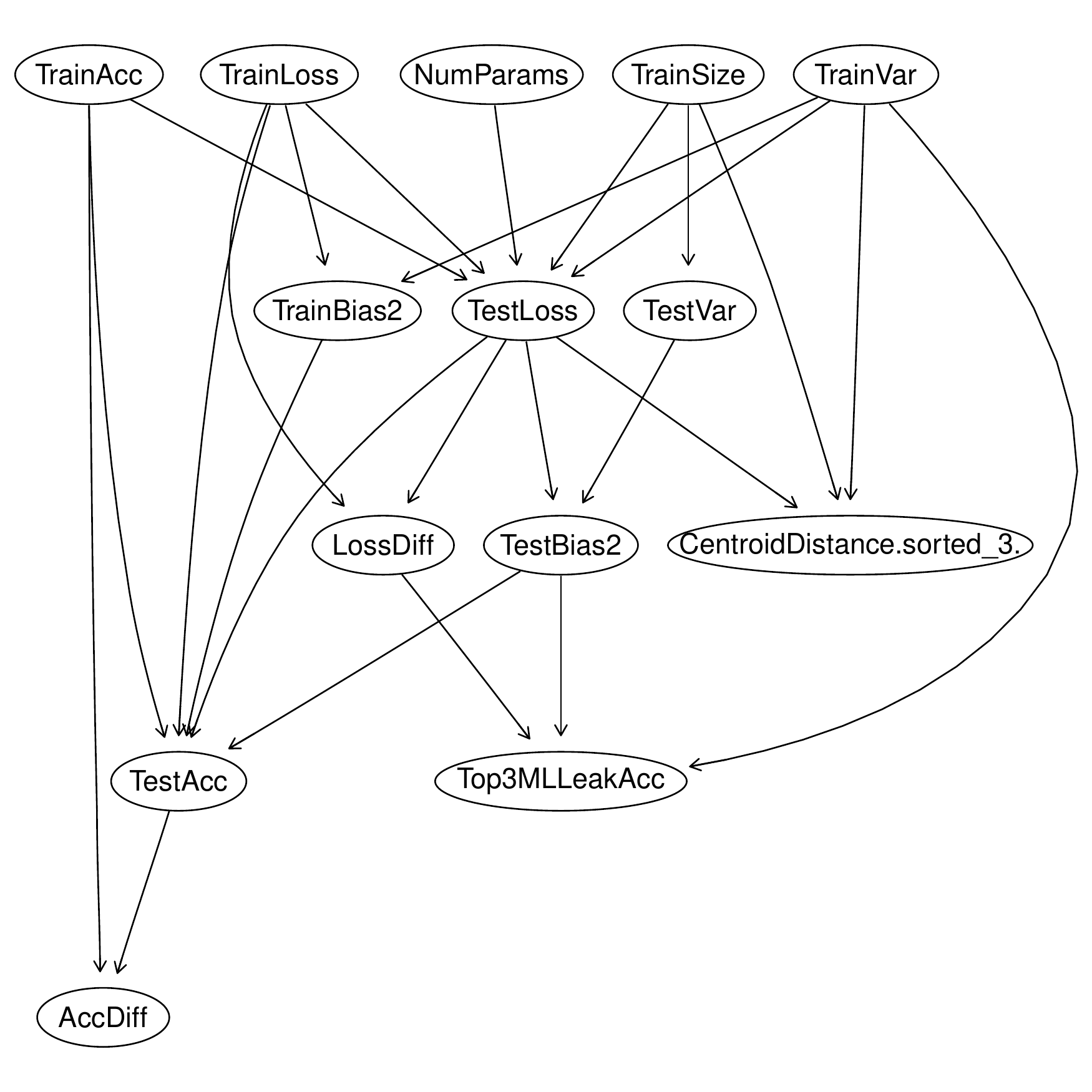}
    \subcaption[]{The causal model \tool infers for the target \mlleaktopacc (MSE-trained models).}
    \label{fig:mse-MLLeakTop3-MLLeakAcc}
\end{subfigure}%
\hfill
\begin{subfigure}[t]{0.45\linewidth}
    \centering
    \includegraphics[width=0.9\linewidth]{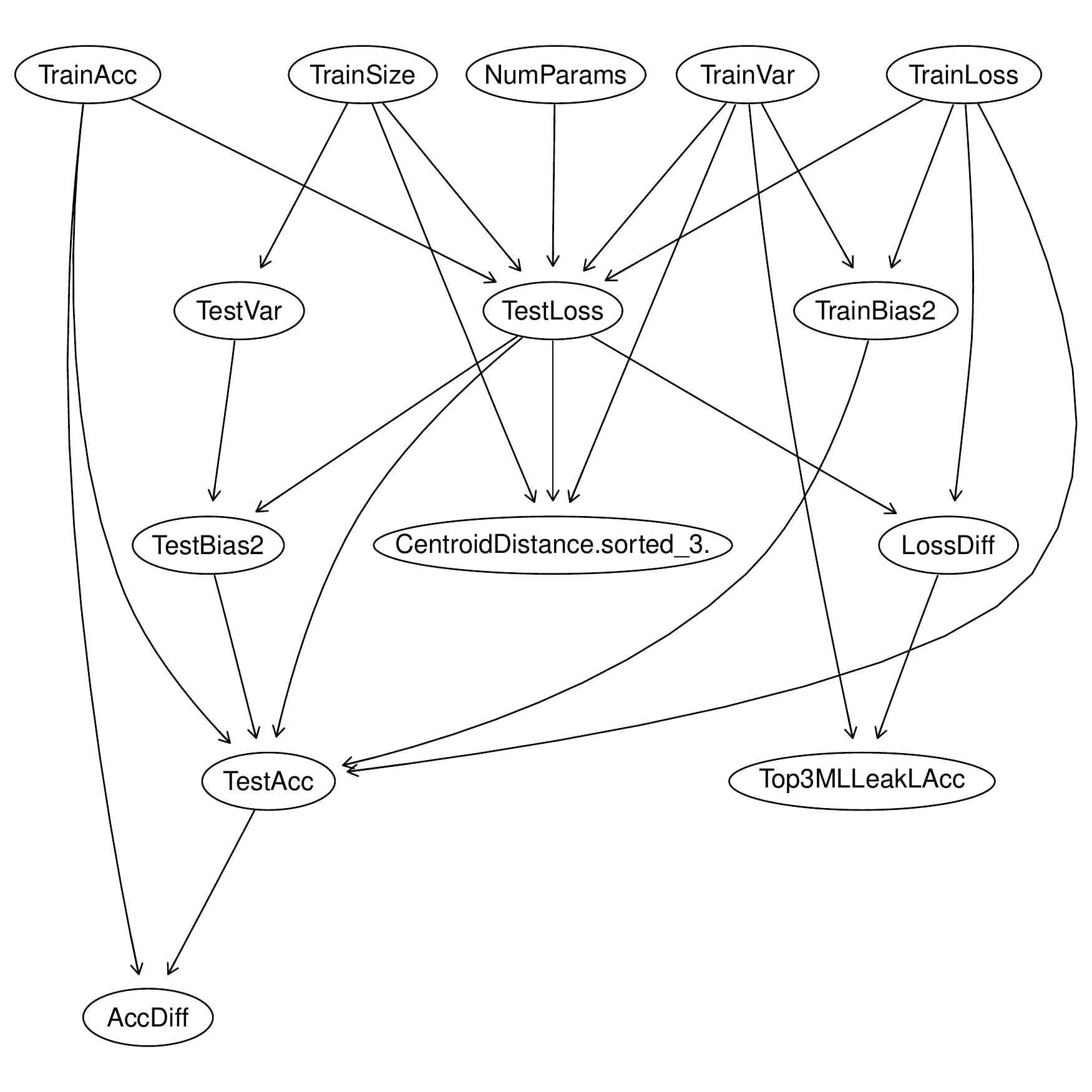}
    \subcaption[]{The causal model \tool infers for the target \mlleaktoplacc (MSE-trained models).}
    \label{fig:mse-MLLeakTop3-MLLeakLAcc}
\end{subfigure}
\caption{\tool graphs for the single shadow model that takes the top-3 prediction vector (with and without label) as input to the attack model.}
\label{fig:MLLeakTop3}
\end{figure*}

\begin{figure*}[t]
\centering
\begin{subfigure}[t]{0.45\linewidth}
    \centering
    \includegraphics[width=\linewidth]{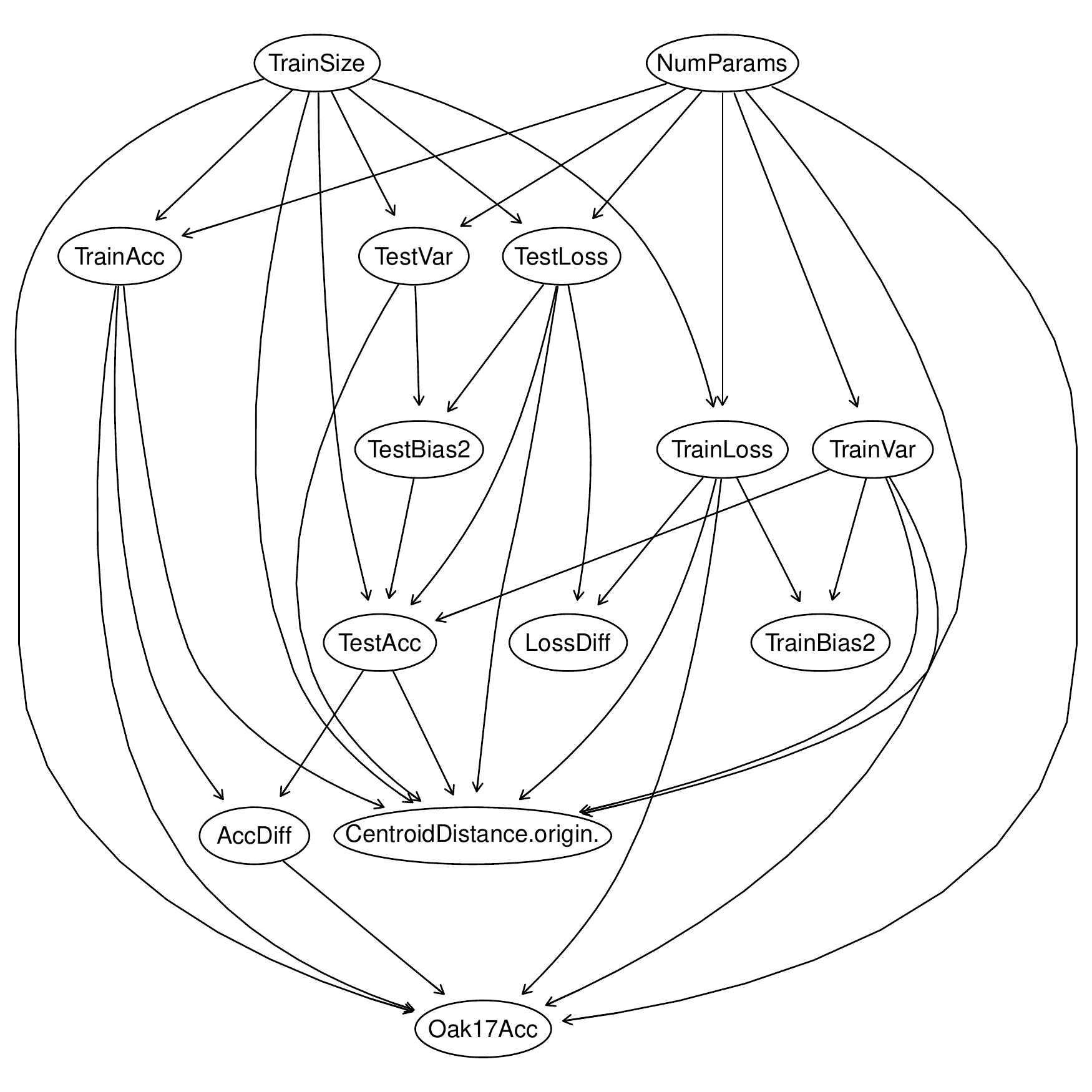}
    \subcaption[]{The causal model \tool infers for the target \oakacc (CE-trained models).}
    \label{fig:mse-ShadowAcc}
\end{subfigure}%
\hfill
\begin{subfigure}[t]{0.45\linewidth}
    \centering
    \includegraphics[width=\linewidth]{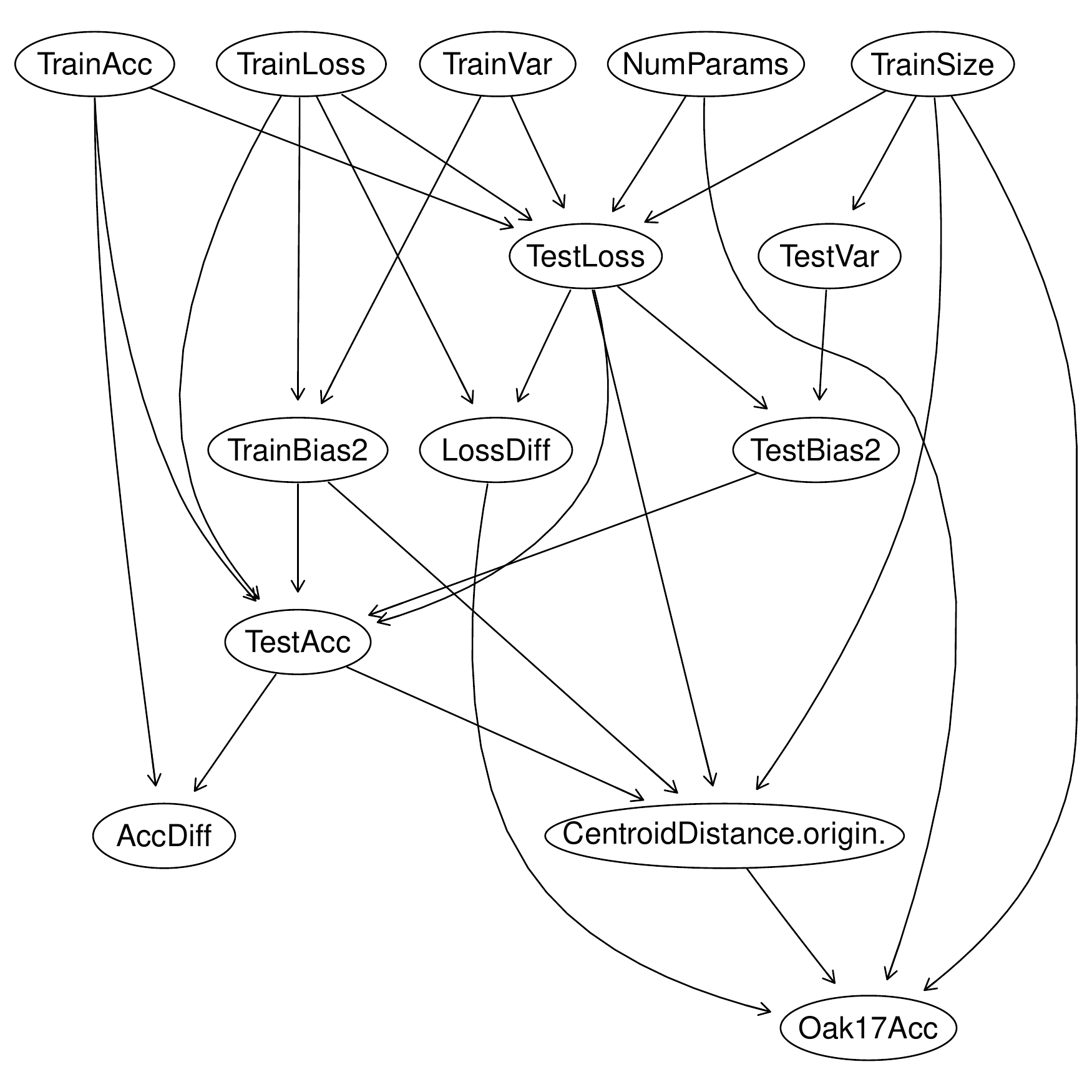}
    \subcaption[]{The causal model \tool infers for the target \oakacc (MSE-trained models).}
    \label{fig:mse-ShadowAcc}
\end{subfigure}
\caption{The causal model \tool infers for the multiple shadow model attack for CE and MSE-trained models, where the target node is \oakacc.}
\label{fig:undefended-Shadow}
\end{figure*}

\begin{figure*}[t]
\centering
\begin{subfigure}[t]{0.45\linewidth}
    \centering
    \includegraphics[width=\linewidth]{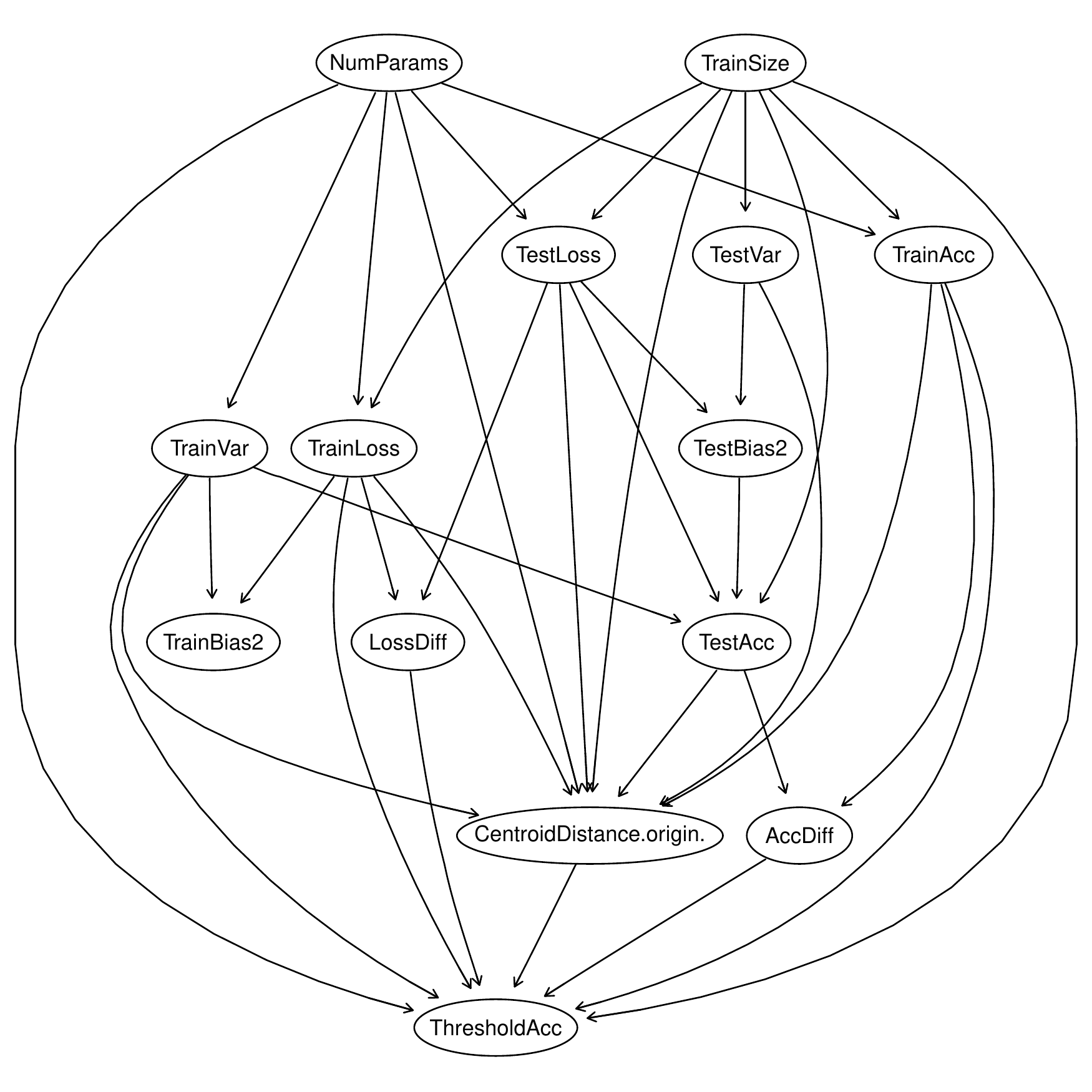}
    \subcaption[]{The causal model \tool infers for the target \threshacc (CE-trained models).}
    \label{fig:mse-ThreshAcc}
\end{subfigure}%
\hfill
\begin{subfigure}[t]{0.45\linewidth}
    \centering
    \includegraphics[width=\linewidth]{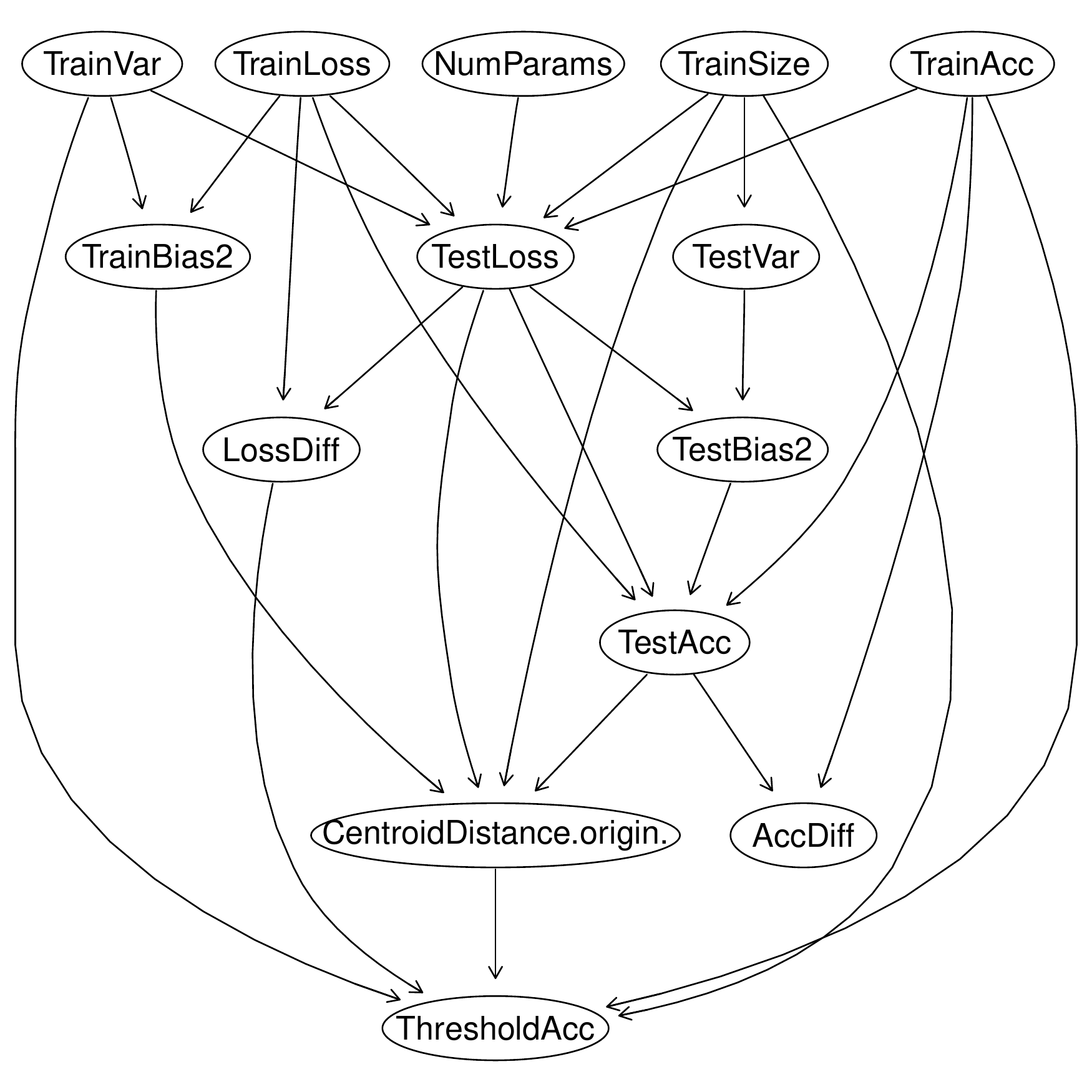}
    \subcaption[]{The causal model \tool infers for the target \threshacc (MSE-trained models).}
    \label{fig:mse-ThreshAcc}
\end{subfigure}
\caption{The causal model \tool infers for the multiple shadow model attack for CE and MSE-trained models, where the target node is \oakacc.}
\label{fig:undefended-Thresh}
\end{figure*}

\begin{figure*}[ht!]
\centering
\begin{subfigure}[t]{0.45\linewidth}
    \centering
    \includegraphics[width=0.99\linewidth]{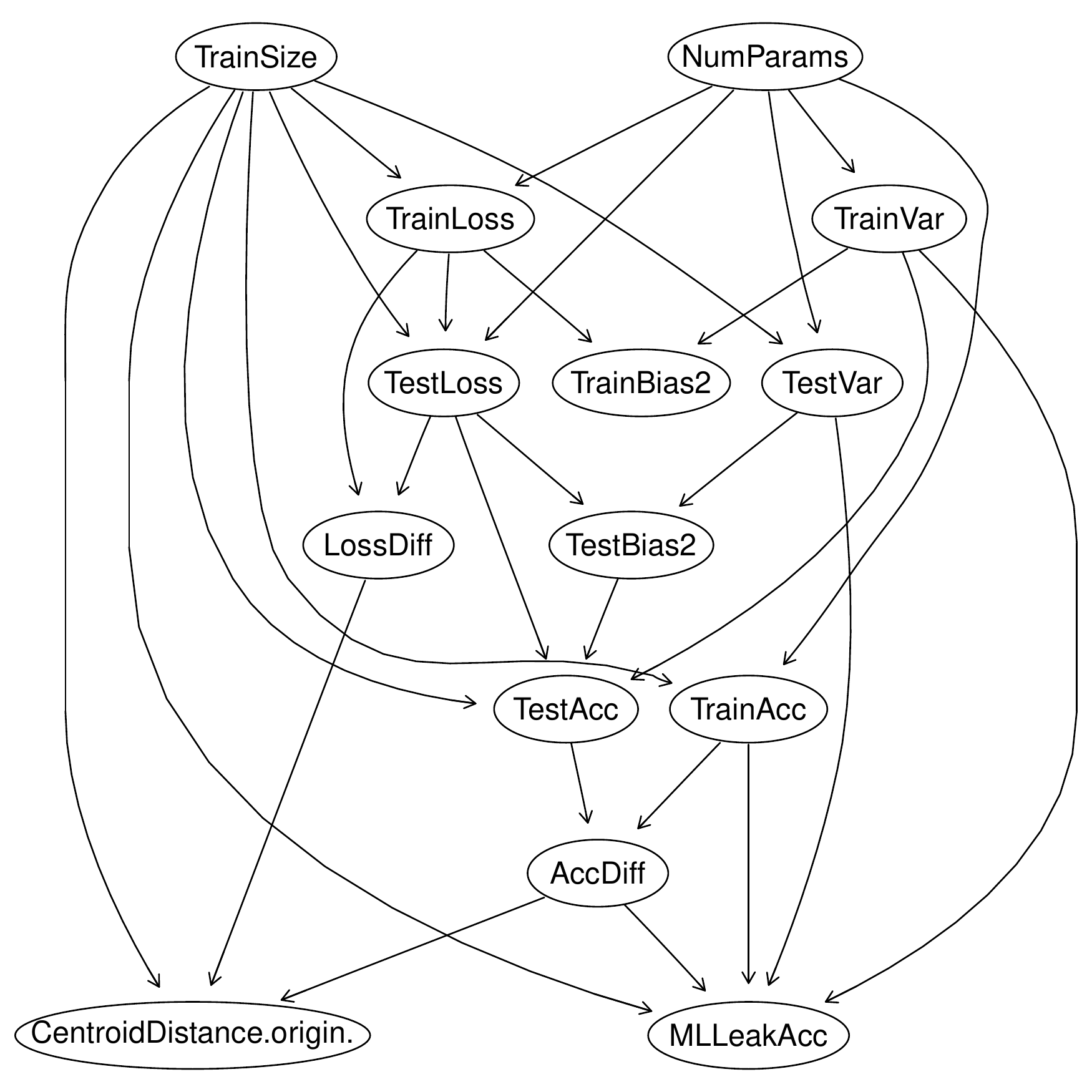}
    \subcaption[]{The causal model \tool infers for the target \mlleakacc 
    (CE-trained models with regularization).}
    \label{fig:ce-MLLeak-MLLeakAcc}
\end{subfigure}%
\hfill
\begin{subfigure}[t]{0.45\linewidth}
    \centering
    \includegraphics[width=0.99\linewidth]{figdata/graphs-w_dk-cont-wd_0.000500/ce-MLLeakLAcc-cv_avg_net_discovery.pdf}
    \subcaption[]{The causal model \tool infers for the target \mlleaklacc (CE-trained models with regularization).}
    \label{fig:ce-MLLeak-MLLeakLAcc}
\end{subfigure}
\begin{subfigure}[t]{0.45\linewidth}
    \centering
    \includegraphics[width=0.9\linewidth]{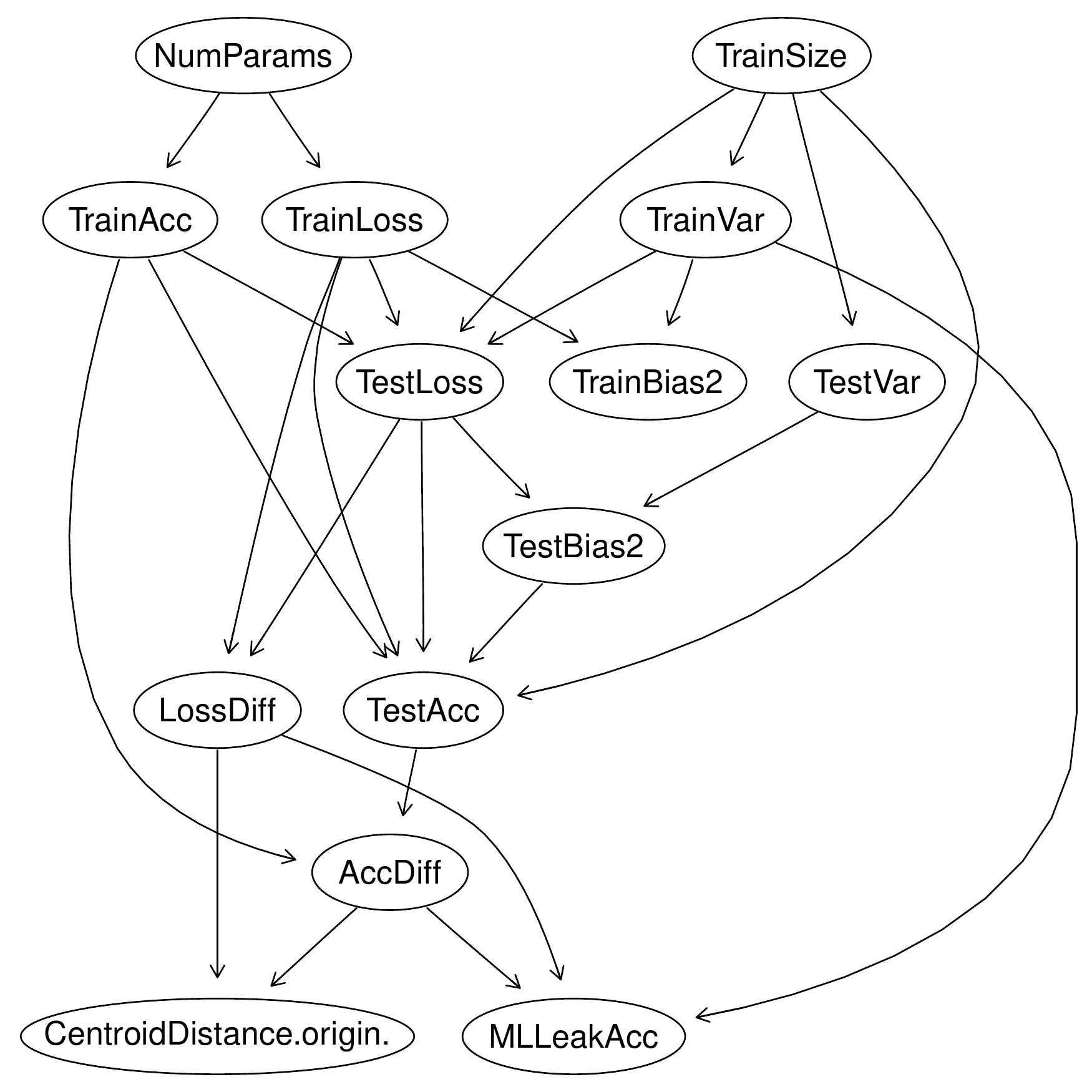}
    \subcaption[]{The causal model \tool infers for the target \mlleakacc (MSE-trained models with regularization).}
    \label{fig:mse-reg-MLLeak-MLLeakAcc}
\end{subfigure}%
\hfill
\begin{subfigure}[t]{0.45\linewidth}
    \centering
    \includegraphics[width=0.9\linewidth]{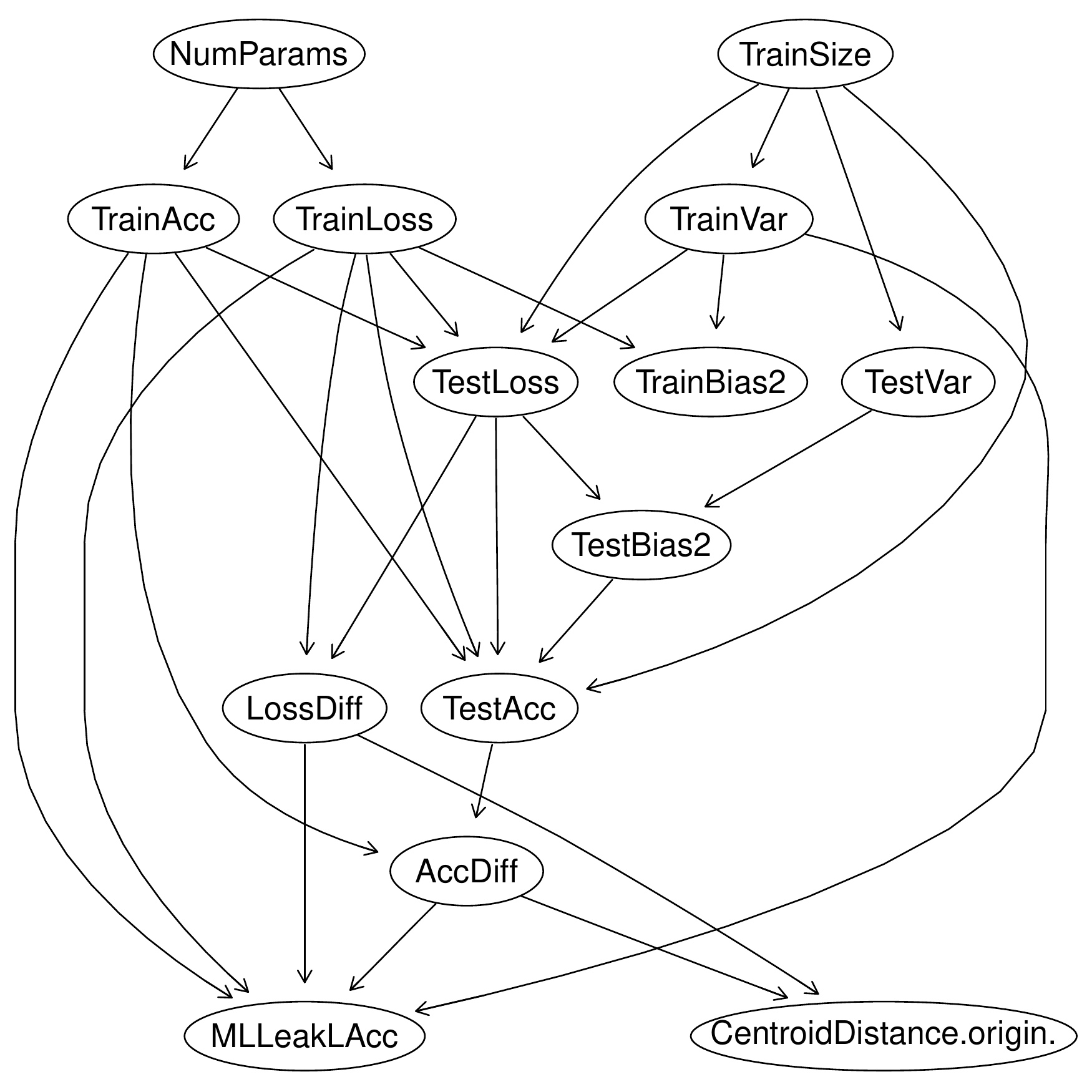}
    \subcaption[]{The causal model \tool infers for the target \mlleaklacc (MSE-trained models with regularization).}
    \label{fig:mse-reg-MLLeak-MLLeakLAcc}
\end{subfigure}
\caption{\tool graphs for the single shadow model with top-10 prediction vector (with and without label) as input to the attack model. The models have been trained with L2-regularization (weight decay=$5\times10^-3$).}
\label{fig:reg-MLLeak}
\end{figure*}

\begin{figure*}[th!]
\centering
\begin{subfigure}[t]{0.45\linewidth}
    \centering
    \includegraphics[width=0.9\linewidth]{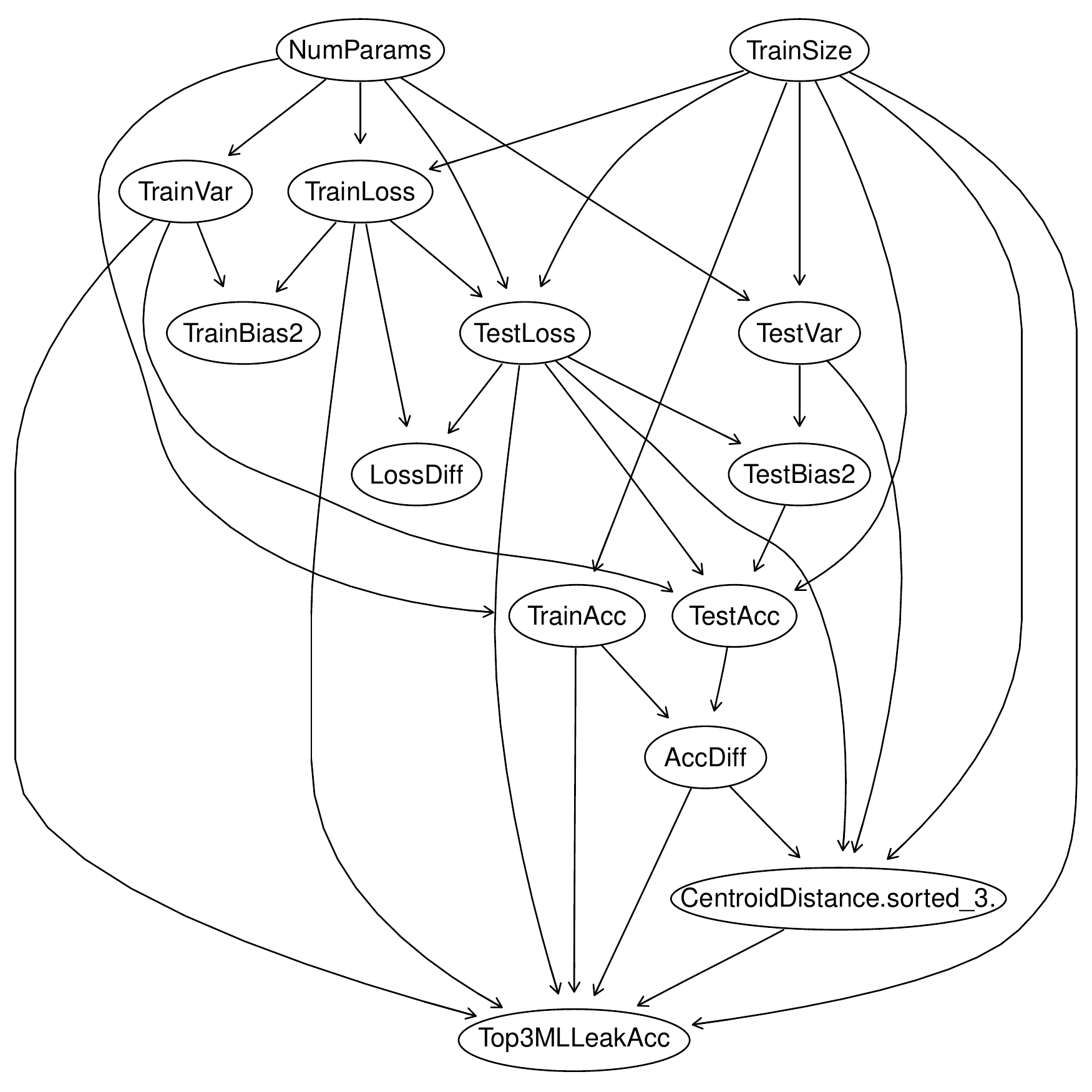}
    \subcaption[]{The causal model \tool infers for the target \mlleaktopacc (CE-trained models with regularization).}
    \label{fig:ce-reg-MLLeakTop3-MLLeakAcc}
\end{subfigure}%
\hfill
\begin{subfigure}[t]{0.45\linewidth}
    \centering
    \includegraphics[width=0.9\linewidth]{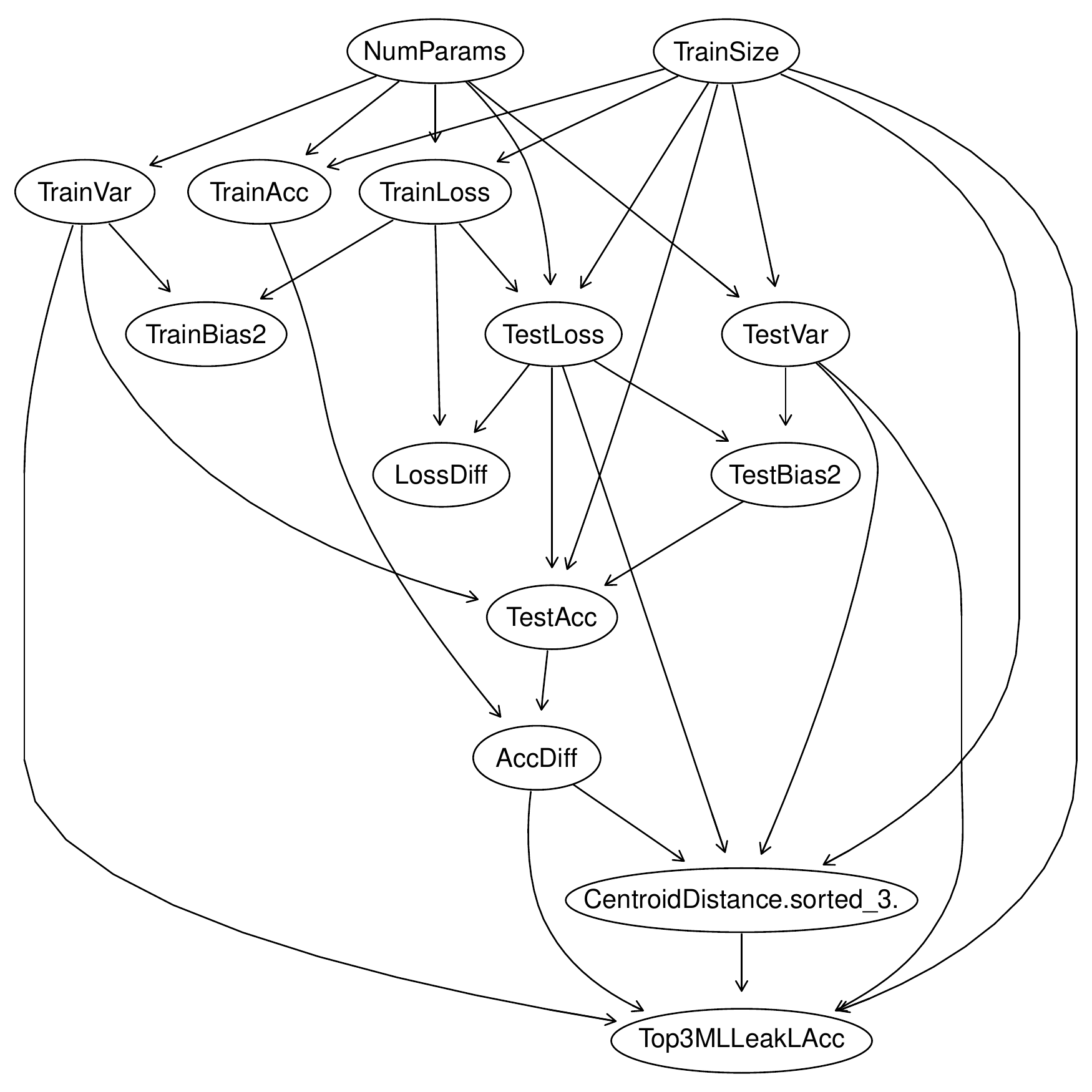}
    \subcaption[]{The causal model \tool infers for the target \mlleaktoplacc (CE-trained models with regularization).}
    \label{fig:ce-reg-MLLeakTop3-MLLeakLAcc}
\end{subfigure}
\begin{subfigure}[t]{0.45\linewidth}
    \centering
    \includegraphics[width=0.9\linewidth]{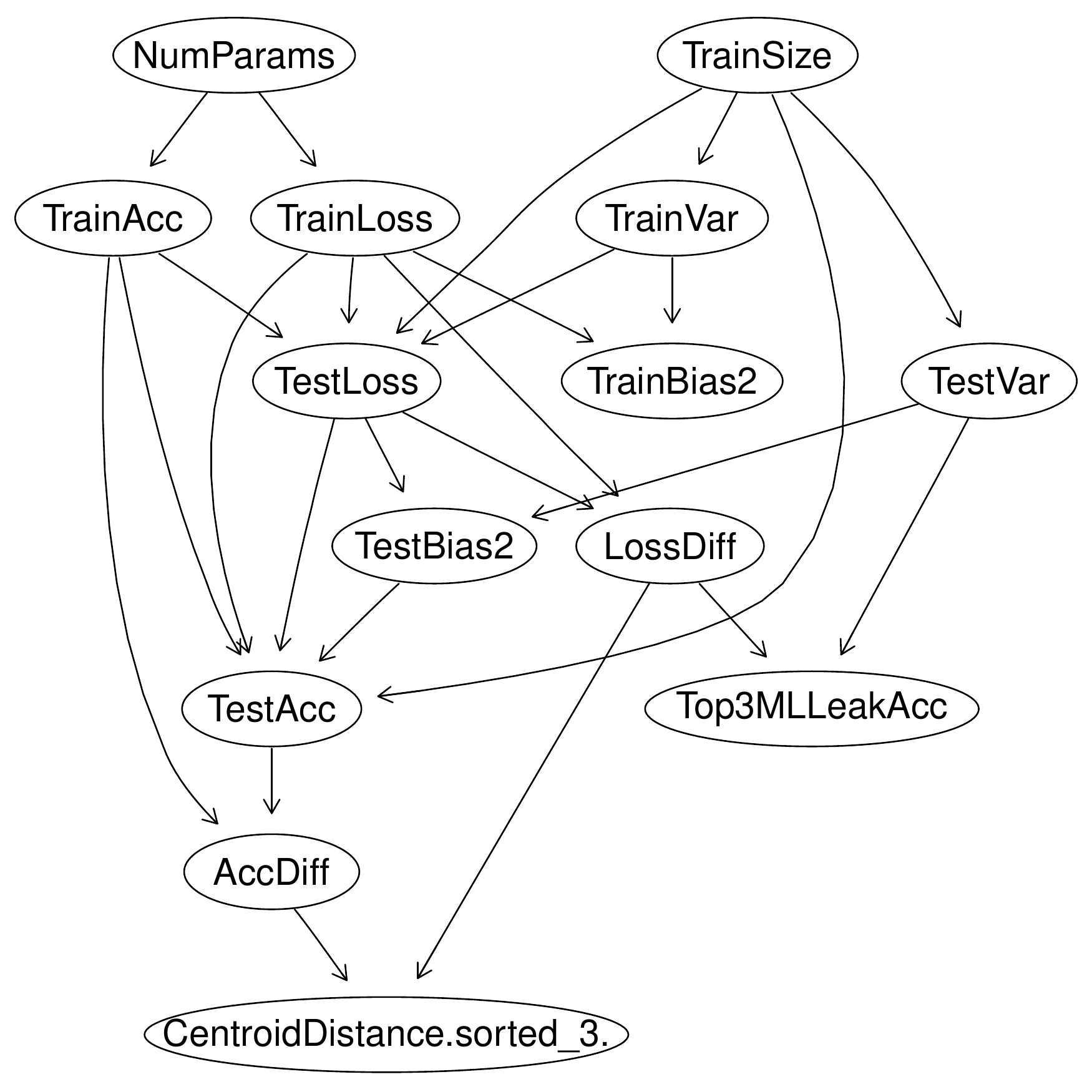}
    \subcaption[]{The causal model \tool infers for the target \mlleaktopacc (MSE-trained models with regularization).}
    \label{fig:mse-reg-MLLeakTop3-MLLeakAcc}
\end{subfigure}%
\hfill
\begin{subfigure}[t]{0.45\linewidth}
    \centering
    \includegraphics[width=0.9\linewidth]{figdata/graphs-w_dk-cont-wd_0.000500/mse-Top3MLLeakLAcc-cv_avg_net_discovery.pdf}
    \subcaption[]{The causal model \tool infers for the target \mlleaktoplacc (MSE-trained models with regularization).}
    \label{fig:mse-reg-MLLeakTop3-MLLeakLAcc}
\end{subfigure}
\caption{\tool graphs for the single shadow model that takes the top-3 prediction vector (with and without label) as input to the attack model. The models have been trained with L2-regularization (weight decay=$5\times10^-3$).}
\label{fig:reg-MLLeakTop3}
\end{figure*}

\begin{figure*}[t]
\centering
\begin{subfigure}[t]{0.45\linewidth}
    \centering
    \includegraphics[width=\linewidth]{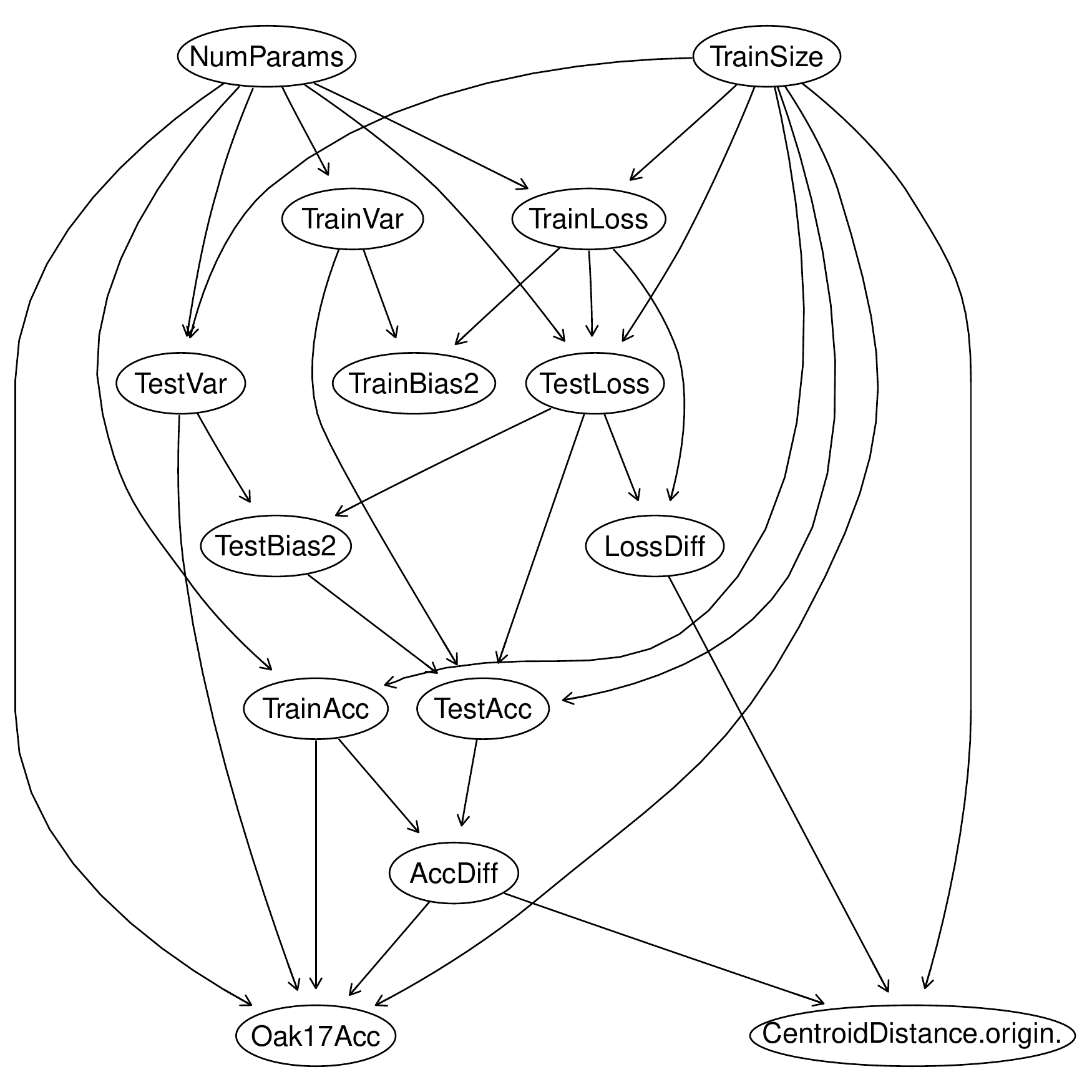}
    \subcaption[]{The causal model \tool infers for the target \oakacc (CE-trained models with regularization).}
    \label{fig:ce-reg-ShadowAcc}
\end{subfigure}%
\hfill
\begin{subfigure}[t]{0.45\linewidth}
    \centering
    \includegraphics[width=\linewidth]{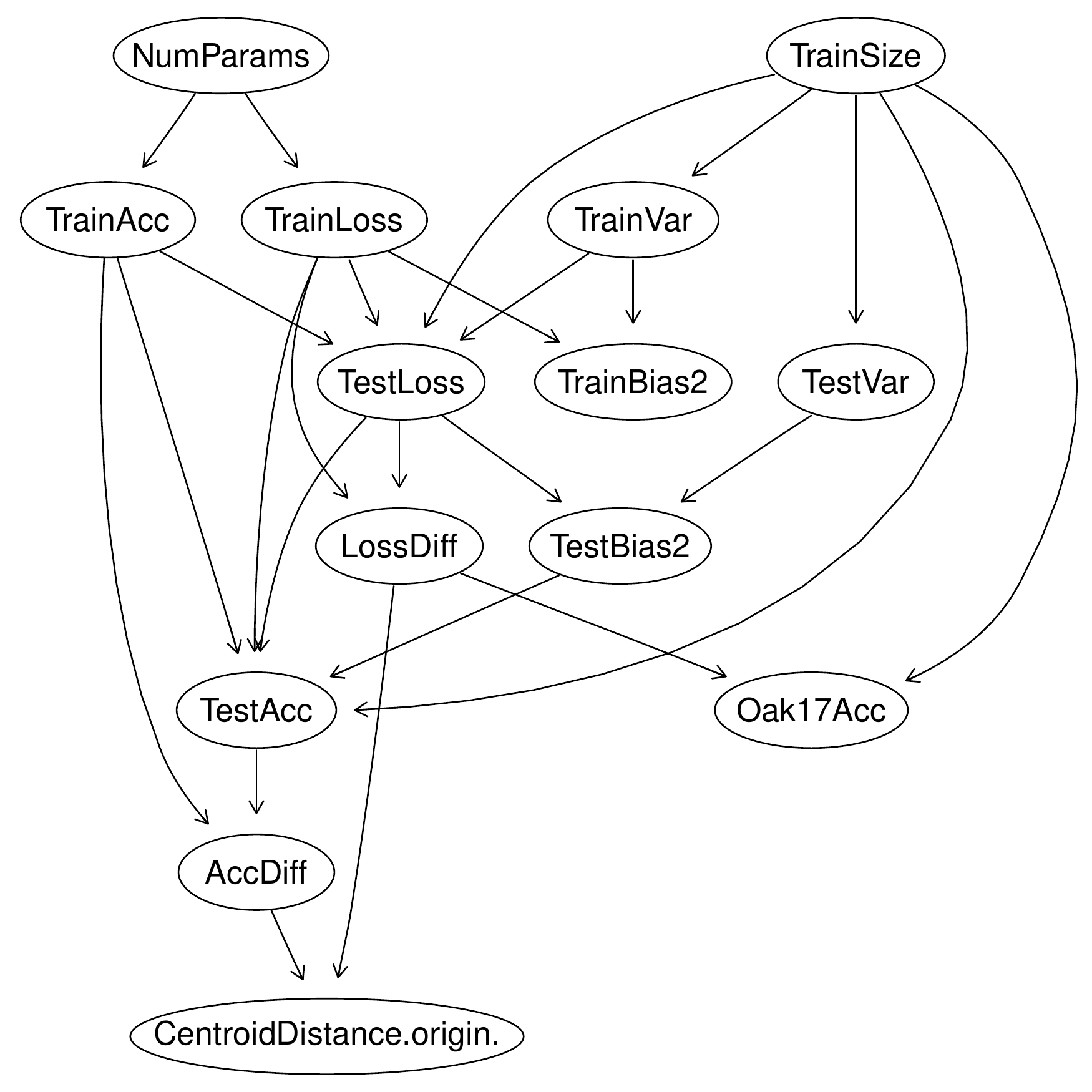}
    \subcaption[]{The causal model \tool infers for the target \oakacc (MSE-trained models with regularization).}
    \label{fig:mse-reg-ShadowAcc}
\end{subfigure}
\caption{The causal model \tool infers for the multiple shadow model attack for CE and MSE-trained models, where the target node is \oakacc. The models have been trained with L2-regularization (weight decay=$5\times10^-3$).}
\label{fig:reg-Shadow}
\end{figure*}

\begin{figure*}[t]
\centering
\begin{subfigure}[t]{0.45\linewidth}
    \centering
    \includegraphics[width=\linewidth]{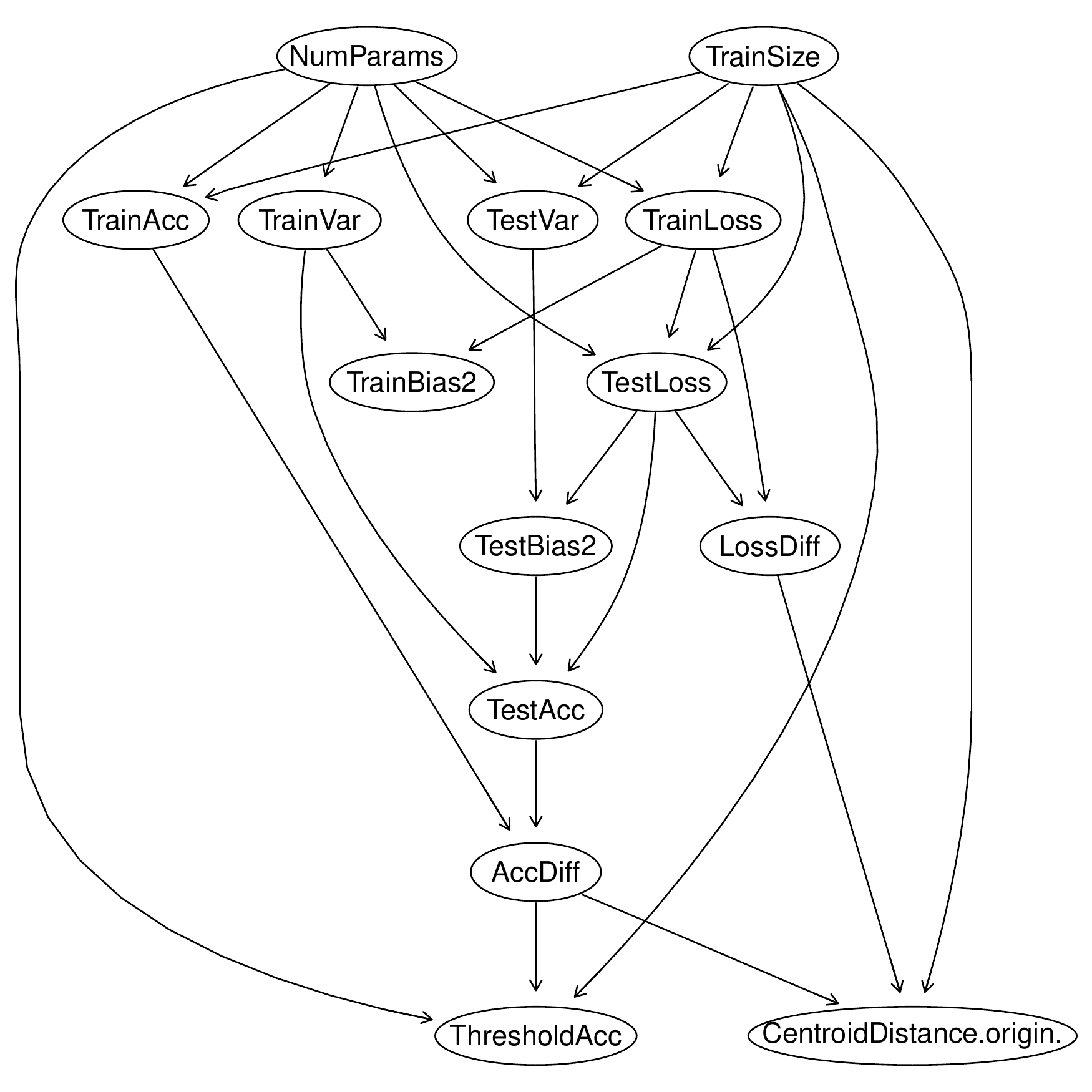}
    \subcaption[]{The causal model \tool infers for the target \threshacc (CE-trained models with regularization).}
    \label{fig:ce-reg-ThreshAcc}
\end{subfigure}%
\hfill
\begin{subfigure}[t]{0.45\linewidth}
    \centering
    \includegraphics[width=\linewidth]{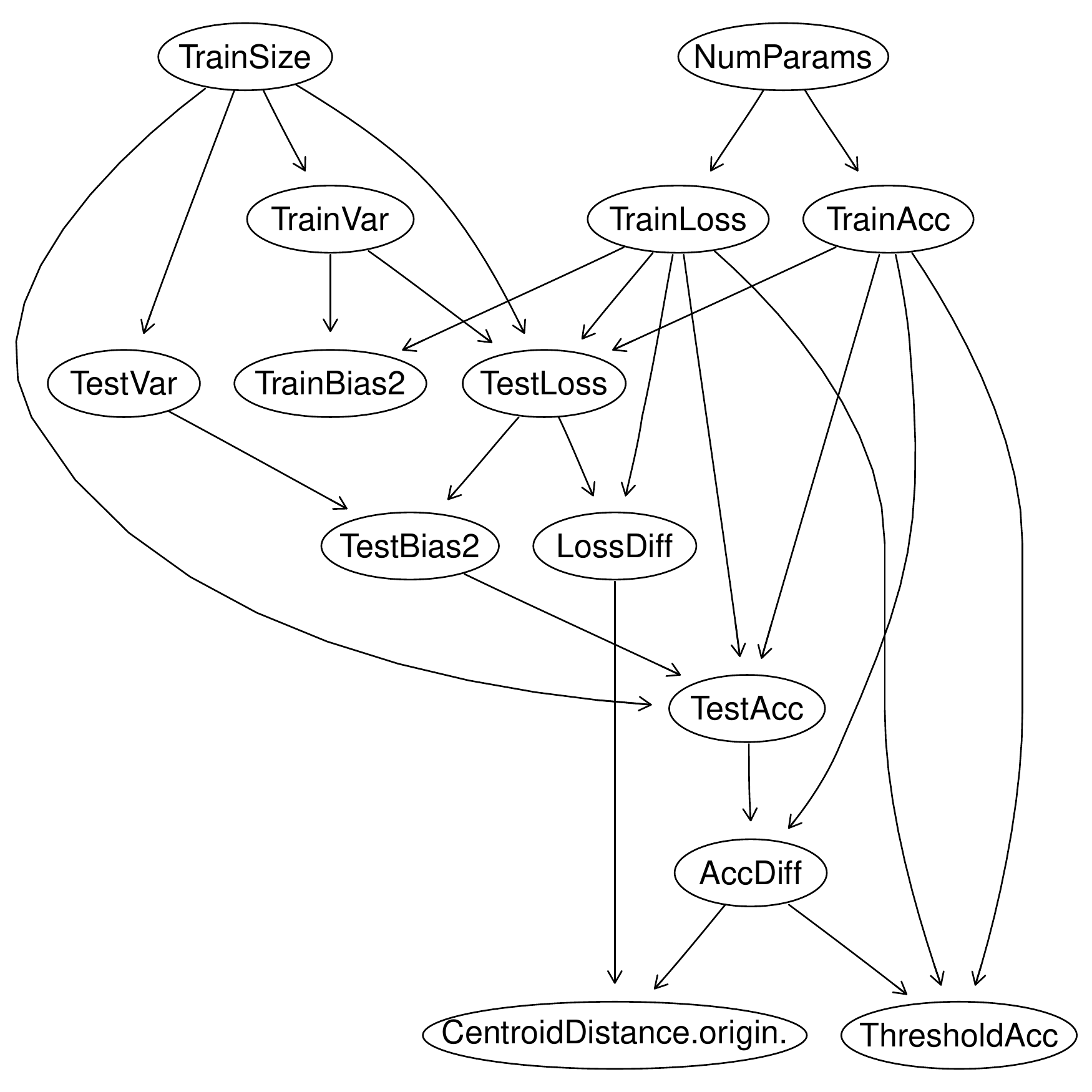}
    \subcaption[]{The causal model \tool infers for the target \threshacc (MSE-trained models with regularization).}
    \label{fig:mse-reg-ThreshAcc}
\end{subfigure}
\caption{The causal model \tool infers for the multiple shadow model attack for CE and MSE-trained models, where the target node is \threshacc. The models have been trained with L2-regularization (weight decay=$5\times10^-3$).}
\label{fig:reg-Shadow}
\end{figure*}

\begin{figure*}[t]
\centering
\begin{subfigure}[t]{0.45\linewidth}
    \centering
    \includegraphics[width=\linewidth]{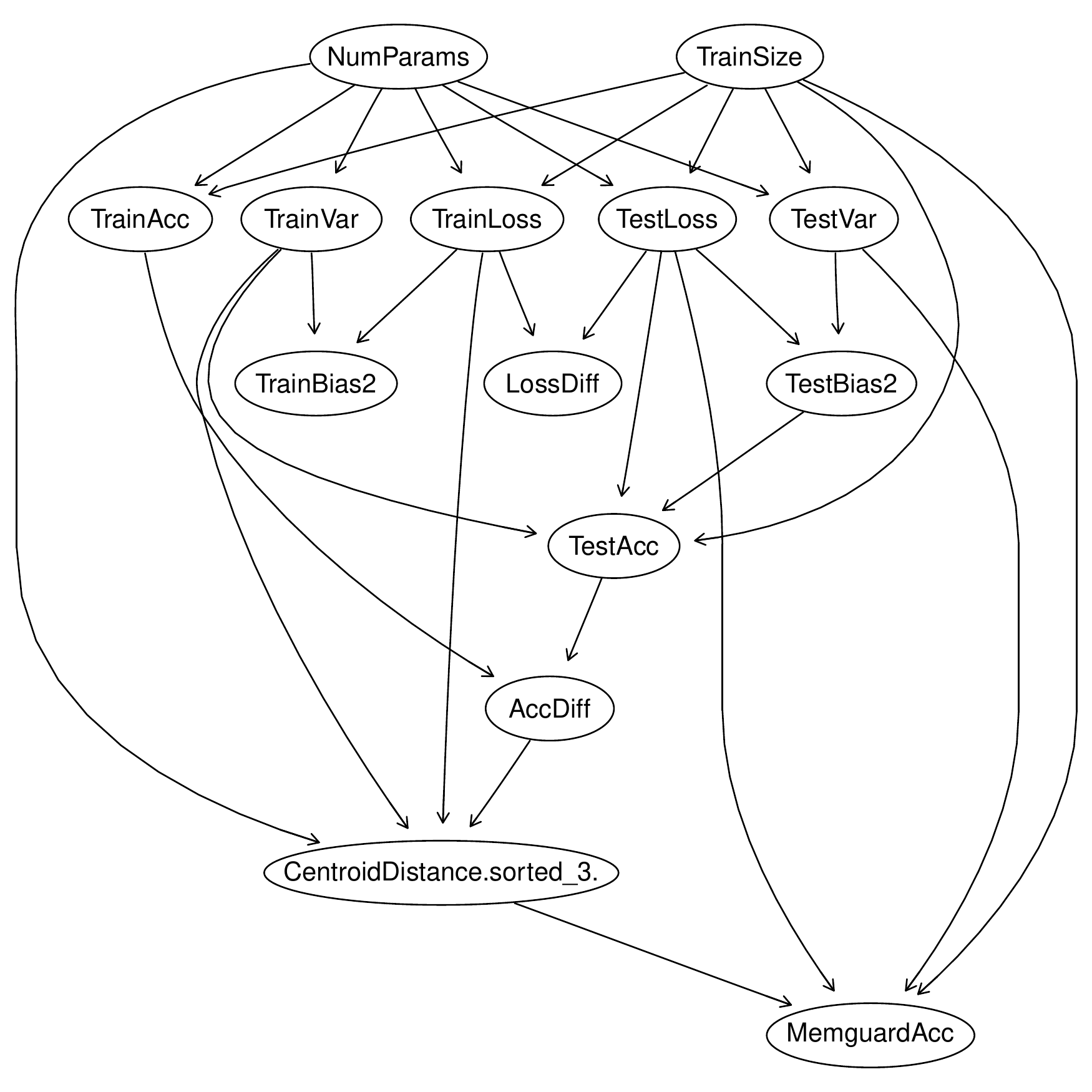}
    \subcaption[]{The causal model \tool infers for the target \memguard (CE-trained models).}
    \label{fig:ce-Memguard}
\end{subfigure}%
\hfill
\begin{subfigure}[t]{0.45\linewidth}
    \centering
    \includegraphics[width=\linewidth]{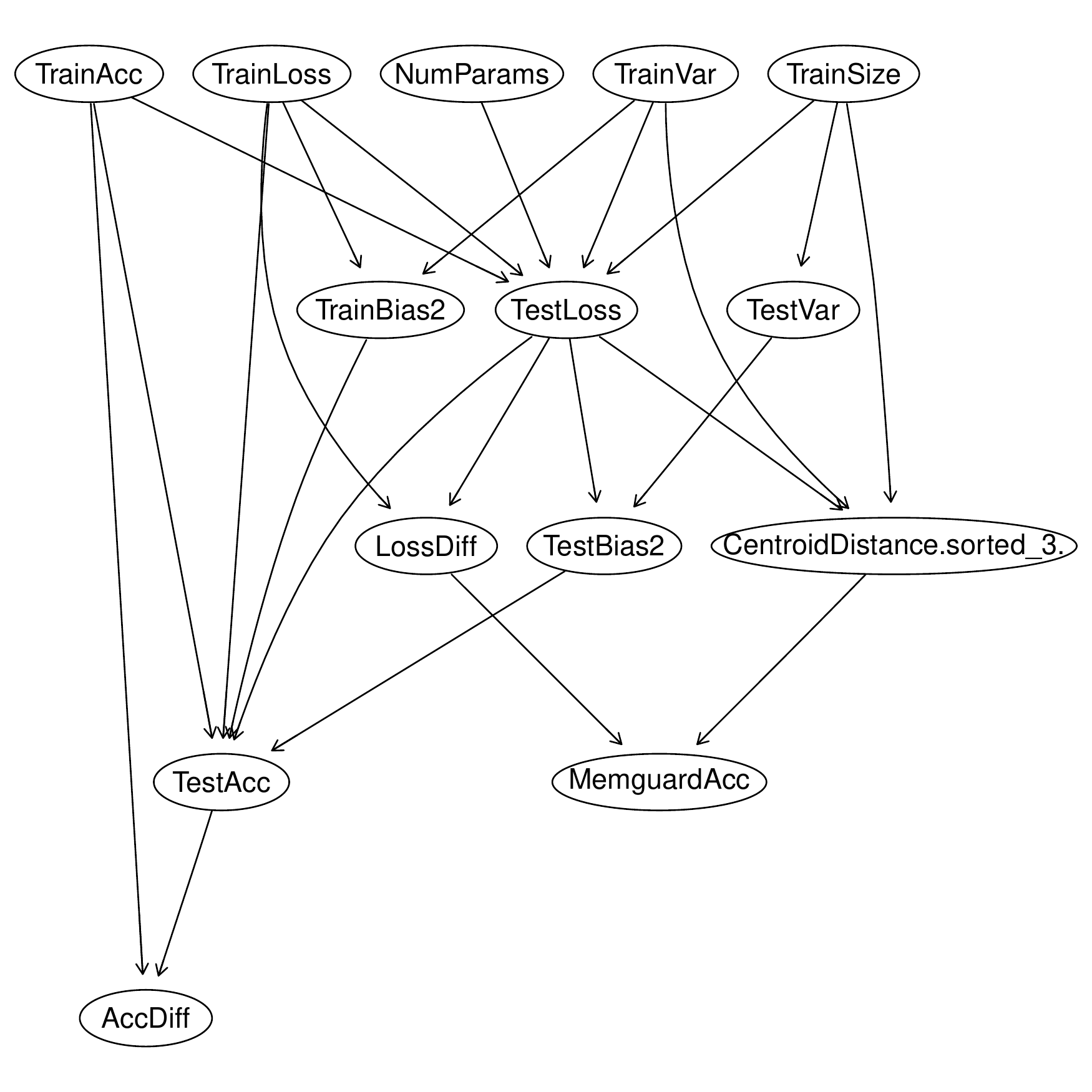}
    \subcaption[]{The causal model \tool infers for the target \memguard (MSE-trained models).}
    \label{fig:mse-Memguard}
\end{subfigure}
\caption{The causal model \tool infers for the single shadow model on MemGuard defended models. The
    target node in this case \memguard represents the accuracy of the MLLeak attack with top-3
    predictions on defended models. }
\label{fig:undefended-Thresh}
\end{figure*}

\end{appendix}

\end{document}